%% file: iclr2026_conference.tex
\documentclass{article} 
\usepackage{iclr2026_conference,times}

\usepackage{hyperref}
\usepackage{url}
\usepackage{graphicx}
\usepackage{booktabs}
\usepackage{multirow}
\usepackage{wrapfig}
\usepackage{amsmath}
\usepackage{amssymb}
\usepackage[table]{xcolor}
\usepackage{caption}
\usepackage{algorithm}
\usepackage{algpseudocode}
\usepackage{listings}
\usepackage{tabularx}
\usepackage{enumitem}
\setlist[itemize]{leftmargin=*}
\lstdefinestyle{artifact}{
  basicstyle=\ttfamily\tiny,
  breaklines=true,
  columns=fullflexible,
  keepspaces=true,
  showstringspaces=false,
  aboveskip=3pt,
  belowskip=5pt
}

\newcommand{\framework}{\textsc{SWE-Duel}}

\title{
\raisebox{-.11cm}{\includegraphics[width=0.7cm]{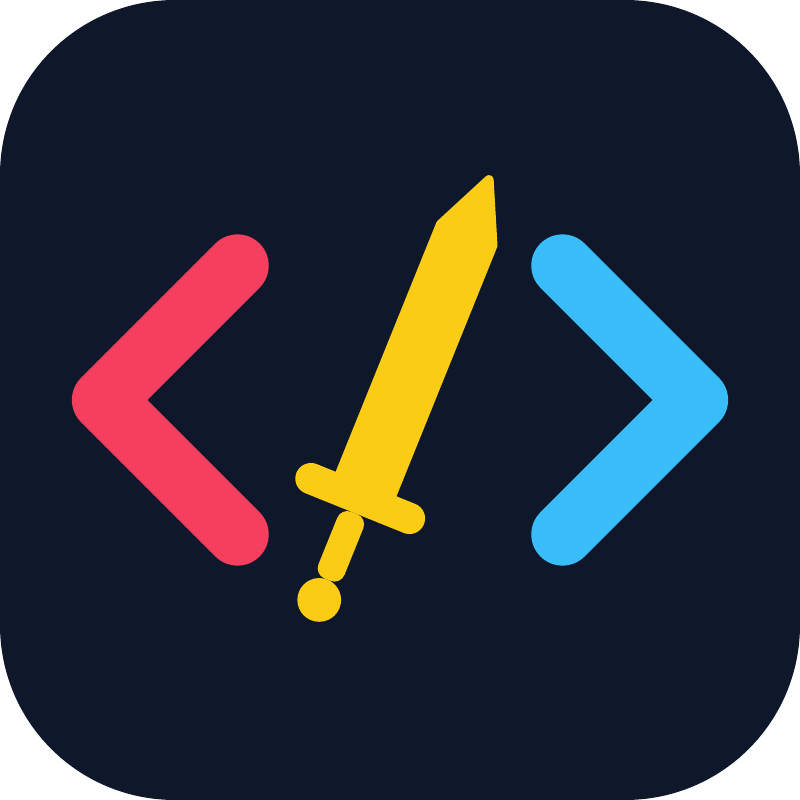}}~
\framework{}: Towards A Self-Scaling Arena for Coding Agents}

\author{Wei Minn, Zhensu Sun\thanks{ Corresponding Author}~, Lwin Khin Shar \& David Lo \\
School of Computing and Information Systems\\
Singapore Management University\\
\texttt{wei.minn.2023@phdcs.smu.edu.sg}; \texttt{\{zssun,lkshar,davidlo\}@smu.edu.sg}
}

\iclrfinalcopy 
\begin{document}

\maketitle

\vspace{-2em}
\begin{center}

\href{https://sweduel.github.io/SWE-Duel/}{
\raisebox{-.15cm}{\includegraphics[width=0.5cm]{Figures/swe-duel-logo.pdf}}~
\url{https://sweduel.github.io/SWE-Duel}
}

\end{center}

\input{sections/abstract}

\noindent
\begin{minipage}{\textwidth}
\centering
\includegraphics[width=\linewidth, trim={0cm 0cm 0cm 0.75cm}, clip]{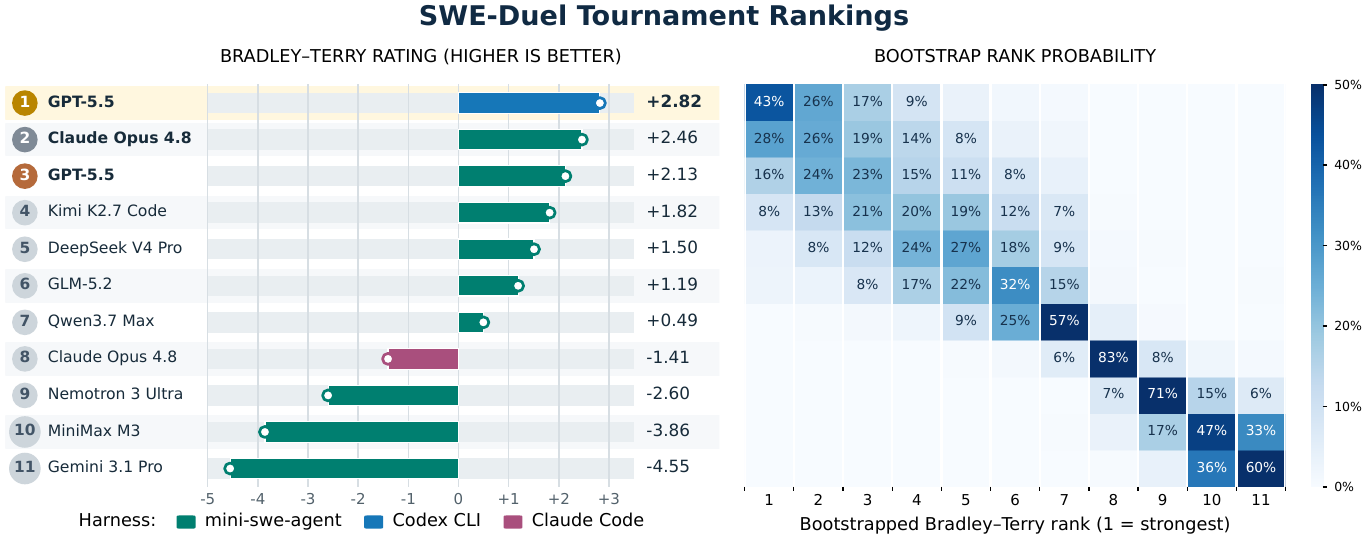}
\captionsetup{hypcap=false}
\captionof{figure}{
    \textbf{SWE-Duel Tournament Rankings
    after a round-robin tournament involving 11 participants.
    }
    Gemini 3.1 Pro with mini-swe-agent is a late entrant, demonstrating that new
    LLMs or harnesses can be evaluated without discarding the current tournament.
    }
\label{fig:headline-elo}
\end{minipage}

\input{sections/1.intro}
\input{sections/related_work}
\input{sections/framework}

\input{sections/experiments}
\input{sections/conclusion}

\bibliography{iclr2026_conference}
\bibliographystyle{iclr2026_conference}

\input{sections/ethics}

\appendix
\input{sections/appendix}

\end{document}

%% file: sections/abstract.tex
\begin{abstract}
Static coding benchmarks are fixed at release. Over time their tasks leak into training data, and stronger agents eventually saturate them. We argue that a benchmark should instead aim to be self-scaling: its tasks and oracles are created at evaluation time, so no content exists in advance to leak, and its difficulty is set by the participating agents rather than fixed at release. We introduce \framework{}, an arena in which two coding agents evaluate each other over mirrored Red–Blue turns.
In each turn, the Red agent works on a real repository across three context-isolated sessions.
It proposes and implements a new feature with feature tests, then embeds a subtle bug in that feature along with sealed bug-triggering tests. The resulting pull request is admitted only if (1) it passes the feature tests and the repository's existing tests while failing exactly the bug tests, and (2) a fresh session of the same agent can repair the bug blind, which rules out unsolvable tasks. The Blue agent then reviews the pull request without being told a bug exists.
It wins only if its revision passes all three test suites, and then the two agents swap roles.
We run a tournament among 11 model–harness participants on 15 repositories in five languages for 1,500 turns. The leaders win in different ways. GPT-5.5 with Codex CLI ranks first as the strongest solver, repairing 98.3\% of the defects it reviews. Claude Opus 4.8 with mini-swe-agent ranks second as the strongest task author, with a 27.3\% Red win rate. Finally, within one model family, tasks written by the newest version defeat a fixed-reference solver far more often than those of older versions (42.9\% vs. 4.0\%), giving preliminary evidence that task difficulty scales with task authoring capability.
\end{abstract}

%% file: sections/1.intro.tex
\section{Introduction}

Benchmarks are the primary instrument for measuring progress in coding agents.
To deliver trustworthy signal over the long run, a benchmark must remain absent from the training data of the models it evaluates and challenging for agents to solve.
This is increasingly difficult to satisfy, as LLMs continually learn from publicly available materials and grow more capable with every generation.
The deeper issue is that most existing coding benchmarks are static once released and built from openly accessible sources.
Such content is eventually absorbed into training corpora, and a fixed task set is eventually saturated by stronger agents.
In other words, from the moment a benchmark is released, it begins to leak into training data and to fall behind the agents it is meant to measure.
We argue that avoiding this consequence requires a benchmark to be \emph{self-scaling}: its tasks and oracles are created at evaluation time, so no content exists in advance to leak, and its difficulty is set by the participating agents rather than fixed at release, so it can rise as those agents improve.

Existing benchmarks fall short of this property in different ways.
Static benchmarks have already run into this problem: OpenAI recently stopped reporting results on SWE-bench Verified~\citep{jimenez_swe-bench_2024} after finding that frontier models could reproduce its gold patches.
Continuously updated benchmarks such as SWE-bench-Live~\citep{zhang_swe-bench_2025} and SWE-rebench~\citep{badertdinov_swe-rebench_2025} mitigate leakage by collecting tasks created after model training cutoffs, but their task difficulty is still bounded by the daily issues faced by human developers.
Most recently, model-generated tasks have emerged as a new direction: BugPilot~\citep{sonwane_bugpilot_2025} collects bugs introduced unintentionally by agents implementing new features, and SwingArena~\citep{xu_swingarena_2026} pairs LLMs as patch submitters and test-writing reviewers in an adversarial arena.
However, BugPilot generates its tasks ahead of time as a fixed set and SwingArena still draws tasks from a fixed pool of human-written GitHub issues, generating only the oracles at evaluation time; neither produces tasks whose difficulty grows with the agent being evaluated.
This motivates the central question of this work: \textbf{Can we create a self-scaling benchmark for coding agents?}

To answer this question, we present \framework{}, a self-scaling arena with a evaluation protocol in which two coding agents evaluate each other over mirrored adversarial turns.
In each turn, the Red agent, working on a given repository across three context-isolated sessions, proposes a new feature, implements it together with feature tests, and finally injects subtle bugs along with bug-triggering tests.
The resulting PR is used in the duel only if 1) it passes the feature tests and the repository's regression tests while failing exactly the bug-triggering tests, and 2) the Red agent can repair its own bugs in a fresh session, which rules out unsolvable tasks.
The Blue agent then reviews the PR and submits a revision, winning the round if and only if the revision passes all three test suites; the two agents swap roles in the next turn.
Because every feature, bug, and oracle is created at evaluation time, these specific tasks cannot appear in earlier training data,
And because tasks are authored by the agents under evaluation, task difficulty is not fixed by a human-written pool; it can in principle rise as the authors improve, up to the limit set by the self-repair gate. Altogether, a single duel profiles a broad slice of the software engineering lifecycle, including requirement specification, feature implementation, test generation, code review, fault localization, and
program repair.

We evaluate \framework{} in a tournament among 11 participants, each pairing a frontier proprietary or open-weight model (e.g., GPT-5.5, Claude Opus 4.8, DeepSeek V4 Pro, and Gemini 3.1 Pro) with one of three agent harnesses, on 15 repositories in five languages. Ten participants play a full round robin, and Gemini 3.1 Pro joins afterwards as a late entrant, giving 50 matches and 1,500 turns.
The leaders win in different ways.
GPT-5.5 with Codex CLI ranks first as the strongest solver, repairing 98.3\% of the defects it reviews, while Claude Opus 4.8 with mini-swe-agent ranks second as the strongest task author, with a 27.3\% Red win rate. A late entrant can be placed with only five matches, and within one model family, tasks written by the newest version defeat a fixed reviewer far more often than those of older versions (42.9\% vs. 4.0\%), giving preliminary evidence that difficulty tracks authoring capability.

We make the following contributions:
\begin{itemize}
\item We introduce \framework{}, a self-scaling arena that evaluates coding agents with each other. It does not leak test content by design and determines the task difficulty with the participants' own authoring ability rather than fixed at release.
\item We build an open arena toolkit for \framework{}~\citep{toolkit_swe-duel_2026}, enabling the community to start their own tournament from scratch, or add a new agent to the existing leaderboard~\citep{leaderboard_swe-duel_2026}.
\item We conduct a 1{,}500-turn tournament over 11 (model, harness) pairs, demonstrating the broad applicability of \framework{} and revealing distinct behavioral profiles across models and harnesses~\citep{dataset_swe-duel_2026}.
\end{itemize}

\begin{figure*}[t]
    \centering
    \includegraphics[width=\textwidth, trim={.65cm .50cm .7cm .25cm},clip]{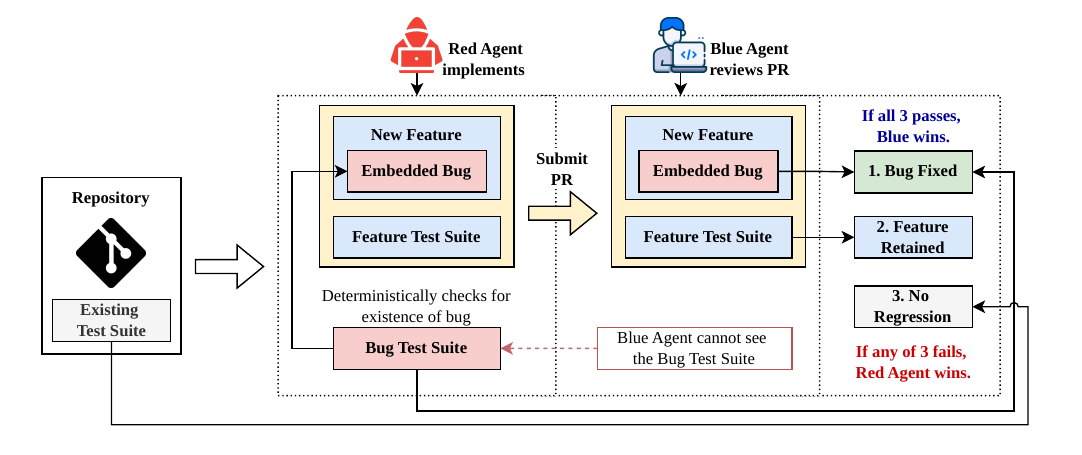}
    \caption{\textbf{A contamination-resistant evaluation in \framework{}}.
    Red agent implements a new feature with an embedded bug, so that the Blue agent cannot simply revert 
    the repository to its original version to remove the bug as doing so would remove the implemented feature.
    Feature retention is deterministically verified with the feature test suite, while
    bug removal is deterministically verified with the bug test suite.
    Existing test suite verifies no regression by the Blue agent's patch.
    }
    \label{fig:intro-overview}
\end{figure*}

%% file: sections/related_work.tex
\section{Related Work}

\noindent \textbf{Repository-level coding benchmarks.}
SWE-bench~\citep{jimenez_swe-bench_2024,openai_introducing_2024}
moved evaluation from isolated functions to real issue resolution.
Later benchmarks add automated environment setup~\citep{vergopoulos_automated_2025},
more languages~\citep{rashid_swe-polybench_2025},
long-horizon software evolution~\citep{le_swe-evo_2026, chen_swe-ci_2026},
the full issue-resolution cycle~\citep{guan_swe-cycle_2026},
and feature addition~\citep{liu_race-bench_2026}.
In all of them, the tasks and reference patches are public and fixed before evaluation.
However, public tasks eventually enter training data of evaluated LLMs~\citep{matton_leakage_2024}.
The benchmarks thus need to be repeatedly updated to replace the leaked ones.
For example, SWE-bench-Live~\citep{zhang_swe-bench_2025} and
SWE-rebench~\citep{badertdinov_swe-rebench_2025} regularly collect new tasks from GitHub to remain temporally fresh.
\citep{guan_is_2026} and DyCodeEval~\citep{chen_dynamic_2025} propose to perturb the tasks to avoid exact memorization, but preserve the original tasks and solutions.
Moreover, the tasks come from human development history (issues, pull requests, and releases), so their difficulty is fixed at release, and a fixed task set is saturated as agents improve.
The usual response is to build a harder benchmark, such as
SWE-Bench Pro~\citep{deng_swe-bench_2025}, but each replacement again has a fixed difficulty.
Refreshing the benchmark does not solve this as the new tasks come from the same distribution, so the tasks are newer but not harder.
In SWE-Duel, both the tasks and their oracles are written at evaluation time by the agents being evaluated.
No task exists in advance to leak, and difficulty is set by the
participants instead of a fixed human-written pool.

\noindent \textbf{Automatic bug generation.}
Early work injects bugs with predefined or learned patterns:
symbolic synthesis of deep faults~\citep{roy_bug_2018},
bug seeding learned from past fixes~\citep{patra_semantic_2021},
learned contextual mutations~\citep{richter_learning_2022}, and
re-introducing fixed security bugs~\citep{zhang_fixreverter_2022}.
These methods show that realistic bugs can be generated automatically, but each bug is a local edit drawn from a pattern set, so the variety of bugs is limited by the patterns.
LLMs remove this limit.
SWE-smith~\citep{yang_swe-smith_2025} prompts LLMs to modify or rewrite functions,
BugPilot~\citep{sonwane_bugpilot_2025} collects the regressions that agents cause by accident while adding features,
and Self-play SWE-RL~\citep{wei_swe-rl_2025} trains one model to both inject and repair bugs.
Both target training data, and they show that LLMs can produce diverse and hard bugs at scale.
BugPilot is the closest to our task construction, since its bugs also arise while an agent adds a feature.
The key difference is what the solver can see.
A BugPilot bug is kept only if it breaks existing tests, so running the test suite reveals it.
A \framework{} bug is admitted only if all visible tests still pass, so the solver has to find it by reading the code, and only a sealed test can confirm the fix.
Our bugs are also deliberate, so the task yield does not depend on agents making mistakes, which may become rarer as agents get more reliable.
For evaluation, CDBench~\citep{romazanov_cdbench_2026} lets an attacker LLM write mutants for a defender LLM, but the mutants are local edits and the defender is shown the mutant diff.
In \framework{}, each agent hides a bug inside a new feature for the other to fix, and a task is admitted only after a fresh session of its author repairs the bug blind, so every task is known to be solvable.

\noindent \textbf{Arena-style evaluation.}
Chatbot Arena~\citep{chiang_chatbot_2024} ranks LLMs from pairwise human votes, and
Arena-Lite~\citep{son_arena-lite_2025} and Auto-Arena~\citep{zhao_auto-arena_2025}
replace human voters with LLM judges and tournament schedules.
Arenas stay fresh because new prompts keep arriving, but their outcomes depend on judges,
whose preferences are noisy and are not the same as correctness.
SwingArena~\citep{xu_swingarena_2026} brings the arena format to software engineering
and lets repository CI decide the outcome: submitters patch GitHub issues and reviewers write tests.
However, its tasks come from a fixed pool of human-written issues, and reviewer tests are validated against a human gold patch.
\framework{} keeps execution-based scoring but removes both human anchors:
the competing agents write the tasks themselves, and task validity is checked by
execution gates and a blind self-repair instead of a gold patch.
As a result, task difficulty is set by the participants and not by a human-written pool.

%% file: sections/framework.tex
\section{\framework{}: A Self-scaling Coding Arena}
\label{sec:framework} 

\framework{} evaluates coding agents through Red--Blue duels (Figure~\ref{fig:intro-overview}).
A duel is played by two participants on a real repository $R$, and each participant plays both roles in turn.
The \emph{Red} side creates a task which is a pull request that adds a new feature to $R$ and hides a bug inside that feature.
The \emph{Blue} side solves the task by reviewing the pull request and submits a fixed version if the bug is noticed.

Formally, a task is a tuple $C = (R, s, d, T_f, T_b)$, where $s$ is the feature specification, $d$ is the pull-request diff that contains both the feature and the bug, $T_f$ is a feature test suite, and $T_b$ is a bug test suite that fails on $d$ only because of the bug.
Blue receives only the public part, $\mathrm{pub}(C) = (R, s, d, T_f)$.
$T_b$ stays sealed, so Blue is not told whether or where a bug exists.
Blue returns a revised diff $d'$ and wins the turn if and only if the repository's existing tests $T_e$, the feature tests $T_f$, and the sealed bug tests $T_b$ all pass on $d'$; otherwise Red wins.
Because the bug lives inside the feature, Blue cannot win by reverting the pull request, since $T_f$ would then fail.

Red's tasks are created and validated before the tournament and stored in a task bank
(Section~\ref{sec:generation}), so every opponent faces the same tasks from a given Red.
In a match, the two participants defend each other's banked tasks on every repository
(Section~\ref{sec:turn}), and a tournament is a set of matches among many participants (Section~\ref{sec:tournament}).

\subsection{Participants and Environment}
\label{sec:environment}

A participant is a pair $(m, h)$, where $m$ is the model and $h$ is the harness that provides
the agent loop, tools, context policy, command execution, and recovery behavior.
The same model under two harnesses counts as two participants, each with its own task bank and records.

Every duel runs inside a Docker container built for one open-source repository pinned to a specific commit.
The repository's existing test suite records the original behavior and is re-run on every later artifact
(the new feature, the embedded bug, and Blue's patch) to detect regressions.
Red sessions, validation gates, and scored Blue runs all start from the same image,
so every participant works in an identical environment.

\subsection{Red Agent: Task Creation}
\label{sec:generation}

\begin{figure*}[t]
\centering

\includegraphics[width=\textwidth, trim={.5cm .250cm .5cm .25cm},clip]{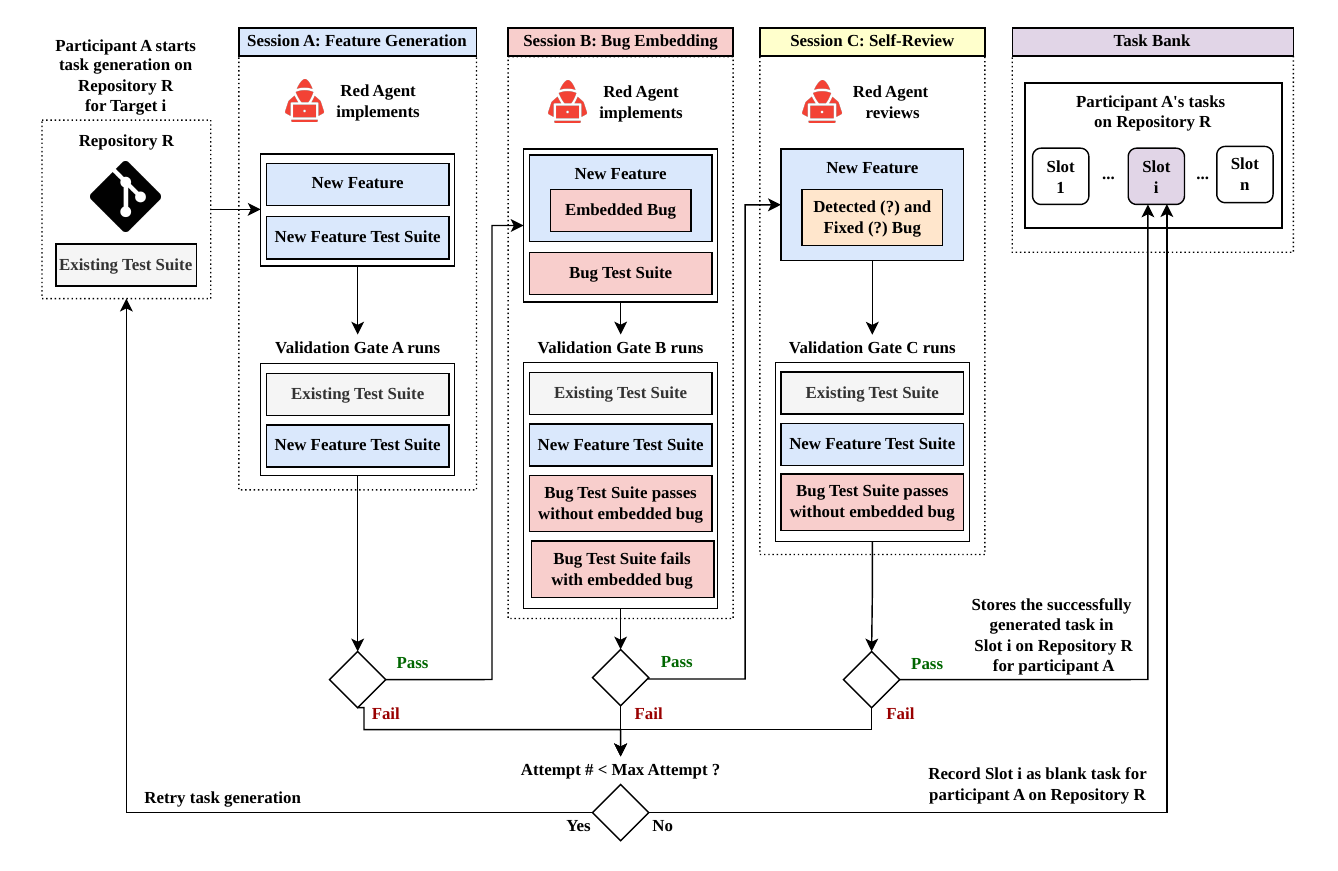}
\caption{\textbf{Three-session task generation.} 
Each of the $n$ targets is stored inside a slot. 
Each slot permits at most five attempts: a challenge successfully generated within 5 attempts
is stored inside its corresponding slot, whereas if it exceeds five attempts, it leaves an exhausted slot.
Validation gate A ensures the verifiability of the implemented feature. 
Validation gate B ensures the verifiability of the embedded bug. 
Validation gate C ensures the fairness of the embedded bug by verifying 
the same-system reviewer detects and removes the bug from the new feature.
}
\label{fig:challenge-generation}
\end{figure*}

Red creates each task in three sessions, and each session is followed by an executable
validation gate (Figure~\ref{fig:challenge-generation}).
Every session starts with a fresh context, so a session knows only what we pass to it.
For each participant and repository, we request $N$ tasks, one per \emph{slot}.
A slot allows at most five attempts.
The first attempt that passes all three gates is admitted and stored in the participant's task bank.
If all five attempts fail, the slot is \emph{exhausted} and stays empty.
We never replace an exhausted slot with a new one, because creating a valid task
(implementing a feature, embedding a bug, writing both test suites, and passing self-repair)
is part of the ability we measure.
An exhausted slot later counts as a missing task in a match (Section~\ref{sec:tournament}).

\noindent \textbf{Session A: Feature Implementation}
\label{sec:session-feature}
Red explores the repository $R$, picks a target module on its own, implements a new feature,
and writes the feature specification $s$ and the feature tests $T_f$.
We denote the resulting feature-only diff by $d_f$.
To keep a participant's tasks on one repository diverse, Red is shown the specifications
and target files of its earlier admitted tasks there and must pick a different module.
The validation gate of session A requires that the existing test suite $T_e$ and generated feature tests $T_f$ pass on $d_f$ and that the diff and the tests meet minimum size thresholds.
Tests that already fail on the pinned commit are recorded and ignored.
This gate establishes that the feature is correct before any bug is added,
so a later failure can be attributed to the bug and not to a broken feature.

\noindent \textbf{Session B: Bug Embedding}
\label{sec:session-bug}
A fresh Red session receives the feature specification $s$ and the diff $d_f$.
It edits the feature code to embed a bug, which gives the final diff $d$, and writes the bug tests $T_b$.
The validation gate of session B requires that
(i) $d$ applies to the pinned commit and passes lint or compilation;
(ii) the build, $T_e$, $T_f$, and the size checks still pass on $d$; and
(iii) $T_b$ passes on $d_f$ but fails on $d$.
Condition (ii) ensures that the bug gives Blue no visible failing test.
Condition (iii) ensures that $T_b$ tests the bug and not the feature,
and that the presence of the bug can be checked by execution.

\noindent \textbf{Session C: Blind Self-Repair}
\label{sec:session-review}
Deterministic gates of session A and B show that the embedded bug exists and is hidden, but not that it can be detected and fixed from what a reviewer sees.
Therefore, a fresh instance of the same participant receives only the public bundle $\mathrm{pub}(C)$ (i.e., repository, feature specification, pull-request diff, and feature tests), which is exactly what Blue will later receive, and submits a revised diff $d'$.
The task is admitted only if $T_e$, $T_f$, and $T_b$ all pass on $d'$,
which is the same condition under which Blue wins a turn.
This keeps unsolvable tasks out, but it also means a task can be no harder than its author's own bug-finding ability.
We view this as a deliberate trade-off between solvability and difficulty.
Admitted tasks are saved in a bank and reused in every match of a tournament, so all opponents face the same tasks from a given Red.

\subsection{Blue Agent: Task Solving}
\label{sec:turn}

Blue receives the public bundle $\mathrm{pub}(C)$ of a task $C$ from Red's bank,
exactly as the reviewer in Session C does (Section~\ref{sec:session-review}).
The bug tests $T_b$ and the bug metadata stay hidden.
Blue starts from a fresh context in the same container image and is asked to review the
pull request as if it came from a contributor.
The prompt does not say that a bug exists.
Blue may edit the code, including rewriting the feature, or leave the pull request as it is,
and then submits its final diff $d'$.

The framework then runs the three test suites on $d'$.
Let $S_e, S_f, S_b \in \{0,1\}$ indicate that $T_e$, $T_f$, and $T_b$ pass.
We deliberately use a hard conjunction:
\begin{equation}
S_{\mathrm{Blue}}=S_eS_fS_b, \qquad
S_{\mathrm{Red}}=1-S_{\mathrm{Blue}}.
\label{eq:score}
\end{equation}
Blue wins the turn only if all three suites pass. There is no partial credit for a plausible but incomplete patch.
Blue can lose in three ways: the bug ships ($S_b=0$), the feature is broken or removed ($S_f=0$),
or existing behavior regresses ($S_e=0$).
If Blue submits the pull request unchanged, Gate B guarantees that $T_b$ fails, so Red wins.

\subsection{Tournament}
\label{sec:tournament}

A match between participants $A$ and $B$ is played over a set of repositories
$\mathcal{R}=\{R_1,\ldots,R_{|\mathcal{R}|}\}$ with $N$ tasks per repository.
Let $C_{c,j,i}$ denote the $i$-th admitted task of participant $c$ on repository $R_j$.
For every $R_j$ and every $i \le N$, the match has two turns:
$B$ plays Blue on $C_{A,j,i}$, and $A$ plays Blue on $C_{B,j,i}$.
A match therefore has $2N|\mathcal{R}|$ turns, and each participant plays Red and Blue
equally often on the same repositories.
If a participant admitted fewer than $N$ tasks on $R_j$ because its slots were exhausted,
each missing task is scored as a Blue win without running Blue.
We call all other turns \emph{contested}.
Missing tasks thus lower their author's match score.
Each turn gives one point to Blue if $S_{\mathrm{Blue}}=1$ (Eq.~\ref{eq:score}) and to Red otherwise.
A participant's match score $Q$ is the sum of its points in both roles.
$A$ wins the match if $Q_A > Q_B$, loses if $Q_B > Q_A$, and draws otherwise.

Following Chatbot Arena~\citep{chiang_chatbot_2024} and Arena-Lite~\citep{son_arena-lite_2025},
\framework{} uses a round-robin format in which every pair of participants plays one match.
To add a new participant to a finished tournament of $M$ participants,
the new participant first creates its own task bank and then plays against only
$\lceil \log_2 M \rceil + 1$ incumbents, selected with the active sampling rule of
Chatbot Arena~\citep{chiang_chatbot_2024}.
The incumbents' task banks and all earlier match results are reused unchanged,
so the number of new matches grows logarithmically with $M$.

%% file: sections/experiments.tex
\section{Experiments}
\label{sec:experiments}


\subsection{Experimental Setup}

\begin{table*}[t]
\centering
\caption{
    \textbf{Main tournament results.} 
    The rows are sorted in descending order by the full-schedule Bradley-Terry log-strength rating in the highlighted column. 
    Full-schedule means the missing-Red auto-wins are retained. 
    Figure~\ref{fig:rank-trajectory} shows the trajectory
    of the rankings throughout round-robin tournament rounds before arriving at this final ranking.
}
\label{tab:match-leaderboard}
\small
\setlength{\tabcolsep}{2pt}
\resizebox{\textwidth}{!}{\input{Tables/tab_match_leaderboard}}
\end{table*}

\textbf{Repository field.}
\framework{} uses 15 popular repositories across 5 programming languages (Python, Java, Go, JavaScript, C/C++) as shown in Table~\ref{tab:repositories} of Appendix~\ref{app:repository-field}. 

\textbf{Agents.}
We choose frontier models at the time of our experiment, including GPT-5.5~\citep{openai_introducing_2026}, Claude Opus 4.8~\citep{anthropic_introducing_2026}, Kimi K2.7 Code~\citep{kimi_kimi_2026}, DeepSeek V4 Pro~\citep{deepseek-ai_deepseek-v4_2026}, GLM 5.2~\citep{zai_glm-52_2026}, Qwen 3.7 Max~\citep{qwenteam_qwen37_2026},
Nvidia Nemotron 3 Ultra~\citep{nvidia_nemotron_2026}, Minimax M3~\citep{minimax_minimax_2026}, and Gemini 3.1 Pro~\citep{google_gemini_2026}.
All these models are paired with mini-swe-agent~\citep{yang_swe-agent_2024} as harness.
In addition, we include two more (model, harness) pairs: GPT-5.5 with Codex CLI~\citep{openai_non-interactive_2026} and Claude Opus 4.8 with Claude Code~\citep{anthropic_run_2026}.
Step budgets, recovery turns, decoding temperature, and the execution environment are detailed in Appendix~\ref{app:harness-access}.
All model APIs are accessed through OpenRouter between June 18 and August 30, 2026; per-model access dates are reported in Appendix~\ref{app:model-access}.



\textbf{Tournament Details.}
In our experiments, we start with an original group of ten participants.
To minimize variance in rankings, each of the original ten participants faced every other participant in a round-robin format, forming $\binom{10}{2}=45$ match pairings.
After the initial round-robin tournament, we demonstrate the late entrance by Gemini 3.1 Pro which plays $5$ active-sampled incumbents, giving $45 + 5 = 50$ matches. Since, we use $15$ repositories in \framework{}, we have $15 \times 2 = 30$ turns in a match. Thus, in total, we have $50$ matches $\times$ $30$ turns per match $=$ $1{,}500$ turns. 
We give automatic win to \emph{Blue} on missing \emph{Red} challenges so as to avoid survivor bias (dropping weak task creators), and also to consider task creation as SWE capability of the Red agent, not just task solving.
For completeness, we also report the alternative tournament ranking without automatic win for \emph{Blue} on missing \emph{Red} challenges, to include only the contested turns.


\textbf{Metric.}
Following existing works in arena-style evaluation of LLMs~\citep{chiang_chatbot_2024,zhao_auto-arena_2025,son_arena-lite_2025}, 
\framework{} reports tournament outcome using \textbf{Bradley--Terry (BT) ratings}~\citep{bradley_rank_1952}. It assigns each participant $p$ a log-strength $\beta_p$ (centered to zero mean), fitted by maximum likelihood over all match outcomes of Section~\ref{sec:tournament} under the model $\Pr(p \text{ beats } q) = \sigma(\beta_p - \beta_q)$ with $\sigma$ the logistic function and a draw counted as half a win for each side, from which we report rank probability distributions and 95\% confidence intervals using 1,000 bootstrap resamples of matches.
\subsection{Main Results}

\textbf{Overall ranking.}
Figure~\ref{fig:headline-elo} depicts the tournament rankings of all 11 agents in BT ratings, and the probability distribution of the rankings based on the participants' BT ratings.
Meanwhile, Table~\ref{tab:match-leaderboard} reports definitive full-schedule BT ranking together with
comprehensive statistics such as Win/Loss/Draw, and Bradley--Terry ratings.
GPT-5.5 with Codex harness leads the final ranking in BT ratings, 
followed in order by Claude Opus 4.8, Kimi-K2.7 Code, and DeepSeek-V4-Pro,
all with mini-swe-agent harness.
Meanwhile, MiniMax-M3 and the late entrant Gemini 3.1 Pro rank the lowest. 
Table~\ref{tab:bt-uncertainty} presents $95$\% confidence intervals for the rankings and BT ratings.
The six participants with strongest BT ratings
(GPT-5.5 under both harnesses, Opus 4.8 with mini-swe-agent, Kimi, DeepSeek, and GLM) form a broad leading group,
whereas the lower group comprises Opus 4.8 with Claude
Code, Nemotron and MiniMax, and the five-match Gemini late entrant.
Appendix~\ref{app:turn-level-bt} reports the complementary turn-level BT refit of the main tournament.

\textbf{Distinct profiles in creating and solving tasks.}
In Figure~\ref{fig:role-rates}, we report normalized, challenge-clustered role rates.
The two leaders reach the top in very different ways.
GPT-5.5 is the strongest solver under both harnesses, with Blue win rates of $98.7$\% with Codex CLI and $97.0$\% with mini-swe-agent.
In contrast, Claude Opus 4.8 with mini-swe-agent is the strongest task creator: its Red win rate of $27.3$\% is the highest in the field and more than twice that of either GPT-5.5 participant ($11.3$\% and $10.4$\%), which suggests that it embeds bugs that are harder to find.
DeepSeek-V4-Pro's Red:Blue record of $18.5$\%--$91.9$\% is seemingly better than that of Kimi-K2.7-Code ($15.6$\%:$90.4$\%).
Yet, Kimi ranks above DeepSeek on the BT ratings (with 95\% CIs overlapping) due to Kimi having one less loss than DeepSeek.
At the lower end, MiniMax-M3 ($2.0$\%) and the Gemini 3.1 Pro ($2.7$\%) have the weakest task creation capabilities, thus often failing to generate admitted tasks, and thereby give the Blue opponent an automatic win on the task solving turn. 
Appendix~\ref{app:estimand-decomposition} and Appendix~\ref{app:red_blue_transfer} ablates the automatic wins, and analyzes the two capabilities separately.

\textbf{Task solving rate and its breakdown.}
Across all $1{,}500$ recorded turns, Blue succeeds $87.9$\% of the time (95\% CI $[84.6\%,90.9\%]$).
Focusing on the $1{,}219$ contested turns, Blue succeeds $85.1$\% (95\% CI $[81.0\%,88.6\%]$): $1{,}037$ clean repairs, $158$ cases where the latent bug ships, and $24$ collapses that break the feature or existing behavior.
Figure~\ref{fig:asymmetry} 
in Appendix~\ref{app:defect-composition}
breaks contested success down by programming language and by bug category, with $95$\% task-clustered intervals.
By language, Blue success is $89.2$\% for Java, $85.7$\% for Go, $83.9$\% for C/C++, $83.5$\% for Python, and $82.9$\% for Node.
By self-reported bug category, logic errors are the hardest to repair ($81.9$\%), followed by security vulnerabilities ($84.9$\%), off-by-one errors ($88.9$\%), type confusion ($90.6$\%), and input validation ($95.5$\%).

\begin{figure}[t]
\centering
\begin{minipage}[t]{0.475\linewidth}
\vspace{0pt}
\centering
\captionof{table}{Fixed-reference solver failure rates across the task-author capability ladder.
Successive versions of one model family (all with mini-swe-agent) each
authored tasks over the same repositories.
The $13$ GLM-5.2 draws are the solver's in-match
tournament solving runs reused verbatim; 
the other $26$ ran fresh with zero infrastructure errors.
}
\label{tab:capability-scaling}
\small
\setlength{\tabcolsep}{2pt}
\resizebox{\linewidth}{!}{%
\begin{tabular}{lrrl}
\toprule
\textbf{Task author} & \textbf{Tasks} & \textbf{Yield} & \textbf{Solver failure (95\% Wilson)} \\
\midrule
GLM-5.1 & 12 & 30.8\% & $1/12 = 8.3\%$ $[1.5, 35.4]$ \\
GLM-5.2 & 13 & 40.6\% & $0/13 = 0.0\%$ $[0.0, 22.8]$ \\
GLM-5.3 & 14 & 37.8\% & $6/14 = 42.9\%$ $[21.4, 67.4]$ \\
\midrule
All versions & 39 & 36.1\% & $7/39 = 17.9\%$ $[9.0, 32.7]$ \\
\bottomrule
\end{tabular}%
}
\end{minipage}%
\hfill
\begin{minipage}[t]{0.475\linewidth}
\vspace{0pt}
\centering
\includegraphics[width=\linewidth, trim={1.5cm 1.4cm 0cm .8cm}, clip]{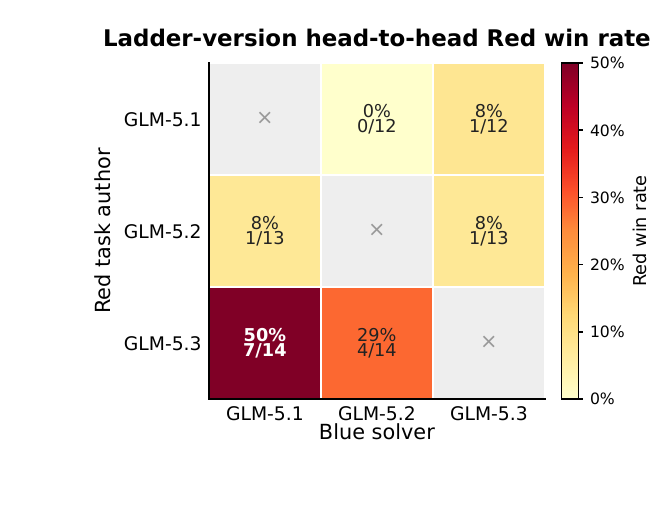}
\caption{
Head-to-head match-ups between the ladder versions (all with mini-swe-agent). 
}
\label{fig:capability-h2h}
\end{minipage}
\end{figure}

\textbf{Adding a new participant.}
To demonstrate how a new agent joins the existing leaderboard, we held Gemini 3.1 Pro with mini-swe-agent out of the ten-player round robin and added it only after all 45 incumbent matches had finished.
The newcomer first created its own task bank over the same 15 repositories under the same protocol.
\framework{} then scheduled $\lceil\log_2 10\rceil+1=5$ matches against incumbents chosen by active sampling, instead of all ten pairings.
The incumbents' task banks and earlier match results were reused unchanged.
Five matches were enough for a coherent placement.
Gemini finished $0$--$1$--$4$ (W/D/L) and ranks last in BT rating (Table~\ref{tab:match-leaderboard}), with the lowest contested Blue success in the field ($61.8$\%, Figure~\ref{fig:role-rates}).
Adding it also left the existing leaderboard untouched: refitting Bradley--Terry on the 45 incumbent-only matches reproduces the incumbent ordering of the full 50-match fit exactly (Kendall's $\tau=1.0$).
Appendix~\ref{app:newcomer} details the onboarding protocol.

\textbf{Tournament and newcomer cost.}
Running this tournament consumed $2{,}109$ USD of API spend: \$670 for Red task creation (including every failed attempt), \$158 for blind self-review validation, and \$1,281 for Blue evaluation, about \$$26$ per match. 
Per-participant totals are in Table~\ref{tab:costs} of Appendix~\ref{sec:resource-budgets}.
Onboarding the newcomer cost $\$243$: \$$126$ to author its task bank once, plus \$117 for its five matches, split between the entrant solving incumbent tasks (\$$61.81$) and incumbents solving the entrant's tasks (\$$55.64$).
A newcomer's marginal cost is therefore its own bank plus $O(\log N)$ active-sampled matches rather than the $O(N)$ pairings of a full round robin: had Gemini 3.1 Pro played all ten incumbents, we estimate the entrance at about \$379 (Appendix~\ref{app:newcomer}).




\subsection{Additional Analysis}
\textbf{Adaptive task difficulty.}
SWE-Duel assumes that stronger agents create harder tasks. The tournament cannot test this directly, because participants differ in more than strength.
We therefore fixed the solver and changed only the task author: GLM-5.1, GLM-5.2, and GLM-5.3 (all with mini-swe-agent) each wrote tasks on the same repositories, and DeepSeek-V4-Pro~$\times$~mini-swe-agent, a reviewer outside the family, solves all admitted tasks.
Table~\ref{tab:capability-scaling} shows the fixed-reference solver's failure rates when solving tasks
from the three task authors.
It fails on 8.3\% (GLM-5.1), 0.0\% (GLM-5.2), and 42.9\% (GLM-5.3) of them (exact trend test $p{=}.015$). 
It suggests that the tasks from two older versions are both too easy for DeepSeek-V4-Pro, while the newest is clearly harder.
We also ran a round-robin tournament among GLM-5.1, GLM-5.2, and GLM-5.3 that allows us to present head-to-head match-ups that further highlight the models' different capabilities (Figure~\ref{fig:capability-h2h}). 
GLM-5.3's tasks defeat GLM-5.1 in 7/14 (50.0\%) and GLM-5.2 in 4/14 (28.6\%) contested turns, while the two older versions only defeats GLM-5.3 once each: 1/12 for GLM-5.1, and 1/13 for GLM-5.2.
Appendix~\ref{app:capability-scaling} discusses the adaptive task difficulty experiment results in detail.

\begin{table*}[t]
\centering
\caption{Fixed reference task solver over the tasks used in the main tournament. 
The endogenous Red win rate is its Red success
over its contested-only tournament turns; 
the reference Red win rate is the fraction of its admitted tasks 
that the fixed reference solver fails to solve blind.
}
\label{tab:reference-attack}
\small
\resizebox{\linewidth}{!}{\input{Tables/tab_reference_attack}}
\end{table*}

\textbf{Fixed-reference task solver.}
In the tournament, a participant's Red win rate depends on which Blue it happened to meet.
To put all participants on one scale, we had a single mid-tier agent (GLM-5.2 with mini-swe-agent) solve all 134 tournament tasks (Table~\ref{tab:reference-attack}).
It fails on 15 of 134 tasks (11.2\%), and the Red win rates against it correlate with
the tournament rates (Spearman $\rho{=}0.66$) .
Claude Opus 4.8 with mini-swe-agent remains the strongest task creator (27.3\% in the tournament,
26.7\% against the reference). Kimi-K2.7-Code drops from 17.9\% to 0 of 13,
which suggests its tournament wins came from a few weaker defenders.
Further details about this experiment are in Appendix~\ref{app:reference-reviewer}.

\textbf{Admission stability.}
A task is admitted after a single blind self-repair run (Section~\ref{sec:session-review}), and that run is stochastic, so some tasks may have been admitted by luck.
To measure this, we re-ran the same check three more times on a stratified sample of 44 tournament tasks (4 per participant). 
Of these, 
35 were self-repaired in all three runs, 8 in two, and one in none.
The pooled self-repair rate is 91.7\% (95\% Wilson CI 85.7--95.3\%).
Admitted tasks are therefore mostly reliably solvable, and lucky admissions are
rare (Appendix~\ref{app:admission-noise}).

\textbf{Bug test case soundness.}
The sealed bug tests are written by Red, so a test that is too specific could 
incorrectly reject a legitimate repair.
We manually checked a seeded random sample of 50 of the 158 turns that 
Blue lost only on the bug tests.
We found 45 genuine misses, three valid repairs rejected by the test,
and two ambiguous cases: a 6.0\% false-negative rate (95\% Wilson CI 2.1--16.2\%).
All three false negatives come from two Go JWT tasks whose tests require one specific error type or option (more detail in Appendix~\ref{app:oracle-audit}).
Five language-stratified examples of challenges are in 
Appendix~\ref{app:worked-challenges}.

\textbf{Ablated tournament for harness.}
\label{sec:harness-ablation}
To investigate the effect of harnesses on the participants, we run an ablated tournament in which the
same model (DeepSeek-V4-Pro, GLM-5.2, or Kimi-K2.7-Code) plays against itself under mini-swe-agent, Codex CLI, and Claude Code.
Just like the main tournament, this ablation tournament covers the same 15 open-source
repositories.
Unlike the main tournament, there are no native pairings that could confound the harness 
effect analysis.
Mini-swe-agent leads with BT $[$95\% CI$]$ score of $+0.73\,[-0.25,+5.56]$ followed by
Codex CLI wtih $-0.25\,[-3.50,+1.34]$, and Claude Code with $-0.49\,[-3.83,+0.65]$
(Appendix~\ref{app:harness-details}). 

%% file: Tables/tab_match_leaderboard.tex
\begin{tabular}{lrrrrrrr}
\toprule
\textbf{Competitor} & \textbf{\shortstack{Matches\\played}} & \textbf{Won} & \textbf{Drawn} & \textbf{Lost} & \textbf{\shortstack{Red win\\ (\%)}} & \textbf{\shortstack{Blue win\\ (\%)}} & \cellcolor{gray!20}\textbf{\shortstack{Bradley--Terry}} \\
\midrule
GPT-5.5 [codex] & 10 & 8 & 1 & 1 & 11.3 & 98.7 & \cellcolor{gray!12}\textbf{+2.82} \\
Claude Opus 4.8 [mini-swe-agent] & 10 & 8 & 0 & 2 & 27.3 & 88.7 & \cellcolor{gray!12}\textbf{+2.46} \\
GPT-5.5 [mini-swe-agent] & 9 & 5 & 3 & 1 & 10.4 & 97.0 & \cellcolor{gray!12}\textbf{+2.13} \\
Kimi-K2.7-Code [mini-swe-agent] & 9 & 5 & 2 & 2 & 15.6 & 90.4 & \cellcolor{gray!12}\textbf{+1.82} \\
DeepSeek-V4-Pro [mini-swe-agent] & 9 & 5 & 1 & 3 & 18.5 & 91.9 & \cellcolor{gray!12}\textbf{+1.50} \\
GLM-5.2 [mini-swe-agent] & 9 & 4 & 2 & 3 & 14.8 & 89.6 & \cellcolor{gray!12}\textbf{+1.19} \\
Qwen3.7-Max [mini-swe-agent] & 9 & 3 & 2 & 4 & 13.3 & 91.1 & \cellcolor{gray!12}\textbf{+0.49} \\
Claude Opus 4.8 [claude-code] & 10 & 2 & 2 & 6 & 7.3 & 87.3 & \cellcolor{gray!12}\textbf{-1.41} \\
Nemotron-3-Ultra [mini-swe-agent] & 10 & 1 & 2 & 7 & 6.7 & 79.3 & \cellcolor{gray!12}\textbf{-2.60} \\
MiniMax-M3 [mini-swe-agent] & 10 & 1 & 0 & 9 & 2.0 & 74.7 & \cellcolor{gray!12}\textbf{-3.86} \\
Gemini-3.1-Pro [mini-swe-agent] & 5 & 0 & 1 & 4 & 2.7 & 72.0 & \cellcolor{gray!12}\textbf{-4.55} \\
\bottomrule
\end{tabular}

%% file: Tables/tab_reference_attack.tex
\begin{tabular}{lrrrrr}
\toprule
\textbf{Participant} & \textbf{$n$\,tasks} & \textbf{Endog.\ Red wins} & \textbf{Endog.\ rate} & \textbf{Ref.\ Red wins/$n$} & \textbf{Ref.\ rate (95\% Wilson)} \\
\midrule
Claude Opus 4.8 [mini-swe-agent] & 15 & 41/150 & 27.3\% & 4/15 & 26.7\% $[10.9,\,52.0]$ \\
DeepSeek-V4-Pro [mini-swe-agent] & 15 & 25/135 & 18.5\% & 3/15 & 20.0\% $[7.0,\,45.2]$ \\
Qwen3.7-Max [mini-swe-agent] & 15 & 18/135 & 13.3\% & 3/15 & 20.0\% $[7.0,\,45.2]$ \\
Nemotron-3-Ultra [mini-swe-agent] & 7 & 10/70 & 14.3\% & 1/7 & 14.3\% $[2.6,\,51.3]$ \\
GLM-5.2 [mini-swe-agent] & 13 & 20/117 & 17.1\% & 1/13 & 7.7\% $[1.4,\,33.3]$ \\
GPT-5.5 [Codex CLI] & 13 & 17/130 & 13.1\% & 1/13 & 7.7\% $[1.4,\,33.3]$ \\
GPT-5.5 [mini-swe-agent] & 13 & 14/117 & 12.0\% & 1/13 & 7.7\% $[1.4,\,33.3]$ \\
Claude Opus 4.8 [Claude Code] & 15 & 11/150 & 7.3\% & 1/15 & 6.7\% $[1.2,\,29.8]$ \\
Kimi-K2.7-Code [mini-swe-agent] & 13 & 21/117 & 17.9\% & 0/13 & 0.0\% $[0.0,\,22.8]$ \\
MiniMax-M3 [mini-swe-agent] & 5 & 3/50 & 6.0\% & 0/5 & 0.0\% $[0.0,\,43.4]$ \\
Gemini-3.1-Pro [mini-swe-agent] & 10 & 2/48 & 4.2\% & 0/10 & 0.0\% $[0.0,\,27.8]$ \\
\midrule
\textbf{All participants} & \textbf{134} & -- & -- & \textbf{15/134} & \textbf{11.2\% $[6.9,\,17.6]$} \\
\bottomrule
\end{tabular}

%% file: sections/conclusion.tex
\section{Conclusion}

We present \framework{}, an arena in which coding agents test each other. 
In a 1,500-turn tournament among 11 participants, GPT-5.5 with Codex CLI was the strongest solver, repairing 98.3\% of the defects it reviewed. Claude Opus 4.8 with mini-swe-agent was the strongest task author, with a 27.3\% Red win rate. A new participant joined with only five matches without changing the existing order. Within one model family, tasks written by the newest version defeated a fixed solver far more often than those of older versions, which gives early evidence that task difficulty rises with the author's ability.



%% file: sections/ethics.tex
\newpage
\section*{AI Use Statement}

In this work, we used generative AI to generate tasks, and attempt solutions to the tasks 
for evaluation of AI coding agents. 
We have not used generative AI tools to conceptualize our framework, propose or refine hypotheses, 
design research methodology or experiments, implement methods, 
clean and reformat dataset, support qualitative and thematic data analysis, or interpret results.
Using generative AI tools for formulating mathematical claims, 
providing critical ingredients for proving mathematical claims, 
assisting in the writing of proofs, or assisting with translation
are not applicable to this work.
We used generative AI tools for 
create and polish figures or images, 
suggest experimental parameters, 
create and edit software code, and
edit and format the paper to improve readability.
AI-generated code was manually verified in pair programming sessions by the first and second authors,
and was further verified using unit and integration test cases.  
We take responsibility for the final content of this work,
including text, claims or artifacts produced with the aid of generative AI.

\section*{Ethics Statement}
\framework{} is deliberately capable of generating subtle defects in plausible
changes to real open-source code, selecting defects that evade the repository's
existing CI and admitting them only when a frontier reviewer can nevertheless
repair them under blind review. It is therefore an automated generator of
CI-evading defects, not merely a conventional code-generation benchmark, and
its prompts, trajectories, sealed tests, and admitted challenge bank are
dual-use artifacts. The observed 27.3\% attack success rate of the strongest
Red participant against frontier-model reviewers is consequently a
disclosure-relevant security finding: it demonstrates that apparently plausible,
test-passing contributions can retain defects despite automated review, rather
than establishing an exploit rate for deployed software.


All generation and evaluation in this study ran in isolated local containers at
pinned repository revisions. We did not target deployed services, contact
maintainers under false pretenses, open pull requests, or submit any generated
feature, defect, test, or patch to a real repository. If future use of the
pipeline reveals an independently existing vulnerability rather than an
introduced benchmark defect, we require coordinated disclosure to the affected
maintainer before publication. These safeguards do not eliminate the dual-use
risk, and we will revisit access decisions if evidence suggests that release
would materially increase practical misuse.

\section*{Reproducibility Statement}

We have open-sourced the \framework{} toolkit~\citep{toolkit_swe-duel_2026}
and the experiment dataset~\citep{dataset_swe-duel_2026}
for reproducibility.




%% file: sections/appendix.tex

\definecolor{appendixcodebg}{HTML}{F7F9FC}
\definecolor{appendixcoderule}{HTML}{CAD3DE}
\definecolor{appendixlinenumber}{HTML}{7B8794}
\definecolor{appendixkeyword}{HTML}{1F4E79}
\definecolor{appendixcomment}{HTML}{5B6573}
\definecolor{appendixstring}{HTML}{9A4D00}
\definecolor{appendixadded}{HTML}{147D4D}
\definecolor{appendixremoved}{HTML}{B54747}
\definecolor{appendixhunk}{HTML}{6B4E9B}
\definecolor{appendixpromptbg}{HTML}{F5F8FC}
\definecolor{appendixpromptrule}{HTML}{9EB4D0}

\lstdefinestyle{artifact}{%
  basicstyle=\ttfamily\scriptsize,
  backgroundcolor=\color{appendixcodebg},
  frame=single,
  rulecolor=\color{appendixcoderule},
  framerule=0.4pt,
  framesep=4pt,
  numbers=left,
  numberstyle=\tiny\color{appendixlinenumber},
  numbersep=5pt,
  xleftmargin=2.25em,
  framexleftmargin=2.25em,
  breaklines=true,
  breakatwhitespace=false,
  breakindent=1.5em,
  postbreak=\mbox{\textcolor{appendixlinenumber}{\tiny$\hookrightarrow$}\space},
  columns=fullflexible,
  keepspaces=true,
  showstringspaces=false,
  showtabs=false,
  tabsize=2,
  numberblanklines=false,
  keywordstyle=\color{appendixkeyword}\bfseries,
  commentstyle=\color{appendixcomment}\itshape,
  stringstyle=\color{appendixstring},
  aboveskip=10pt,
  belowskip=7pt
}

\lstdefinelanguage{appendixdiff}{%
  morecomment=[f][\color{appendixremoved}]{-},
  morecomment=[f][\color{appendixadded}]{+},
  morecomment=[f][\color{appendixhunk}]{@@}
}

\makeatletter
\newcommand{\diffnumber}[1]{%
  \ifnum\c@lstnumber>\z@ \tiny\color{appendixlinenumber}#1\fi}
\newcommand{\diffhunkstart}[1]{\global\c@lstnumber=\numexpr(#1)-1\relax}
\newcommand{\difffilebreak}{\global\c@lstnumber=-3\relax}
\makeatother

\lstdefinestyle{artifactdiff}{%
  style=artifact,
  escapechar=`,
  numberstyle=\diffnumber,
  firstnumber=-2
}

\makeatletter
\renewcommand*\theHlstnumber{%
  \ifx\lst@@caption\@empty \lst@neglisting \else \theHlstlisting \fi
  .\thelstnumber.\the\lst@lineno}
\makeatother

\lstdefinelanguage{appendixjavascript}{%
  sensitive=true,
  morekeywords={const,let,var,function,return,if,else,for,while,new,typeof,
    undefined,async,await,true,false,null},
  morecomment=[l]{//},
  morecomment=[s]{/*}{*/},
  morestring=[b]',
  morestring=[b]"
}

\newcommand{\listingheading}[1]{\textbf{#1}\mbox{}\par}

\section{Task Freshness and Task Bank Release Policy}
\label{app:freshness-policy}
The same task cannot remain both public and unseen. In \framework{}, a generated
bank is static within a tournament so every opponent faces controlled, reusable
tasks;
freshness is renewed across tournaments by generating a new task bank. 
Full publication of replicated results would enable exact external replay,
but it would both make the tasks unsuitable for future unseen-task claims. 
We encourage the generating of new task bank for each replication study 
to limit the possibility of the released tasks from previous replication 
studies leaking into the training data of newer LLMs.
Any disclosed task bank should be marked \emph{retired} 
and cannot support evaluation of models released in the future, and
other unseen-task-related works. 

\input{sections/worked_challenges}

\section{Repository Field}
\label{app:repository-field}

Table~\ref{app:repository-field} shows the 15 repositories across 5 languages used in \framework{} along with exact commit number for each of the repositories.

\begin{table*}[h]
\centering
\caption{The fixed repository field. 
Commit is the 12-hex prefix of the checked-out SHA. 
SLoC gives a rough estimate on the size/complexity of the repository via
the number of nonblank lines in Git-tracked files of the repository's
implementation language, 
including tests but excluding documentation and build outputs. 
Tests is the number of test cases in existing-suite;
$\dagger$ denotes configured executable targets when the runner does
not expose case totals. Time is the median wall-clock time of those existing-suite
runs; commands that compile include compilation time.}
\label{tab:repositories}
\scriptsize
\setlength{\tabcolsep}{4pt}
\resizebox{\columnwidth}{!}{%
\begin{tabular}{llrrrr}
\toprule
\textbf{Repository} & \textbf{Commit} & \textbf{Language} & \textbf{SLoC} & \textbf{Tests} & \textbf{Time (s)} \\
\midrule
Flask & \texttt{7fff56f5172c} & Python & 13,751 & 480 & 3.7 \\
Jinja & \texttt{15206881c006} & Python & 18,903 & 909 & 4.2 \\
SQLAlchemy & \texttt{5237d923ddb0} & Python & 511,172 & 18,020 & 161.8 \\
chi & \texttt{3b171578ca44} & Go & 9,923 & 254 & 36.1 \\
golang-jwt/jwt & \texttt{7ceae619e739} & Go & 5,503 & 201 & 5.6 \\
gorilla/csrf & \texttt{9dd6af1f6d30} & Go & 2,760 & 50 & 3.9 \\
Helmet & \texttt{638e43becc81} & Node & 3,373 & 115 & 10.0 \\
Express & \texttt{dbac741a49a5} & Node & 17,339 & 1,130 & 6.3 \\
node-jsonwebtoken & \texttt{ed59e76ea37a} & Node & 3,949 & 512 & 3.4 \\
OWASP Java HTML Sanitizer & \texttt{4e186b8a8c84} & Java & 22,492 & 332 & 82.4 \\
java-jwt & \texttt{695fd2bea64b} & Java & 11,976 & $1\dagger$ & 81.5 \\
JJWT & \texttt{0df975627c2c} & Java & 66,626 & 20 & 127.6 \\
cJSON & \texttt{c859b25da029} & C & 18,999 & $1\dagger$ & 0.7 \\
libexpat & \texttt{c61098da494e} & C & 36,207 & $1\dagger$ & 22.6 \\
simdjson & \texttt{1bcf71bd8505} & C++ & 313,998 & $4\dagger$ & 59.0 \\
\bottomrule
\end{tabular}
}
\end{table*}

\section{Additional Experimental Detail}
\label{app:details}

\subsection{Harness Setup}
\label{app:harness-access}

All participants receives an initial step budget of 50 harness steps.
The 50-step initial ceiling applies separately to Red feature generation, 
Red bug embedding, blind self-review, and each Blue task-solving sessions. 
Each of the sessions are also subject to a 30-minute wall-clock limit.
If required artifacts remain missing,
up to three recovery turns of six steps each is given to the participant.
In this work, we choose mini-swe-agent~\citep{yang_swe-agent_2024},
Codex CLI~\citep{openai_non-interactive_2026}, and 
Claude Code~\citep{anthropic_run_2026}, all of which are popular harnesses 
that post strong performances across various coding benchmarks.
Note that different models use different amount of compute resources, 
especially depending on the accompanying harness.
Thus, the harness step counts do not represent equal compute. 
Appendix~\ref{sec:resource-budgets} reports a token-level 
compute-controlled analysis.
All the agents, validation gates, and turn scorers run in separate 
Docker container instances of the same
repository-specific image to ensure a consistent environment.
The inference temperature for all the models is fixed at $0.2$.


\subsection{Model API Access Dates}
\label{app:model-access}

All participants access their models through the OpenRouter API\footnote{\url{https://openrouter.ai}}.
Table~\ref{tab:api-access} lists the first and last recorded API access date for each model.
The windows span task creation (including failed attempts and blind self-review validation), evaluation in the main and harness-ablation tournaments, and the post-hoc studies of Appendices~\ref{app:admission-noise}, \ref{app:reference-reviewer}, and~\ref{app:capability-scaling}: the adaptive task difficulty study adds the GLM-5.1 and GLM-5.3 windows (adaptive task creation), and extends the DeepSeek V4 Pro window (fixed-reference task solver for the adaptive tasks).
They are identical across the harnesses paired with a model because all harnesses route to the same OpenRouter endpoint; all dates are in UTC.

\begin{table}
\centering
\caption{Per-model API access dates. Each entry gives the first and last
recorded model call through the OpenRouter API (UTC). All calls of a model use
the same OpenRouter model identifier regardless of the paired harness.}
\label{tab:api-access}
\small
\setlength{\tabcolsep}{4pt}
\resizebox{\columnwidth}{!}{%
\begin{tabular}{llrr}
\toprule
\textbf{Model} & \textbf{OpenRouter model ID} & \textbf{First access} & \textbf{Last access} \\
\midrule
GPT-5.5 & \texttt{openai/gpt-5.5} & 2026-06-18 & 2026-08-29 \\
Claude Opus 4.8 & \texttt{anthropic/claude-opus-4.8} & 2026-06-18 & 2026-08-29 \\
Kimi K2.7 Code & \texttt{moonshotai/kimi-k2.7-code} & 2026-06-18 & 2026-08-29 \\
DeepSeek V4 Pro & \texttt{deepseek/deepseek-v4-pro} & 2026-06-18 & 2026-09-19 \\
GLM 5.1 & \texttt{z-ai/glm-5.1} & 2026-09-18 & 2026-09-18 \\
GLM 5.2 & \texttt{z-ai/glm-5.2} & 2026-06-18 & 2026-08-30 \\
GLM 5.3 & \texttt{z-ai/glm-5.3} & 2026-09-18 & 2026-09-18 \\
Qwen 3.7 Max & \texttt{qwen/qwen3.7-max} & 2026-07-11 & 2026-08-29 \\
Nemotron 3 Ultra & \texttt{nvidia/nemotron-3-ultra-550b-a55b} & 2026-07-11 & 2026-08-29 \\
MiniMax M3 & \texttt{minimax/minimax-m3} & 2026-06-18 & 2026-08-29 \\
Gemini 3.1 Pro & \texttt{google/gemini-3.1-pro-preview} & 2026-06-19 & 2026-08-29 \\
\bottomrule
\end{tabular}%
}
\end{table}



\section{Evaluation and Outcome Classification}
\label{app:match-execution}


Algorithm~\ref{alg:match-execution} describes the end-to-end procedure 
encompassing task cache lookup, and turn-based match scoring. 
The $\operatorname{Load}(r,R_j,i)$ procedure returns the $i$th admitted task for Red
participant $r$ in bank-index order. 
After Blue submits a diff, the existing test suite, and the test suites for feature, and embedded bugs executes. 
Existing test suite failure classifies the attempt as a regression. 
If existing tests pass but the feature suite fails, 
the attempt is a feature collapse. 
If both pass but hidden bug tests fail, the defect shipped. 
Blue wins if and only if all the test suites pass. 

\begin{algorithm*}[t]
\caption{Deterministic execution and scoring of a symmetric match}
\label{alg:match-execution}
\begin{algorithmic}[1]
\Require Participants $A,B$; repositories $\mathcal{R}$; target count $N$;
task bank $\mathcal{B}$; solving cache $\mathcal{D}$
\Ensure Match totals $(Q_A,Q_B)$ and outcome $O$
\State $Q_A \gets 0$; $Q_B \gets 0$; $T \gets [\,]$
\ForAll{$R_j \in \mathcal{R}$}
    \For{$i \gets 1$ \textbf{to} $N$}
        \ForAll{$(r,b) \in [(A,B),(B,A)]$} \Comment{swap Red and Blue roles}
            \State $C \gets \operatorname{Load}(\mathcal{B},r,R_j,i)$
            \If{$C=\varnothing$}
                \State $(S_{\mathrm{Blue}},S_{\mathrm{Red}}) \gets (1,0)$
                \Comment{automatic Blue win for missing Red task}
            \Else
                \If{$\mathcal{D}$ contains $(C,b)$}
                    \State $F \gets \mathcal{D}[C,b]$ \Comment{reuse immutable solving run}
                \Else
                    \State $P \gets \Call{PublicBundle}{C}$
                    \State $F \gets \Call{Solve}{b,P}$
                    \Comment{sealed defect metadata and bug tests remain hidden}
                    \State $\mathcal{D}[C,b] \gets F$
                \EndIf
                \State $S_e \gets \mathbb{1}[\Call{ExistingTests}{C,F}\ \mathrm{pass}]$
                \State $S_f \gets \mathbb{1}[\Call{FeatureTests}{C,F}\ \mathrm{pass}]$
                \State $S_b \gets \mathbb{1}[\Call{HiddenBugTests}{C,F}\ \mathrm{pass}]$
                \State $S_{\mathrm{Blue}} \gets S_e S_f S_b$;
                $S_{\mathrm{Red}} \gets 1-S_{\mathrm{Blue}}$
            \EndIf
            \State $Q_b \gets Q_b+S_{\mathrm{Blue}}$;
            $Q_r \gets Q_r+S_{\mathrm{Red}}$
            \State \Call{Append}{$T,(R_j,i,r,b,C,S_{\mathrm{Red}},S_{\mathrm{Blue}})$}
        \EndFor
    \EndFor
\EndFor
\If{$Q_A>Q_B$}
    \State $O \gets A\text{-wins}$
\ElsIf{$Q_B>Q_A$}
    \State $O \gets B\text{-wins}$
\Else
    \State $O \gets \mathrm{draw}$
\EndIf
\State \Return $(Q_A,Q_B,O,T)$
\end{algorithmic}
\end{algorithm*}

\section{Rank Trajectories}
\label{app:rank-trajectory}

\begin{figure}[h]
\centering
\includegraphics[width=0.75\columnwidth]{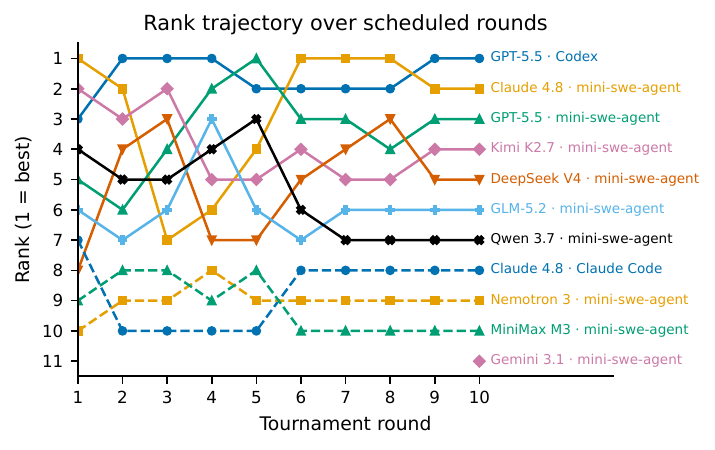}
\caption{BT rank recomputed after each scheduled tournament round. A late
entrant begins only when it first plays. This arrival-order diagnostic is not a
ranking rule or convergence criterion.}
\label{fig:rank-trajectory}
\end{figure}

Figure~\ref{fig:rank-trajectory} re-fits BT using only the matches available at
each scheduled round. It documents when observations arrived, rather than a
second estimand: the reported leaderboard fits the complete persisted schedule
in Table~\ref{tab:match-leaderboard}.

\input{sections/oracle_audit}
\section{Main Tournament Bradley--Terry Confidence Intervals}

Table~\ref{tab:bt-uncertainty} shows BT intervals and central 95\% rank 
intervals use the 1,000-draw match-bootstrap samples.
No two adjacent participants in the ranking ordering is distinguishable at 95\%. 
Such overlapping confidence intervals among nearby participants are common in head-to-head arenas~\citep{arenaai_llm_2026,designarenaai_design_2026}. 

\begin{center}
\centering
\captionof{table}{
BT point estimate and ranking confidence intervals corresponding to Figure~\ref{fig:headline-elo}.
}
\label{tab:bt-uncertainty}
\setlength{\tabcolsep}{5pt}
\resizebox{.7\textwidth}{!}{
\input{Tables/tab_bt_uncertainty}
}
\end{center}

\section{Session-C Admission Stability}
\label{app:admission-noise}

Admission requires that a fresh blind instance of the same model--harness
repair the embedded defect (Section~\ref{sec:session-review}). 
All models' inference is configured with temperature $0.2$, 
so the reviewer's trajectory (and hence its repair) is a stochastic draw 
rather than a deterministic property. 
A task that no fresh reviewer repairs 
(for e.g. fails all $k$ independent re-runs) is therefore  
a direct signature of a \emph{lucky admission}. 
Thus, we re-run the identical Session-C probe 
using the same model--harness pair, prompt template, public test bundles, 
sealed bug-test oracles 
and other harness settings such as recovery loop, step and time budgets.
The probe runs for $k{=}3$ additional times per task on a stratified 
sample of the admitted tasks the headline tournament actually solved: 
four tasks per participant, 
round-robined across repositories, 
drawn from the $134$ tournament-contested tasks by a per-participant 
seeded random generator.
Each of the $44\times 3 = 132$ runs launches a fresh reviewer instance in an
isolated workspace and persists its verdict, trajectory, and bug-test output.

\begin{table*}[h]
\centering
\caption{Session-C admission-noise stability study over the headline tournament
bank. For each participant, its four sampled tournament tasks
distribute over the repaired-$x/3$ buckets according to how many of the
$k{=}3$ fresh blind re-runs of the exact Session-C probe repaired them;
``Repairs/runs'' pools the $44\times 3=132$ per-run verdicts with Wilson
95\% intervals. 
}
\label{tab:admission-noise}
\small
\resizebox{\linewidth}{!}{\input{Tables/tab_admission_noise}}
\end{table*}

\begin{figure}[h]
\centering
\includegraphics[width=0.70\columnwidth]{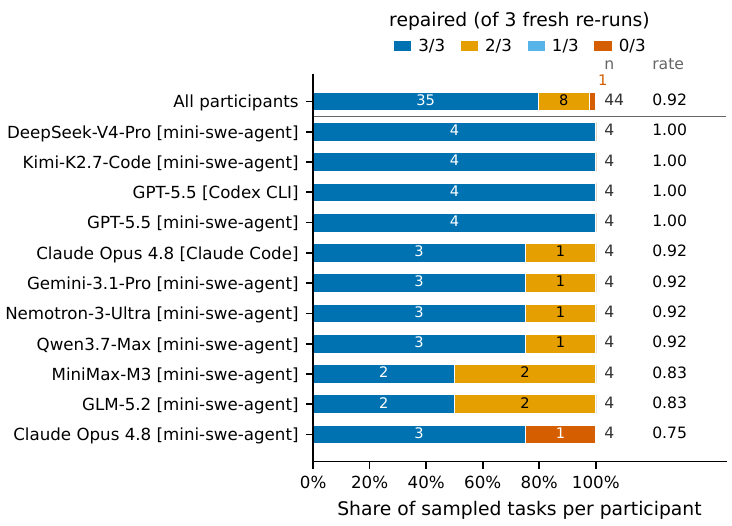}
\caption{Session-C admission-noise stability study. Each row stacks, as a
share of its sampled tasks, how many of the three fresh blind re-runs
repaired them (segment labels give absolute counts; $n$ at each bar tip
gives the row's sample size); the right column reports the pooled per-run
repair rate, and the top row aggregates all $44$ tasks. The
distribution concentrates on reliable solvability ($3/3$), with a single
lucky admission ($0/3$).}
\label{fig:admission-noise}
\end{figure}

\textbf{Result.}
The distribution demonstrates the admitted tasks' reliable solvability: 
$35$ of $44$ tasks ($79.5\%$) are repaired $3/3$,
$8$ ($18.2\%$) are repaired $2/3$, none $1/3$, and a single task
($2.3\%$) is repaired $0/3$---the study's one lucky admission. The pooled
per-run repair rate is $121/132 = 91.7\%$ (Wilson 95\% CI
$[85.7\%, 95.3\%]$), and the average per-run repair probability
over the tournament bank is approximately $0.92$.
Thus, admission mostly certifies reliable solvability rather than a lucky draw. 
Table~\ref{tab:admission-noise} and Figure~\ref{fig:admission-noise} 
report the per-participant breakdown.
``Repaired $x/3$'' counts how many of the fresh blind reviewers repaired the
task. $3/3$ tasks are reliably solvable, so admission was not a
fluke; $2/3$ and $1/3$ tasks sit near the reviewer's ability boundary,
where repair depends on sampling luck; $0/3$ tasks are lucky admissions.
per-participant pooled rates range from $75.0\%$ to $100.0\%$, and all nine
tasks outside the $3/3$ bucket embed logic or off-by-one defects.

\textbf{Insights.}
Since the pooled per-run repair rate estimates
the degree to which admission implies reliable solvability,
the prevalence of tasks that are successfully admitted with $3/3$ Session-C re-runs 
shows that admission verdict is a stable property of the admitted tasks.
Meanwhile, the $\{1/3, 0/3\}$ tail represents the noise 
floor of the admission filter and flags that tournament turns against 
those tasks are high-variance events even for a capable Blue.
Overall, $79.5\%$ of the sampled tasks are reliably repaired $3/3$, 
the $18.2\%$ boundary mass at $2/3$ bounds the share of
contested turns whose solvability is partly sampling luck, 
and the lucky-admission tail is a single task. 
Admission noise therefore does not materially 
attenuate the reported Blue success rates.

\section{Fixed-reference Task Solver}
\label{app:reference-reviewer}

In the main tournament, a participant's Red success mixes 
the skill of its task author with 
(i)~how difficult its own admitted tasks are and 
(ii)~which solvers its tasks happened to meet. 
Thus, the match record alone cannot separate these factors.
We therefore designate one mid-tier participant as a fixed reference
solver: GLM-5.2 with mini-swe-agent, whose contested Blue success of
$.874$ over its $111$ in-frame tournament solving runs matches the field
aggregate of $.851$.
We have the fixed-reference solver solve everyone of $134$ distinct admitted task in
the main tournament blind through the exact task-solving evaluation
path of a match sub-turn: same Blue prompt, step and wall-clock budgets, and
regression and feature test suites, and sealed bugfix scoring.
Thus, 
the task difficulty reading is the same measuring stick for everyone,
and is exogenous to the task creation process.
Comparing it with the endogenous rate from the main round-robin tournament shows
how much of a participant's scheduled Red success transfer to a neutral solver.

Since the reference solver had already solved $111$ of the $134$ tasks in-match,
those turn results are reused for this fixed-reference study rather than re-run. 
Thus, the $23$ tasks remained uncovered by the round-robin main tournament, 
out of which $13$ were authored by the reference itself 
(admission self-review records are deliberately \emph{not} reused for the
reference-authored tasks since 
those records are the selection event rather than a
measurement),
and $10$ stem from the one pairing involving the reference and the late entrant to the tournament (Gemini 3.1 Pro with mini-swe-agent) that was not part of the active-sampled match-ups in the $50$-match main tournament schedule.
These $23$ get fresh blind runs. 

\begin{figure}[t]
\centering
\includegraphics[width=0.6\columnwidth]{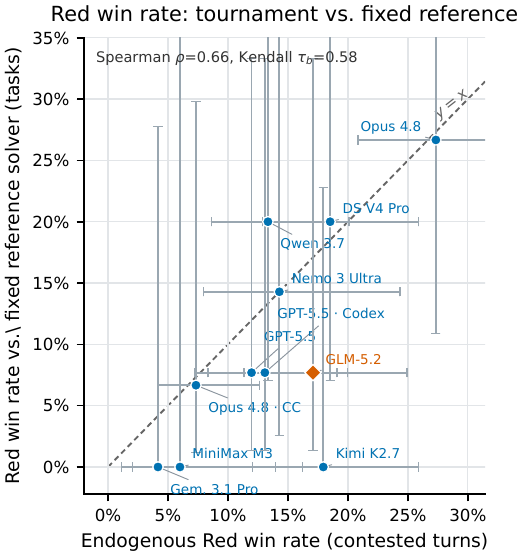}
\caption{Task difficulty: scheduled field vs.\ fixed reference. Each point
is one participant's Red win rate against the full scheduled solver field
(x-axis) and against the fixed reference solver (y-axis). 
Error bars are Wilson 95\% intervals.
Points without a harness tag use mini-swe-agent. 
}
\label{fig:reference-attack}
\end{figure}

\textbf{Result.}
The fixed solver fails to solve $15$ of the
$134$ tasks ($11.2\%$, Wilson 95\% $[6.9, 17.6]$), and across the eleven
participants the two difficulty scales agree at Spearman $\rho = 0.66$
(Kendall $\tau_b = 0.58$). 
The strongest task creator of the main tournament keeps the top
slot on the fixed-reference scale (Claude Opus 4.8 with mini-swe-agent:
$27.3\%$ endogenous, $26.7\%$ reference). 
Nemotron-3-Ultra stays exactly at $14.3\%$ on both scales.
The weakest task generators stay at the bottom (Gemini 3.1 Pro and MiniMax-M3 
cause zero fixed-reference solver failures in $10$ and $5$ tasks, respectively). 
Kimi-K2.7-Code wins $17.9\%$ of its Red turns against
the other participants, yet wins $0\%$ against the fixed-reference solver, 
so its success exploits weaknesses of the other participants rather 
than its intrinsic difficulty; conversely, Qwen3.7-Max ($13.3\%\to20.0\%$) and
DeepSeek-V4-Pro ($18.5\%\to20.0\%$) author tasks that are \emph{harder}
for the fixed-reference solver than the other participants. 
Table~\ref{tab:reference-attack} reports the per-participant comparison and
Figure~\ref{fig:reference-attack} the scatter plot visualization of it.

\textbf{Insights.}
The moderate-to-strong positive rank agreement between the main round-robin 
tournament and fixed-reference evaluation 
(Spearman $\rho = 0.66$, and Kendall $\tau_b=0.58$) 
means the endogenous difficulty ordering is substantially a
property of the \emph{generated tasks} rather than the resultant of
which solvers a task author happened to meet.
One caveat: with $5$--$15$ tasks per participant, the confidence intervals are wide. 
Our future extension plans more in-depth study of the finer-grained ordering
that supports rank-level agreement claims. 

\section{Scaling Task Difficulty with Task Generator Capability}
\label{app:capability-scaling}

While the fixed reference solver of Appendix~\ref{app:reference-reviewer} 
removes the solver-field confound,  it still compares the participants 
of different origin that happen to be in the field.
Thus, their task difficulties are a property of the generated tasks, 
not that it changing with task authoring capability. 
We therefore hold the solver fixed and move the \emph{task author} 
along a known capability axis. 
Successive versions of one model family are exactly such an axis, 
so we built a three-rung ladder from GLM-5.1 $\to$
GLM-5.2 $\to$ GLM-5.3, all running the same mini-swe-agent harness.
Each version authored tasks over the same repository field under the identical
task generation protocol and validation gates. 
GLM-5.1 and GLM-5.3 wrote fresh banks ($12$ and $14$ admitted tasks, respectively);
the $13$ GLM-5.2 tasks are its admitted tasks from the main tournament.
We use DeepSeek-V4-Pro with mini-swe-agent as the fixed reference solver for 
this study, as it is another mid-tier solver whose contested Blue success of 
$.919$ sits near the field aggregate of $.851$, 
and a model outside the ladder family.
The fixed-reference solver than then blind-solved every one
of the $39$ ladder tasks through the exact task-solving path of a match sub-turn:
same Blue prompt, step and wall-clock budgets, regression and feature test suites, and sealed bugfix oracle. 
Since the fixed-reference solver had already solved the $13$ GLM-5.2 
tasks in-match during the main round-robin tournament, 
those turn records are reused as part of this study. 

\textbf{Result.}
Solver failure rises with the capability frontier
(Table~\ref{tab:capability-scaling}): $1/12$ ($8.3\%$, Wilson 95\%
$[1.5,35.4]$) for GLM-5.1, $0/13$ for GLM-5.2, and $6/14$ ($42.9\%$,
$[21.4,67.4]$) for GLM-5.3. The upward trend across the ladder is
significant (exact conditional Cochran--Armitage trend test over the $39$
draws, $z{=}2.32$, one-sided $p{=}.015$), and the newest version is clearly
separated from the pooled older ones ($6/14$ vs.\ $1/25$, Fisher
$p{=}.005$), while the two older versions are indistinguishable from each
other. All seven failures are complete collapses (Blue composite $0.00$),
and the solver works \emph{harder} on the tasks it loses (mean $37.3$
vs.\ $31.5$ steps), so failures are exhausted budgets ending in broken
behavior rather than early give-ups. 


\textbf{Composition vs. difficulty.}
The defect-category mix shifts along the ladder (security vulnerabilities
$4\to2\to7$; logic errors $2\to9\to2$), and GLM-5.2's bank is dominated by
logic errors that the solver always repaired. Composition alone, however,
does not explain the jump: within security-vulnerability tasks only, the
older versions' tasks never defeat the solver ($0/6$) while GLM-5.3's do in
$5/7$ (Fisher $p{=}.016$) even though that category was among the most
solvable for the full field in main tournament ($84.9\%$ Blue success).
Admission yield does not explain it either: per-attempt yields are $12/39$
($30.8\%$), $13/32$ ($40.6\%$), and $14/37$ ($37.8\%$), non-monotonic across
the ladder, with the difficulty jump occurring where yield slightly
\emph{falls}. 
Thus, difficulty rises alongside rather than because of 
other task generation statistics.

\textbf{Head-to-head tournament among the ladder versions.}
The fixed solver sits outside the family, so it cannot say how the three
versions fare against one another. Thus, we conducted a
complete round-robin tournament where 
every version plays Red and Blue on every repository once per pairing
(Figure~\ref{fig:capability-h2h}). 
The head-to-head tournament produces endogenous match-ups which correlated with 
the ladder. 
GLM-5.3's tasks defeat GLM-5.1 in $7/14$ contested turns and GLM-5.2 in $4/14$, 
while the two older versions almost never defeat anyone 
($3/50$ pooled across their four cells; 
one-sided exact Fisher against GLM-5.3's $11/28$: $p{=}.0004$). 
The newest version is also the strongest solver, 
conceding $2/25$ contested turns to its siblings' tasks versus $4/26$ 
for GLM-5.2 and $8/27$ for GLM-5.1; it won both of its matches 
($7$--$1$ and $4$--$1$ in Red wins) and GLM-5.2 won the third
($1$--$0$), a clean $2$--$1$--$0$ sweep of the standings. 
The sibling readings agree with the fixed-reference solver where they overlap: 
the outside solver failed on $6/14$ of GLM-5.3's tasks, 
the siblings on $4/14$ and $7/14$, and on the two older versions' tasks 
no solver in either study concedes more than a single turn. 
Siblings share a training lineage and may share blind spots, so
the matrix corroborates rather than replaces the fixed-solver measurement.


\textbf{Insights.}
With solver, harness, repositories, and protocol held fixed, tasks authored
by the newest version of the family are an order of magnitude harder for the
same solver than the pooled older versions' tasks ($42.9\%$ vs.\ $4.0\%$
failure).
Together with the trend test, this is direct evidence for the 
self-scaling property of \framework{} design: 
as the authoring agents improve, the tasks they produce get harder
for a fixed solver. 
One fixed-reference solver (DeepSeek V4 Pro with mini-swe-agent) 
is the single measuring stick just like Appendix~\ref{app:reference-reviewer}
where the rank agreement of $\rho{=}.66$ suggests
the reading transfers to other mid-tier solvers, and the within-family
round-robin (Figure~\ref{fig:capability-h2h}) supplies two sibling solvers
whose failure rates climb the same way, 
but siblings share the family's training lineage, 
so they corroborate rather than independently re-measure the ladder. 
The ladder has three rungs from one model family; 
such ladders are cheap to extend (the fresh rungs here cost \$15.27 total), 
so the task difficulty scaling trend can be continuously verified as new versions ship.
At $12$--$14$ tasks per level and one draw per task at temperature $0.2$, 
adjacent-version resolution is limited
(the admission stability study in Appendix~\ref{app:admission-noise} 
measured a $.917$ per-run repair rate, and the two older levels' 
intervals overlap almost entirely),
so even though the task difficulty scaling is not strictly per-version monotone, 
the data still supports it roughly scaling with the capability frontier.

\section{Role Decomposition and Contested-only BT Refits}
\label{app:estimand-decomposition}

\begin{figure*}[h]
\centering
\includegraphics[width=0.9\textwidth]{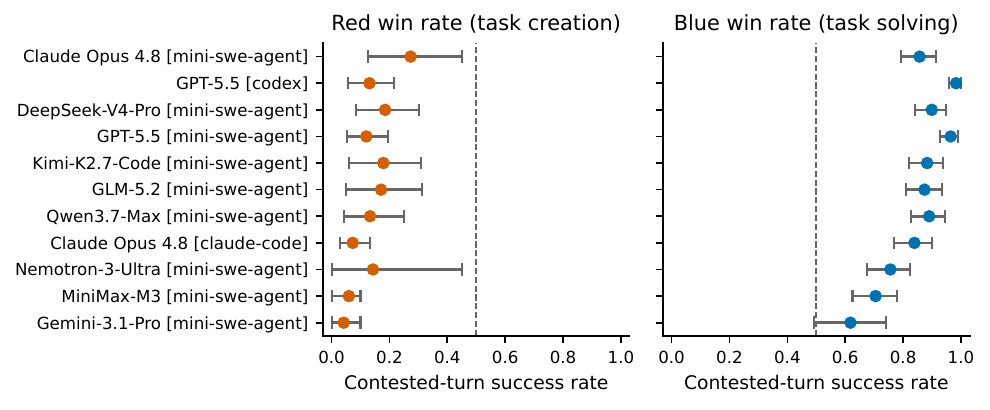}
\caption{Task-clustered contested task-creation and task-solving success by
competitor. 
Missing-Red auto-wins are excluded.}
\label{fig:role-rates}
\end{figure*}

\textbf{Role Decomposition.} 
Table~\ref{tab:decomposition} separates the two capabilities that the
end-to-end match score intentionally combines: 
Red task generation capability, and Blue task solving capability. 
Thus, we analyze the task solving-only BT fit of the main round-robin tournament.
Coverage is the number of admitted tasks divided by the 15 scheduled 
repository slots; an unsuccessful chain of $5$ attempts counts as one 
exhausted slot. 
The task-solving intervals are task-clustered bootstrap intervals.
Figure~\ref{fig:role-rates} decomposes each system into task creation and task
solving.
Nemotron-3-Ultra's contested Red win rate is $10/70$, 
but nine of the ten wins come from one of its seven admitted tasks. 
A bootstrap draw that omits both winning clusters rates the task creator at zero 
while a draw that includes the nine-win task several times is rated near $0.5$. 
Thus, the result is the field's widest task-creation interval, $[0.000,\,0.450]$.
Appendix~\ref{app:attack-anatomy} analyzes the responsible task in detail. 

\begin{table*}[h]
\centering
\caption{
Task solving and generation-coverage decomposition for the headline tournament. 
Participants are shown in the order of full-schedule tournament's BT ratings. 
Solving BT is a role-conditioned fit on contested Blue turns only.
Coverage is admitted slots divided by the 15 repository slots. 
}
\label{tab:decomposition}
\small
\setlength{\tabcolsep}{3pt}
\resizebox{\textwidth}{!}{\input{Tables/tab_decomposition}}
\end{table*}

\textbf{On preventing saturation.}
As we have seen in this section, models are generally better at finding bugs and solving tasks than they are at authoring tasks.
Plus, the admission of a task is contingent upon the task author's task solving 
capability.
Thus, the shrinking gaps in task solving capabilities among 
frontier models could shrink the gaps in their task authoring capabilities,
potentially leading to saturation of the arena.
Two solutions (one short-term, and one long-term) address this problem:
\begin{itemize}
    \item \textbf{(Short-term, temporary)} 
    Appendix~\ref{app:repository-field} shows the 15 repositories used in our 
    experiments. 
    They are selected for being relatively small-sized projects 
    in their respective languages/domains. 
    This is because we need their test suites to be lightweight enough 
    for quick execution of validation gates for the sake of our experiments,
    and for the agent participants to not face too much difficulty 
    with generating tasks and burn the expensive LLM API resources 
    (Appendix~\ref{sec:resource-budgets}).
    Thus, practitioners adopting \framework{} may extend to cover bigger 
    repositories with more complex codebases that has wider surface area 
    for feature generation and more subtle and nuanced bug embedding.
    This would help better highlight the gap in coding capabilities of 
    different frontier models, and prevent saturation of the arena that would 
    otherwise be confined to the same set of 15 lightweight repositories.
    \item \textbf{(Long-term, permanent)}
    A future extension of this work could focus on the solvability checks
    that does not require blind self-reviewing by the task author,
    and thus is not capped by the task solving capability of the task author.
    For example, the solvability check gate could involve the privileged information
    such as informing the self-review session about the existence of the bug, 
    or to look for the specific type of the bug.
    This would still prove the bug is fair and solvable, 
    but now the task author has more leeway to embed more 
    subtle and challenging bugs that no longer needs to
    be detected and fixed by the blind self-review session.

\end{itemize}

\textbf{Contested-only refit.} 
The contested-only refit moves GPT-5.5 with mini-swe-agent from third to first,
GPT-5.5 with Codex from first to second, Claude Opus 4.8 with mini-swe-agent
from second to fifth, Nemotron-3-Ultra from ninth to third, and MiniMax-M3 from
tenth to seventh. The corresponding durable-slot coverage rates are $5/15$
($33.3$\%) for MiniMax-M3, $7/15$ ($46.7$\%) for Nemotron-3-Ultra, and $10/15$
($66.7$\%) for Gemini-3.1-Pro; the table reports all participants.
The rank moves are dominated by the task generation coverage penalty that the
full-schedule estimand imposes via automatic wins. 
In Nemotron-3-Ultra's case, with $7/15$ slots admitted, 
each of its ten matches concedes eight uncontested Blue points,
totalling 80 auto-win turns across the main tournament.
The contested-only refit deletes exactly those 80 concessions, 
and re-weights records to the higher win-rate contested-only Red wins. 
Thus, its refit coefficient ($+0.145$) is the third highest, jumping in 
ranking from ninth to third. 
The position is nonetheless fragile 
as its clustered interval $[-0.35,+0.75]$  zero at its lower bound,
and nine of its ten contested Red wins arise from a single admitted task.
Appendix~\ref{app:attack-anatomy} dissects that task, and its ten solving runs.

\begin{figure*}[t]
\centering
\includegraphics[width=\textwidth, trim={2cm 0 0 0},clip]{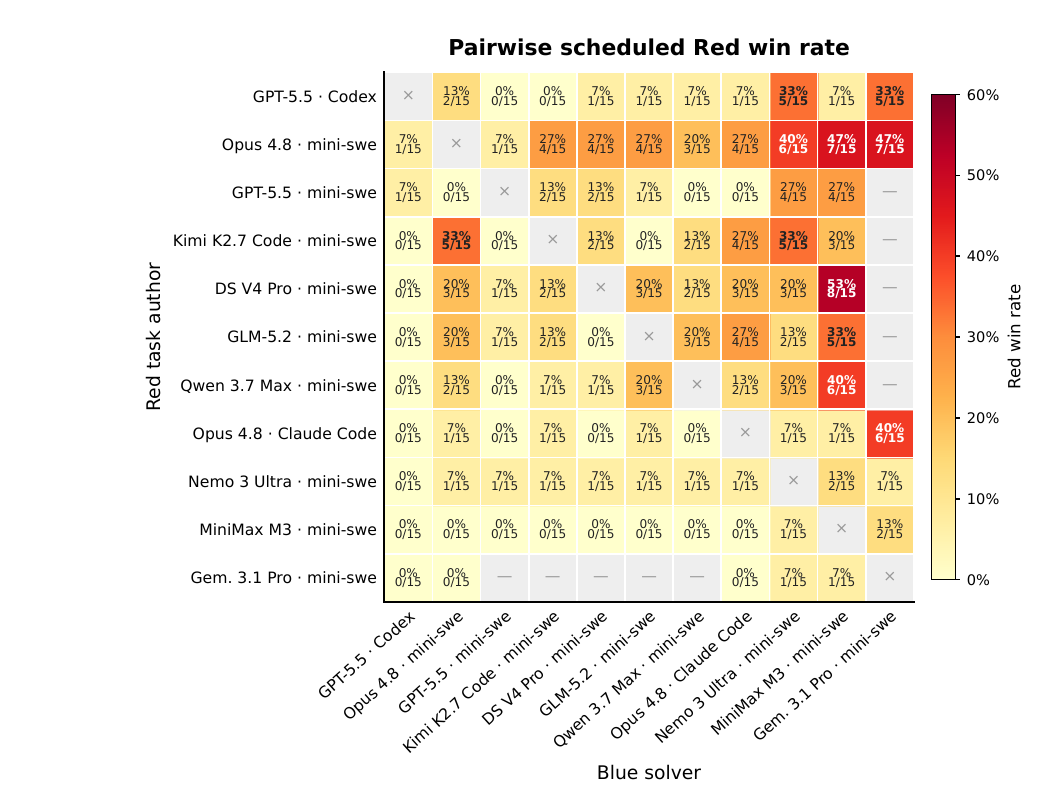}
\caption{Red--Blue transfer matrix, ordered by the tournament's Bradley--Terry
log-strength rating. Each observed cell
reports the Red win rate as a percentage and as raw wins over the 15
scheduled repository turns ($n=15$). 
}
\label{fig:h2h}
\end{figure*}

\section{Red--Blue transfer}
\label{app:red_blue_transfer}

Figure~\ref{fig:h2h} depicts the Red/Blue success rates between any two 
participants from the main round-robin tournament.
Each non-diagonal cell has one scheduled turn per repository in the
main tournament, 
so the displayed $x/15$ values should be read as raw event counts, 
not as precise estimates. 
Some cells have no turns due to the late entrant (Gemini 3.1 Pro with
mini-swe-agent) only having active-sampled matchups with a subset of
the existing participant pool.
A future Red--Blue transfer study should add a new axis by freezing 
a common task subset per task author before comparing solvers.

\section{Turn-level Bradley--Terry Analysis}
\label{app:turn-level-bt}

The match-level estimand in Section~\ref{sec:tournament} compresses each match
into a single outcome. 
This section reports the complementary turn-level estimand, 
which keeps every turn in a match as independent matches: 
each turn contributes one Red-versus-Blue Bradley--Terry outcome under the same
draw-as-half-win convention, so the BT fit measures per-turn strength rather
than per-match win aggregation. 
We fit two variants on the main round-robin tournament's $1{,}500$ persisted 
turn records ($1{,}219$ contested turns and $281$ missing-Red auto-wins.
The \emph{full-schedule} variant keeps all turns and scores an auto-win as a Blue
win.
The \emph{contested-only} variant drops the $281$ auto-win turns.

Table~\ref{tab:turn-bt} reports the point estimates and 95\% intervals
for both BT ratings, and tournament rankings.
We perform $1{,}000$ bootstrap draws to resample the generated tasks.
rather than individual turns because one task is defended by many participants, 
and those turns are correlated through the shared task. 
The auto-win turns due to empty Red--repository slots 
form a single cluster, since they are repeated
concessions of one generation failure rather than independent events. 
In the full-schedule variant, every participant's turns 
is balanced by design (fifteen Red and fifteen Blue turns per match), 
whereas contested-only turns scales with admitted bank size 
($103$--$269$ turns).
Appendix~\ref{app:estimand-decomposition} further analyzes the different
effects by automatic wins, and contested-only turns in the tournament.


\textbf{Auto-win policy, not granularity, drives estimand sensitivity.}
The full-schedule turn-level rankings is much closer to the main tournament's
full-schedule match-level ranking (Kendall's $\tau=0.782$) 
than to the contested-only ordering ($\tau=0.382$), 
with the contested-only ordering itself at $\tau=0.527$ from the match ordering. 

\textbf{Where the rankings change.} In the full-schedule turn-level BT rankings,
the ranking changes are confined to the top seven, 
whose intervals all overlap since the full-schedule main tournament: 
the match-level leader GPT-5.5 with Codex drops to fourth ($+0.18$), 
conceding $2\times 10$ auto-win turns from its $13/15$ coverage, 
while the two full-coverage strong task creators rise:
Claude Opus 4.8 with mini-swe-agent to first ($+0.30$) and
DeepSeek-V4-Pro to second ($+0.25$).
This reflects the per-turn weight on Opus's field-best $27.3$\% Red win rate. 
The bottom four positions are identical to the main tournament's full-schedule 
match ordering (Opus with Claude Code, Nemotron-3-Ultra,
MiniMax-M3, Gemini-3.1-Pro).
MiniMax-M3 ($[-0.84,-0.01]$ conceded $100$ auto-win
turns from its five admitted slots), 
and Gemini ($[-0.98,-0.12]$, conceded five slots combined with 
the field's lowest contested Blue success (Figure~\ref{fig:role-rates}). 
Nemotron-3-Ultra's $[-0.70,+0.19]$ still spans zero despite $80$ concessions, 
because the single task that concentrates its Red wins 
(Appendix~\ref{app:attack-anatomy}) keeps its contested record strong.

\textbf{Rank uncertainty is larger in turn-level BT.} 
Turn-level analysis provides a more conservative treatment of ranking 
confidence intervals than match-level. 
This is to be expected as we have observed that participants are generally 
better at solving tasks than generating difficult tasks, thus every participant
has a non-trivial number of Blue wins that moderates their BT ratings.

\begin{table}[t]
\centering
\caption{
Turn-level Bradley--Terry ratings for the main tournament.
BT values are point estimates with 95\% percentile intervals 
from 1{,}000 task-cluster bootstrap draws; 
rank intervals are central 95\% intervals over the same draws. 
The full-schedule variant treats missing-Red auto-wins as Blue wins; 
the contested-only variant drops them.
}
\label{tab:turn-bt}
\small
\setlength{\tabcolsep}{4pt}
\resizebox{\columnwidth}{!}{\input{Tables/tab_turn_bt}}
\end{table}



\section{A Concentrated Red-win Record}
\label{app:attack-anatomy}

Appendix~\ref{app:estimand-decomposition} singles out Nemotron-3-Ultra 
for having the largest contested-only rank gain and 
the widest clustered Red win rate interval. 
Nine of its ten contested Red wins arise from one admitted artifact, 
task \texttt{71a623dd} on the Go \texttt{gorilla/csrf} repository. 
This section analyzes the artifact of the task and its ten 
persisted solving runs to characterize what its difficulty measures. 

\textbf{Task details.}
The feature adds a \texttt{SkipPaths} option: request paths equal to a
configured path skip CSRF validation exactly, and paths beneath a configured
path ending in a slash skip by prefix. The sealed defect is a one-condition
change in the prefix matcher: the request path must additionally be strictly
longer than the cleaned skip path, so a request equal to the configured prefix
itself (for instance \texttt{POST /webhook/} under skip path \texttt{/webhook/})
fails validation instead of being skipped. The sealed bug test requires the
lenient reading; no visible signal does. The visible feature suite enforces
the strict behavior for the unslashed path (\texttt{/webhook} must be rejected)
and accepts only strict descendants, and the PR's own doc comment over-claims
the lenient semantics, contradicting those tests. A solver must therefore
settle a specification conflict in which every visible signal
corroborates the defective behavior.

\textbf{The ten solving runs.}
One participant solved the task (GPT-5.5 with Codex CLI) 
while the other nine lost.
Each of the 9 losses is a latent bug ship 
(existing and feature suites pass, only the sealed suite fails), 
and one of the nine also appears in the audit sample of
Appendix~\ref{app:oracle-audit} as a genuine miss (solving run \texttt{56210ac0}).
Eight of the nine explicitly inspected or reasoned about the strict-length condition; 
five filed low- or info-severity findings that framed the discrepancy as a documentation error rather than a functional one.
Only one solver (GLM-5.2) never mentions the matching condition at all. 
The single success is the one solving run that separated the two path forms 
(the unslashed path must stay protected while the prefix path itself must skip)
and added its own matcher tests; 
it too labeled the issue low-severity, but repaired it.

\textbf{The task contains a genuine bug.}
The manual audit reveals the sampled solving run is a true failure of
the participants in identifying the bug, rather than an unsound bug-test oracle.
But it does not test whether \emph{Blue} solver can locate the hidden bug, 
as most solvers located the exact condition according to the manual inspection
of the Blue agents' reasoning traces.
Consequently, Nemotron-3-Ultra's third-place contested refit is heavily influenced
by this single task for its Red win rate estimand. 
This concentration reveals an interesting observation information about 
the task author: five of its seven admitted tasks defeat no one, 
so its bank is one exceptionally well-constructed defect plus six ordinary tasks. 
Thus, the clustered intervals are high variance for 
Red capability estimand for Nemotron-3-Ultra participant.

\section{Defect Composition}
\label{app:defect-composition}

Bug types are categorized using the label given the \emph{Red} agents, 
and persisted in task metadata. 
They include logic, security, off-by-one, type-confusion, 
resource, input-validation, and race-related defects. 

\begin{minipage}{\columnwidth}
\centering
\includegraphics[width=.7\columnwidth]{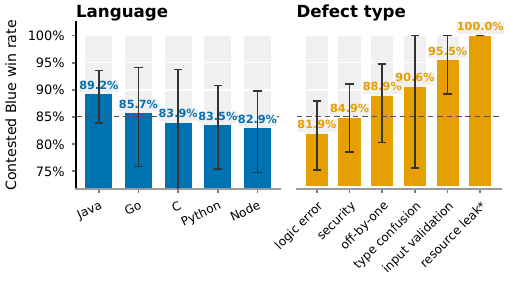}
\captionof{figure}{\textbf{Contested Blue win rates by language (left) and defect type
(right)} with 95\% task-clustered bootstrap intervals. The resource-leak
bar (*) is based on one task (nine turns) and is not interpreted.
Missing-Red auto-wins are excluded.}
\label{fig:asymmetry}
\end{minipage}

\begin{minipage}{\columnwidth}
\centering
\includegraphics[width=.5\columnwidth]{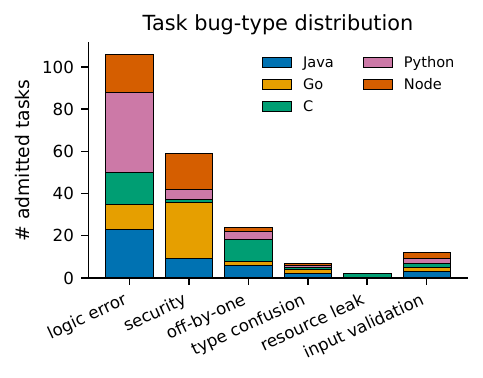}
\captionof{figure}{Distribution of admitted tasks across the observed defect
taxonomy. Taxonomy frequency describes the generated bank and should not be
interpreted as the prevalence of defects in human software.}
\label{fig:bugtype-dist}
\end{minipage}

\section{Harness Ablation Study}
\label{app:harness-details}

In the harness tournament, we select three harnesses 
(mini-swe-agent, Codex CLI, and Claude Code), and 
three models (DeepSeek-V4-Pro, GLM-5.2, or Kimi-K2.7-Code).
Each of the selected model creats tasks under all three selected harnesses
on all $15$ repositories.
Thus, each harness participant pools the three base models ($15$-task slots each)
into a single strength per harness of $45$ tasks.
We conducted a round-robin tournament where each model plays against 
same-model but different-harness participant.
Thus, no cross-model sub-match contributes to this tournament.
The BT column of Table~\ref{tab:harness} presents the match-level 
Bradley--Terry model refit with the models-aggregated harnesses as players.
Each row of Table~\ref{tab:harness} represents the performance of each harness
aggregated over three base models.
Since there are $3$ harnesses selected for this tournament, each harness gets $2$
matchups, giving $45\times 2 = 90$ Red turns for each harness throughout the 
harness--ablation tournament.
Involving Blue turns as well the opposite direction, each harness records
$90$ Red + $90$ Blue = $180$ total turns throughout the 
harness ablation tournament.

\begin{figure*}[h]
\centering
\includegraphics[width=0.5\textwidth]{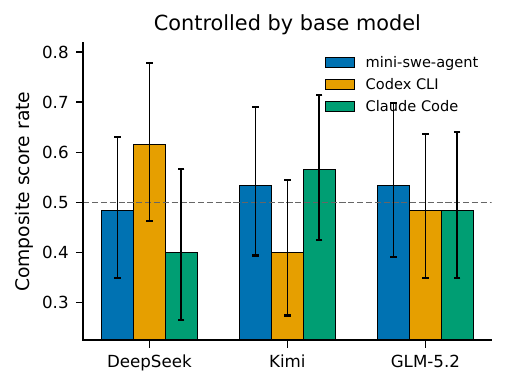}
\caption{
Harness ablation controlled by base model. 
Composite two-role score rate for each fixed base model under 
each of the three harnesses; 
whiskers are 95\% task-clustered percentile intervals 
(60 turns, 45 clusters per bar).
}
\label{fig:harness}
\end{figure*}

\textbf{Statistical resolution.}
Since turns are decisive Red--Blue outcomes, we derive the 95\% CI for score rate 
with using clustered Bernoulli draws.
The three aggregate point estimates span $0.03$ score-rate points 
while each interval spans about $0.17$, and even the largest within-model contrast 
(DeepSeek V4 Pro: Codex CLI $0.62\,[0.46,0.78]$ vs.\ Claude 
Code $0.40\,[0.27,0.57]$) overlaps on $[0.46,0.57]$. 
Table~\ref{tab:harness-pairings} shows the head-to-head match records between
each harness pair, and its the score margins corroborates the point estimates
in Table~\ref{tab:harness}: $+2\,[-16,+22]$, $+2\,[-16,+20]$, and $-8\,[-28,+12]$
points out of 90 paired turns, so paired differences below roughly $0.2$ 
score-rate points are undetectable at this budget. 
Figure~\ref{fig:harness} breaks down the performance of each harness
when controlled by base model.
Within every model the three intervals overlap, so just like models-aggregated
statistics, model-controlled bar crossings are descriptive, and cannot establish 
a per-model harness ranking.

\begin{table}[h]
\centering
\caption{
Harness ablation aggregated over three fixed base models. 
Score sums both roles over the full schedule;
Rate is the score rate with a 95\% task-clustered percentile interval; 
Red win and Blue win are per-role turn success rates over the same full schedule. 
BT is the match-level Bradley--Terry log-strength fit to the nine
same-model sub-match outcomes pooled across the base models, with a 95\%
percentile interval from the ordinary match bootstrap used by the headline
rating.}
\label{tab:harness}
\small
\setlength{\tabcolsep}{3pt}
\input{Tables/tab_harness_ablation}
\end{table}

\textbf{Harness effect vs.\ model--harness fit.}
The two CLI harnesses are prompt- and tool-engineered around their providers'
model families, whereas mini-swe-agent is a minimalist harness without a
prompt tuned for specific models. 
In the ablation, all three are cross-paired with DeepSeek V4 Pro, GLM 5.2, 
and Kimi K2.7 Code. Thus, there are no native (unfair) pairings that could 
confound harness effect analysis.
Native pairings occur only in the main tournament (GPT-5.5 with Codex CLI
and Claude Opus 4.8 with Claude Code) where we just want to observe their 
performances for the sake of curiosity, and not meant for harness probe study.

\begin{table*}[h]
\centering
\caption{
Aggregate score for each harness pair.
Margin is (A$-$B) points with a 95\% clustered percentile interval on the 
paired per-turn difference. 
The direct comparisons resolve 
no ordering at this sample size and the apparent cycle in pair totals 
is consistent with sampling noise.
}
\label{tab:harness-pairings}
\input{Tables/tab_harness_pairings}
\end{table*}

\section{Cost}
\label{sec:resource-budgets}



\begin{table*}[t]
\centering
\caption{Generation, blind self-review validation, and Blue evaluation API cost
by competitor. Generation cost includes failed attempts in slots that never
produced a task, preventing successful-only accounting.}
\label{tab:costs}
\input{Tables/tab_costs}
\end{table*}

\begin{figure*}[t]
\centering
\includegraphics[width=0.92\textwidth]{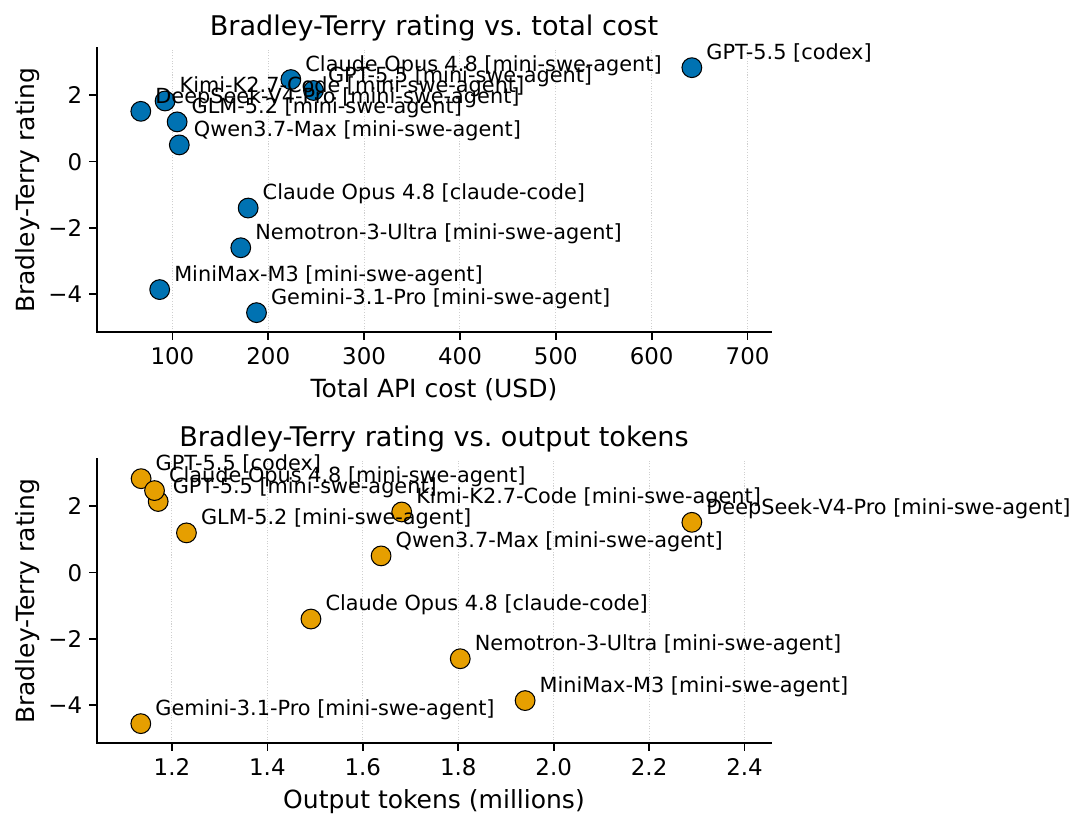}
\caption{Secondary efficiency diagnostic relating the tournament's
Bradley--Terry log-strength rating to API cost
and output-token use. These observational associations combine model pricing,
generation coverage, and behavior; they are not a controlled efficiency
comparison.}
\label{fig:efficiency}
\end{figure*}

Cost is partitioned into Red generation, blind self-review validation, 
and Blue evaluation.
A successfully created task includes the costs previous failed attempts 
within the same slot. 
The full-schedule also counts costs of all the failed attempts in exhausted slots 
that never appear in a contested turn.
Table~\ref{tab:costs} shows the breakdown of costs by generation, 
self-review validation, and blue evaluation for each participant 
throughout the main round-robin tournament, 
and Figure~\ref{fig:efficiency} shows the scatter plot visualizing 
the performance of the participants by total output tokens and API cost (in USD). 
GPT-5.5 with Codex cost \$641.89, 2.6 times the next-highest total 
and 7.4 times MiniMax-M3. 
These totals also vary with generation yield (some participants taken
multiple attempts to generate a task, causing the cost to increase), 
and number of matches (the Gemini 3.1 Pro late entrant plays only five 
active-sampled matches).

\textbf{Turn-level Blue spend.}
Table~\ref{tab:defense-spend} breaks down the main tournament's $1{,}219$ 
contested turns into each participant's Blue turns. 
Participant totals range from 0.64M to 1.58M output tokens, and turn medians from 5.9k to 11.1k.
Nine of eleven correlation intervals include zero. 
Nemotron-3-Ultra has a positive association ($r=.384$ [.213, .532]), 
whereas GPT-5.5 with mini-swe-agent has a small negative one 
($r=-.107$ [-.197, -.004]). 
Difficult tasks can induce longer trajectories in the \emph{Blue} agent,
while easier tasks can be solved in shorter trajectories.
Thus, this analysis is descriptive and exploratory.

\begin{table*}[h]
\centering
\caption{Contested Blue task-solving resource use. ``Turns'' is the 
number of all contested Blue turns for a given participant. 
``Success'' is the rate of successful task solving out of all turns; 
``Output M'' is the total output tokens over all contested turns, in millions;
``Median k'' is the median per-turn output tokens, in thousands.
The final column is Pearson's
correlation between binary task-solving success and $\log(1+\text{output tokens})$,
with a task-cluster bootstrap 95\% interval.}
\label{tab:defense-spend}
\resizebox{\textwidth}{!}{\input{Tables/tab_defense_spend}}
\end{table*}

\textbf{Output-token-matched task-solving ranking.}
As a supplementary sensitivity analysis, we cap every participant at 638,976
recorded output tokens, the smallest available participant total. In each of
1,000 draws we randomly order that participant's complete-case contested
solving runs, admit turns without replacement while the cap permits, and refit the
role-conditioned model
$P(\text{Blue wins})=\operatorname{sigmoid}(d_{\mathrm{Blue}}-a_{\mathrm{Red}})$.
Table~\ref{tab:token-budget} reports medians and 95\% intervals over this
subsampling distribution. GPT-5.5/Codex and GPT-5.5/mini-swe remain first and
second; the matched order is close to the complete-case solving order ($\tau=1.000$).

\begin{table*}[h]
\centering
\caption{
Output-token-budget-matched contested task-solving ranking at 638,976
tokens per participant. 
This task-solving-only estimand excludes Red task
creation, uses provider-reported output tokens,
and is not meant to replace the main tournament ranking.
}
\label{tab:token-budget}
\resizebox{\textwidth}{!}{\input{Tables/tab_token_budget}}
\end{table*}

\section{Newcomer Onboarding and Marginal Cost}
\label{app:newcomer}

The main tournament admitted one late entrant, Gemini-3.1-Pro with
mini-swe-agent, after the initial ten-player round-robin had completed. 
The entrant followed the identical task creation, validation gates, 
and self-review procedure to author its own bank over the 
same 15 repository slots, then sampled 
$\lceil\log_2 N\rceil+1=5$ matches out of $N=10$ incumbents, 
using the active-sampling algorithm presented by 
Chatbot Arena~\citep{chiang_chatbot_2024}
(Section~\ref{sec:tournament}). 
Ten of its 15 slots were admitted; 
the five exhausted slots gives missing-Red auto-wins 
for its opponents exactly as for incumbents. 
Nothing about the completed tournament is recomputed: 
incumbent banks, the 45 completed matches, 
and all cached solving runs are reused unchanged,
and the Bradley--Terry fit simply extends.

\textbf{Marginal entrance cost.}
Table~\ref{tab:newcomer} juxtaposes the main tournament's \$2,109.21 total API 
spend and the late entrant's costs. 
The marginal entrance cost of \$243.40 comprises \$125.95 of 
task creation (generation cost of \$106.18 and \$19.77 of blind 
self-review validation, including every failed attempt in
the entrant's slots) and \$117.45 of match evaluation: 
\$61.81 for the entrant's 55 fresh solving runs on incumbent
tasks and \$55.64 for the incumbents' 48 recorded solving runs on 
the entrant's 10 admitted tasks. 

\begin{table}[h]
\centering
\caption{API-cost ledger: the main tournament's cost split and the
marginal cost of the late entrance. Shares are fractions of the tournament
total; per-participant totals are in Table~\ref{tab:costs}.}
\label{tab:newcomer}
\small
\input{Tables/tab_newcomer}
\end{table}

\textbf{Late entrant placement.}
The five-match entrance resolved a coherent placement. 
Gemini-3.1-Pro finished with W/D/L record of $0$--$1$--$4$, and the 
full-schedule BT point estimate ($-4.55$, 95\% CI $[-12.57,-2.45]$),
ranking last with a bootstrap median rank of 11 
(central 95\% interval 9--11). 
Its contested solving success is $.618$ $[.491,.741]$, the
lowest of the participant pool, and its contested Red win rate $.042$. 
Refitting BT on the 45 incumbent-only matches reproduces 
the incumbent ordering of the full 50-match fit exactly 
(Kendall's $\tau=1.0$).
Thus, the late entrant extends the leaderboard without perturbing it. 

\textbf{Counterfactual entrance cost.}
Playing all ten incumbents instead of 5 active-sample matches 
would have added five matches. 
Using the observed per-run averages 
(\$1.12 per entrant solving run, \$1.16 per incumbent solving run on an
entrant task) and the unsampled incumbents' generation coverage 
(69 additional entrant solving runs, plus $5\times10=50$ additional 
incumbent solving runs on the entrant's bank), 
the extra matches would have cost roughly \$136, so a full
ten-pairing entrance is estimated at about \$379.
Thus, the active sampling procedure saved about a
third of the entrance's evaluation spend, 
and an entrant's marginal cost scales with
$\lceil\log_2 N\rceil+1$ matches rather than the $O(N)$ pairings 
of a full round robin.

\newpage
\section{Textual interface}

\begin{figure*}[h]
\centering
\includegraphics[width=\textwidth]{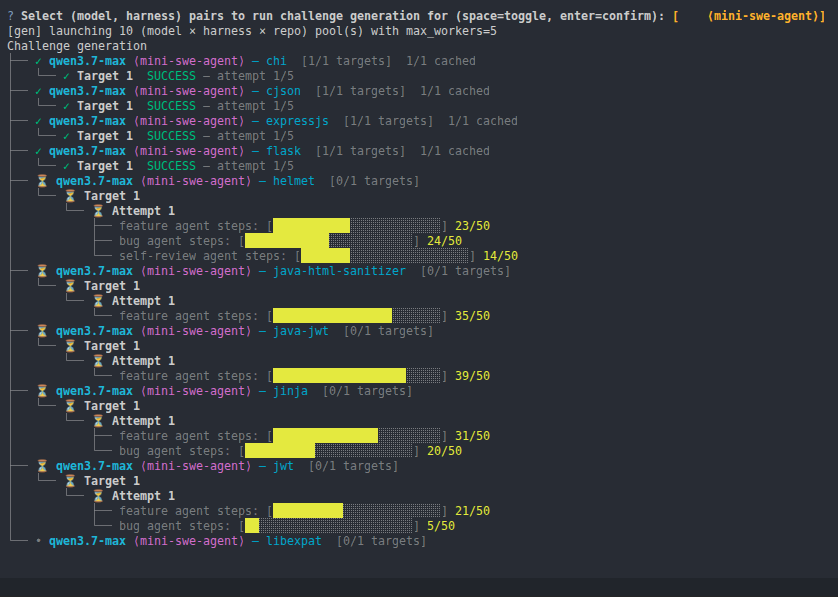}
\caption{Textual interface for task generation. Rows correspond to numbered
generation slots; each row tracks the feature/bug sessions, gate outcomes, and
attempt history of that slot.}
\label{fig:tui-challenge-generation}
\end{figure*}

\begin{figure*}[h]
\centering
\includegraphics[width=\textwidth]{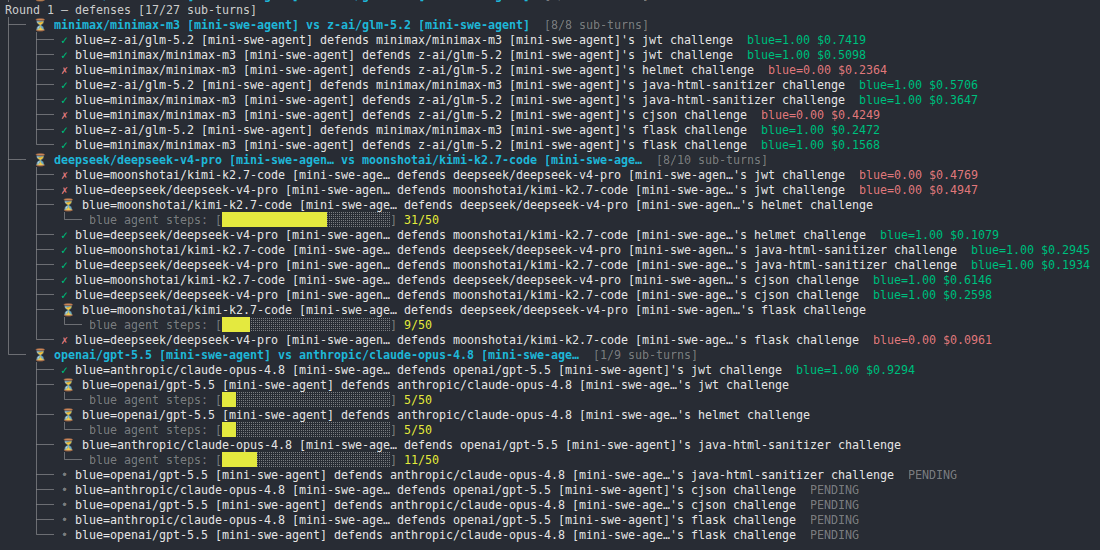}
\caption{Textual interface for tournament scheduling. Pairings are drawn from
the persisted schedule, solving runs are served from the cache, and completed
matches report per-side outcomes and cost.}
\label{fig:tui-tournament}
\end{figure*}

The framework ships an interactive terminal interface for task generation
and tournament management. Figure~\ref{fig:tui-challenge-generation} shows the
task-generation view, and
Figure~\ref{fig:tui-tournament} shows the tournament view.

%% file: sections/worked_challenges.tex
\section{Complete Worked Tasks}
\label{app:worked-challenges}


\subsection{Python: SQLAlchemy \texttt{LRUCache.get\_or\_create}}
\noindent\textbf{Artifacts.} Task \texttt{e1a27fac-1285-4690-ab55-cb8fabca312d};
successful solving run \texttt{cd538560-b183-4eba-a59f-dc7862e90ba8}.

\paragraph{Feature request.}
Add \texttt{LRUCache.get\_or\_create(key, creator)}, which returns and refreshes
an existing cached value on a hit, or calls a zero-argument creator on a miss,
stores the created value through normal LRU size management, and returns it. If
the creator raises, no entry is stored; if it reentrantly populates the key, that
value is returned instead of being overwritten.

\listingheading{Complete PR diff.}
\begin{lstlisting}[style=artifactdiff, language=appendixdiff]
--- a/lib/sqlalchemy/util/_collections.py
+++ b/lib/sqlalchemy/util/_collections.py
`\diffhunkstart{506}`@@ -506,6 +506,29 @@
         else:
             return default
 
+    def get_or_create(self, key: _KT, creator: Callable[[], _VT]) -> _VT:
+        """Return a value from the cache, creating it if absent.
+
+        This provides an atomic-looking convenience for the common cache
+        pattern of retrieving a value and lazily populating it on a miss,
+        while preserving normal LRU access tracking for hits.
+        """
+
+        value = self.get(key)
+        if value is not None:
+            return value
+
+        value = creator()
+
+        item = self._data.get(key)
+        if item is not None:
+            item[2][0] = self._inc_counter()
+            return item[1]
+
+        self._data[key] = (key, value, [self._inc_counter()])
+        self._manage_size()
+        return value
+
     def __getitem__(self, key: _KT) -> _VT:
         item = self._data[key]
         item[2][0] = self._inc_counter()
\end{lstlisting}

\paragraph{Embedded defect (line 517).}
The initial \texttt{get(key)} uses \texttt{None} as the implicit default, so a
cached \texttt{None} is treated as a miss. A side-effecting creator can run and
valid cached state can be overwritten or evicted.

\listingheading{Complete sealed bug test.}
\begin{lstlisting}[style=artifact, language=Python]
from sqlalchemy import util

def test_lru_get_or_create_treats_cached_none_as_existing_value():
    cache = util.LRUCache(2, threshold=0)
    calls = []

    cache["none-key"] = None
    cache["other-key"] = "other-value"

    value = cache.get_or_create(
        "none-key", lambda: calls.append("created") or "new-value"
    )
    cache["new-key"] = "new-entry"

    assert value is None
    assert calls == []
    assert "none-key" in cache
    assert "other-key" not in cache
    assert "new-key" in cache
\end{lstlisting}

\listingheading{Complete successful Blue patch.}
\begin{lstlisting}[style=artifactdiff, language=appendixdiff]
--- a/lib/sqlalchemy/util/_collections.py
+++ b/lib/sqlalchemy/util/_collections.py
`\diffhunkstart{506}`@@ -506,6 +506,28 @@
         else:
             return default
 
+    def get_or_create(self, key: _KT, creator: Callable[[], _VT]) -> _VT:
+        """Return a cached value, creating and caching it if absent.
+
+        If the creator populates *key* reentrantly, that value takes
+        precedence over the value returned by the creator.
+        """
+        item = self._data.get(key)
+        if item is not None:
+            item[2][0] = self._inc_counter()
+            return item[1]
+
+        value = creator()
+
+        item = self._data.get(key)
+        if item is not None:
+            item[2][0] = self._inc_counter()
+            return item[1]
+
+        self._data[key] = (key, value, [self._inc_counter()])
+        self._manage_size()
+        return value
+
     def __getitem__(self, key: _KT) -> _VT:
         item = self._data[key]
         item[2][0] = self._inc_counter()
\end{lstlisting}

\subsection{Go: chi \texttt{Mux.AllowedMethods}}
\noindent\textbf{Artifacts.} Task \texttt{af3fb594-76dd-4f6e-8570-42130eb346d3};
successful solving run \texttt{dff4a747-68d3-4e68-9850-5f54b288432e}.

\paragraph{Feature request.}
Add \texttt{(*Mux).AllowedMethods(path string) []string}, returning a stable,
sorted list of methods registered for a concrete path. Lookup normalizes an
empty path to \texttt{/}, supports parameters, mounted subrouters, catch-alls,
and custom methods, and returns \texttt{nil} when no route matches.

\listingheading{Complete PR diff.}
\begin{lstlisting}[style=artifactdiff, language=appendixdiff]
--- a/mux.go
+++ b/mux.go
`\diffhunkstart{4}`@@ -4,6 +4,7 @@
 	"context"
 	"fmt"
 	"net/http"
+	"sort"
 	"strings"
 	"sync"
 )`\difffilebreak`
`\diffhunkstart{361}`@@ -360,6 +361,41 @@
 	return mx.Find(rctx, method, path) != ""
 }
 
+// AllowedMethods returns the HTTP methods with a registered handler for path.
+//
+// The path is matched the same way as an incoming request path, including
+// parameters, catch-all routes, mounted subrouters, and custom methods
+// registered with RegisterMethod. The returned methods are sorted for stable
+// introspection output. If no route matches path, it returns nil.
+func (mx *Mux) AllowedMethods(path string) []string {
+	if path == "" {
+		path = "/"
+	}
+
+	methods := make([]string, 0, len(methodMap))
+	for method := range methodMap {
+		methods = append(methods, method)
+	}
+	sort.Strings(methods)
+
+	rctx := NewRouteContext()
+	allowed := methods[:0]
+	for _, method := range methods {
+		rctx.Reset()
+		if mx.Find(rctx, method, path) != "" {
+			allowed = append(allowed, method)
+			if len(rctx.RoutePatterns) > 2 {
+				break
+			}
+		}
+	}
+	if len(allowed) == 0 {
+		return nil
+	}
+
+	return append([]string(nil), allowed...)
+}
+
 // Find searches the routing tree for the pattern that matches
 // the method/path.
 //
\end{lstlisting}

\paragraph{Embedded defect (lines 387--389).}
After the first method match, lookup stops whenever nested mounting produced
more than two route patterns. Other methods registered at the same nested path
are omitted, which can produce an incomplete \texttt{Allow} or CORS response.

\listingheading{Complete sealed bug test.}
\begin{lstlisting}[style=artifact, language=Go]
package acaatest

import (
	"net/http"
	"reflect"
	"testing"

	"github.com/go-chi/chi/v5"
)

func TestAllowedMethodsAcrossNestedMountedSubrouters(t *testing.T) {
	leaf := chi.NewRouter()
	leaf.Delete("/items/{id}", noopBugHandler)
	leaf.Patch("/items/{id}", noopBugHandler)

	versioned := chi.NewRouter()
	versioned.Mount("/v1", leaf)

	r := chi.NewRouter()
	r.Mount("/api", versioned)

	got := r.AllowedMethods("/api/v1/items/abc123")
	want := []string{http.MethodDelete, http.MethodPatch}
	if !reflect.DeepEqual(got, want) {
		t.Fatalf("AllowedMethods returned %v, want %v", got, want)
	}
}

func noopBugHandler(w http.ResponseWriter, r *http.Request) {}
\end{lstlisting}

\listingheading{Complete successful Blue patch.}
\begin{lstlisting}[style=artifactdiff, language=appendixdiff]
--- a/mux.go
+++ b/mux.go
`\diffhunkstart{4}`@@ -4,6 +4,7 @@
 	"context"
 	"fmt"
 	"net/http"
+	"sort"
 	"strings"
 	"sync"
 )`\difffilebreak`
`\diffhunkstart{361}`@@ -360,6 +361,38 @@
 	return mx.Find(rctx, method, path) != ""
 }
 
+// AllowedMethods returns the HTTP methods with a registered handler for path.
+//
+// The path is matched the same way as an incoming request path, including
+// parameters, catch-all routes, mounted subrouters, and custom methods
+// registered with RegisterMethod. The returned methods are sorted for stable
+// introspection output. If no route matches path, it returns nil.
+func (mx *Mux) AllowedMethods(path string) []string {
+	if path == "" {
+		path = "/"
+	}
+
+	methods := make([]string, 0, len(methodMap))
+	for method := range methodMap {
+		methods = append(methods, method)
+	}
+	sort.Strings(methods)
+
+	rctx := NewRouteContext()
+	allowed := methods[:0]
+	for _, method := range methods {
+		rctx.Reset()
+		if mx.Find(rctx, method, path) != "" {
+			allowed = append(allowed, method)
+		}
+	}
+	if len(allowed) == 0 {
+		return nil
+	}
+
+	return append([]string(nil), allowed...)
+}
+
 // Find searches the routing tree for the pattern that matches
 // the method/path.
 //
\end{lstlisting}

\subsection{Node: node-jsonwebtoken \texttt{requiredClaims}}
\noindent\textbf{Artifacts.} Task \texttt{fb6f0e8c-b6e9-4bd5-a2eb-4970d8856ae1};
successful solving run \texttt{f1b3beb2-769c-438a-a9a1-bf656a5d45d7}.

\paragraph{Feature request.}
Add a \texttt{requiredClaims} option to \texttt{jwt.verify} accepting one claim
name or an array. After signature and registered-claim checks, every listed
claim must be present (not \texttt{undefined}) or verification fails with
\texttt{JsonWebTokenError}; the option must be a nonempty string or nonempty
array of nonempty strings.

\listingheading{Complete PR diff.}
\begin{lstlisting}[style=artifactdiff, language=appendixdiff]
--- a/verify.js
+++ b/verify.js
`\diffhunkstart{54}`@@ -54,6 +54,13 @@
     return done(new JsonWebTokenError('allowInvalidAsymmetricKeyTypes must be a boolean'));
   }
 
+  if (options.requiredClaims !== undefined) {
+    const requiredClaims = Array.isArray(options.requiredClaims) ? options.requiredClaims : [options.requiredClaims];
+    if (requiredClaims.length === 0 || !requiredClaims.every(function (claim) { return typeof claim === 'string' && claim.trim() !== ''; })) {
+      return done(new JsonWebTokenError('requiredClaims must be a non-empty string or an array of non-empty strings'));
+    }
+  }
+
   const clockTimestamp = options.clockTimestamp || Math.floor(Date.now() / 1000);
 
   if (!jwtString){`\difffilebreak`
`\diffhunkstart{241}`@@ -234,6 +241,7 @@
       }
     }
 
+
     if (options.maxAge) {
       if (typeof payload.iat !== 'number') {
         return done(new JsonWebTokenError('iat required when maxAge is specified'));`\difffilebreak`
`\diffhunkstart{266}`@@ -258,6 +266,16 @@
       });
     }
 
+    if (options.requiredClaims) {
+      const requiredClaims = Array.isArray(options.requiredClaims) ? options.requiredClaims : [options.requiredClaims];
+      for (let i = 0; i < requiredClaims.length; i++) {
+        const claim = requiredClaims[i];
+        if (typeof payload[claim] === 'undefined') {
+          return done(new JsonWebTokenError('jwt claim "' + claim + '" is missing'));
+        }
+      }
+    }
+
     return done(null, payload);
   });
 };
\end{lstlisting}

\paragraph{Embedded defect (lines 269--277).}
With \texttt{complete: true}, verification returns the complete token before
reaching the new check. Mandatory authorization claims can therefore be absent
without error.

\listingheading{Complete sealed bug test.}
\begin{lstlisting}[style=artifact, language=appendixjavascript]
'use strict';

const assert = require('assert');
const jwt = require('../');

describe('requiredClaims with complete option', function () {
  const secret = 'shhhhh';

  it('should still reject a token missing a required claim when complete is true', function (done) {
    const token = jwt.sign({ role: 'admin' }, secret, { algorithm: 'HS256' });
    jwt.verify(token, secret, { requiredClaims: 'permission', complete: true }, function (err, decoded) {
      assert.ok(err, 'expected an error when a required claim is missing');
      assert.strictEqual(err.name, 'JsonWebTokenError');
      assert.ok(err.message.includes('jwt claim "permission" is missing'));
      assert.strictEqual(decoded, undefined);
      done();
    });
  });
});
\end{lstlisting}

\listingheading{Complete successful Blue patch.}
\begin{lstlisting}[style=artifactdiff, language=appendixdiff]
--- a/verify.js
+++ b/verify.js
`\diffhunkstart{54}`@@ -54,6 +54,13 @@
     return done(new JsonWebTokenError('allowInvalidAsymmetricKeyTypes must be a boolean'));
   }
 
+  if (options.requiredClaims !== undefined) {
+    const requiredClaims = Array.isArray(options.requiredClaims) ? options.requiredClaims : [options.requiredClaims];
+    if (requiredClaims.length === 0 || !requiredClaims.every(function (claim) { return typeof claim === 'string' && claim.trim() !== ''; })) {
+      return done(new JsonWebTokenError('requiredClaims must be a non-empty string or an array of non-empty strings'));
+    }
+  }
+
   const clockTimestamp = options.clockTimestamp || Math.floor(Date.now() / 1000);
 
   if (!jwtString){`\difffilebreak`
`\diffhunkstart{255}`@@ -248,6 +255,16 @@
       }
     }
 
+    if (options.requiredClaims) {
+      const requiredClaims = Array.isArray(options.requiredClaims) ? options.requiredClaims : [options.requiredClaims];
+      for (let i = 0; i < requiredClaims.length; i++) {
+        const claim = requiredClaims[i];
+        if (typeof payload[claim] === 'undefined') {
+          return done(new JsonWebTokenError('jwt claim "' + claim + '" is missing'));
+        }
+      }
+    }
+
     if (options.complete === true) {
       const signature = decodedToken.signature;
\end{lstlisting}

\subsection{Java: OWASP sanitizer description lists}
\noindent\textbf{Artifacts.} Task \texttt{eef5eb63-11e9-4e2f-9810-1bc40816e6a2};
successful solving run \texttt{369e5c91-9b26-48c4-8cf5-eebe3cce83f3}.

\paragraph{Feature request.}
Add a pre-packaged \texttt{Sanitizers.DESCRIPTION\_LISTS} policy allowing
\texttt{dl}, \texttt{dt}, and \texttt{dd} without attributes, composable with
policies such as \texttt{FORMATTING} or \texttt{BLOCKS}.

\listingheading{Complete PR diff.}
\begin{lstlisting}[style=artifactdiff, language=appendixdiff]
--- a/owasp-java-html-sanitizer/src/main/java/org/owasp/html/Sanitizers.java
+++ b/owasp-java-html-sanitizer/src/main/java/org/owasp/html/Sanitizers.java
`\diffhunkstart{96}`@@ -96,6 +96,22 @@
     .allowTextIn("table")  // WIDGY
     .toFactory();
 
+  /**
+   * Allows description list elements: {@code <dl>}, {@code <dt>}, and
+   * {@code <dd>}.  These are useful for glossaries, metadata displays,
+   * and FAQ sections.  Does not allow any attributes on these elements.
+   * <p>
+   * Combine with {@link #FORMATTING} to allow inline formatting inside
+   * {@code <dt>} and {@code <dd>} elements, e.g.
+   * <pre>
+   *   PolicyFactory sanitizer = Sanitizers.DESCRIPTION_LISTS
+   *       .and(Sanitizers.FORMATTING);
+   * </pre>
+   */
+  public static final PolicyFactory DESCRIPTION_LISTS = new HtmlPolicyBuilder()
+      .allowElements("dl", "dt", "dd", "td")
+      .toFactory();
+
   private static final AttributePolicy INTEGER = new AttributePolicy() {
     public String apply(
         String elementName, String attributeName, String value) {
\end{lstlisting}

\paragraph{Embedded defect (line 112).}
The allowlist accidentally includes \texttt{td}. This visually similar
copy-paste token preserves table cells despite the policy's documented
description-list-only contract.

\listingheading{Complete sealed bug test.}
\begin{lstlisting}[style=artifact, language=Java]
import static org.junit.Assert.*;

import org.junit.Test;
import org.owasp.html.Sanitizers;

@SuppressWarnings("javadoc")
public class TestAcaaBug {

  @Test
  public void testDescriptionListsDoesNotAllowTableDataCell() {
    String input = "<td>Should be stripped</td>";
    String want = "Should be stripped";
    String got = Sanitizers.DESCRIPTION_LISTS.sanitize(input);
    assertEquals(want, got);
  }

  @Test
  public void testDescriptionListsStripsTableElementsInsideDl() {
    String input = "<dl><dt>Term</dt><dd>Definition</dd><td>sneaky cell</td></dl>";
    String want = "<dl><dt>Term</dt><dd>Definition</dd>sneaky cell</dl>";
    String got = Sanitizers.DESCRIPTION_LISTS.sanitize(input);
    assertEquals(want, got);
  }

  @Test
  public void testDescriptionListsDoesNotAllowTableHeaderCell() {
    String input = "<th>Table header</th>";
    String want = "Table header";
    String got = Sanitizers.DESCRIPTION_LISTS.sanitize(input);
    assertEquals(want, got);
  }
}
\end{lstlisting}

\listingheading{Complete successful Blue patch.}
\begin{lstlisting}[style=artifactdiff, language=appendixdiff]
--- a/owasp-java-html-sanitizer/src/main/java/org/owasp/html/Sanitizers.java
+++ b/owasp-java-html-sanitizer/src/main/java/org/owasp/html/Sanitizers.java
`\diffhunkstart{96}`@@ -96,6 +96,22 @@
     .allowTextIn("table")  // WIDGY
     .toFactory();
 
+  /**
+   * Allows description list elements: {@code <dl>}, {@code <dt>}, and
+   * {@code <dd>}.  These are useful for glossaries, metadata displays,
+   * and FAQ sections.  Does not allow any attributes on these elements.
+   * <p>
+   * Combine with {@link #FORMATTING} to allow inline formatting inside
+   * {@code <dt>} and {@code <dd>} elements, e.g.
+   * <pre>
+   *   PolicyFactory sanitizer = Sanitizers.DESCRIPTION_LISTS
+   *       .and(Sanitizers.FORMATTING);
+   * </pre>
+   */
+  public static final PolicyFactory DESCRIPTION_LISTS = new HtmlPolicyBuilder()
+      .allowElements("dl", "dt", "dd")
+      .toFactory();
+
   private static final AttributePolicy INTEGER = new AttributePolicy() {
     public String apply(
         String elementName, String attributeName, String value) {
\end{lstlisting}

\subsection{C/C++: cJSON boolean arrays}
\noindent\textbf{Artifacts.} Task \texttt{b39a0522-ce05-4a16-8486-56425000ef52};
successful solving run \texttt{aadee125-8b6a-4c3a-9ccb-7935b025dd48}.

\paragraph{Feature request.}
Add \texttt{cJSON\_CreateBoolArray(const cJSON\_bool *booleans, int count)},
returning a cJSON array whose boolean items correspond to the input values.

\listingheading{Complete PR diff.}
\begin{lstlisting}[style=artifactdiff, language=appendixdiff]
--- a/cJSON.c
+++ b/cJSON.c
`\diffhunkstart{2603}`@@ -2603,6 +2603,46 @@
 }
 
 /* Create Arrays: */
+CJSON_PUBLIC(cJSON *) cJSON_CreateBoolArray(const cJSON_bool *booleans, int count)
+{
+    size_t i = 0;
+    cJSON *n = NULL;
+    cJSON *p = NULL;
+    cJSON *a = NULL;
+
+    if ((count < 0) || (booleans == NULL))
+    {
+        return NULL;
+    }
+
+    a = cJSON_CreateArray();
+
+    for(i = 0; a && (i < (size_t)count); i++)
+    {
+        n = cJSON_CreateBool(booleans[i]);
+        if (!n)
+        {
+            cJSON_Delete(a);
+            return NULL;
+        }
+        if(!i)
+        {
+            a->child = n;
+        }
+        else
+        {
+            suffix_object(p, n);
+        }
+        p = n;
+    }
+
+    if (a && a->child) {
+        a->child->prev = a->child;
+    }
+
+    return a;
+}
+
 CJSON_PUBLIC(cJSON *) cJSON_CreateIntArray(const int *numbers, int count)
 {
     size_t i = 0;`\difffilebreak`
--- a/cJSON.h
+++ b/cJSON.h
`\diffhunkstart{219}`@@ -219,6 +219,7 @@
 
 /* These utilities create an Array of count items.
  * The parameter count cannot be greater than the number of elements in the number array, otherwise array access will be out of bounds.*/
+CJSON_PUBLIC(cJSON *) cJSON_CreateBoolArray(const cJSON_bool *booleans, int count);
 CJSON_PUBLIC(cJSON *) cJSON_CreateIntArray(const int *numbers, int count);
 CJSON_PUBLIC(cJSON *) cJSON_CreateFloatArray(const float *numbers, int count);
 CJSON_PUBLIC(cJSON *) cJSON_CreateDoubleArray(const double *numbers, int count);
\end{lstlisting}

\paragraph{Embedded defect (lines 2639--2641).}
The array's tail pointer, \texttt{a->child->prev}, points to the first element
rather than the last element \texttt{n}. Subsequent constant-time append uses
the wrong tail and can corrupt list order or structure.

\listingheading{Complete sealed bug test.}
\begin{lstlisting}[style=artifact, language=C]
#include <stdio.h>
#include <stdlib.h>
#include <assert.h>
#include "cJSON.h"

void test_bug(void) {
    cJSON_bool booleans[3] = {1, 0, 1};
    cJSON *array = cJSON_CreateBoolArray(booleans, 3);

    assert(array != NULL);
    assert(cJSON_IsArray(array));
    assert(cJSON_GetArraySize(array) == 3);

    assert(array->child != NULL);
    assert(array->child->prev != NULL);
    assert(array->child->prev == cJSON_GetArrayItem(array, 2));

    cJSON *new_item = cJSON_CreateBool(1);
    cJSON_AddItemToArray(array, new_item);

    assert(cJSON_GetArraySize(array) == 4);
    assert(cJSON_GetArrayItem(array, 3) == new_item);
    assert(array->child->prev == new_item);

    cJSON_Delete(array);
}

int main(void) {
    test_bug();
    printf("All tests passed!\n");
    return 0;
}
\end{lstlisting}

\listingheading{Complete successful Blue patch.}
\begin{lstlisting}[style=artifactdiff, language=appendixdiff]
--- a/cJSON.c
+++ b/cJSON.c
`\diffhunkstart{2602}`@@ -2602,6 +2602,46 @@
     return item;
 }
 
+CJSON_PUBLIC(cJSON *) cJSON_CreateBoolArray(const cJSON_bool *booleans, int count)
+{
+    size_t i = 0;
+    cJSON *n = NULL;
+    cJSON *p = NULL;
+    cJSON *a = NULL;
+
+    if ((count < 0) || (booleans == NULL))
+    {
+        return NULL;
+    }
+
+    a = cJSON_CreateArray();
+
+    for(i = 0; a && (i < (size_t)count); i++)
+    {
+        n = cJSON_CreateBool(booleans[i]);
+        if (!n)
+        {
+            cJSON_Delete(a);
+            return NULL;
+        }
+        if(!i)
+        {
+            a->child = n;
+        }
+        else
+        {
+            suffix_object(p, n);
+        }
+        p = n;
+    }
+
+    if (a && a->child) {
+        a->child->prev = n;
+    }
+
+    return a;
+}
+
 /* Create Arrays: */
 CJSON_PUBLIC(cJSON *) cJSON_CreateIntArray(const int *numbers, int count)
 {`\difffilebreak`
--- a/cJSON.h
+++ b/cJSON.h
`\diffhunkstart{219}`@@ -219,6 +219,7 @@
 
 /* These utilities create an Array of count items.
  * The parameter count cannot be greater than the number of elements in the number array, otherwise array access will be out of bounds.*/
+CJSON_PUBLIC(cJSON *) cJSON_CreateBoolArray(const cJSON_bool *booleans, int count);
 CJSON_PUBLIC(cJSON *) cJSON_CreateIntArray(const int *numbers, int count);
 CJSON_PUBLIC(cJSON *) cJSON_CreateFloatArray(const float *numbers, int count);
 CJSON_PUBLIC(cJSON *) cJSON_CreateDoubleArray(const double *numbers, int count);
\end{lstlisting}

%% file: sections/oracle_audit.tex
\section{Manual Audit for Sealed-oracle False Negatives}
\label{app:oracle-audit}

The sealed bug test is authored by \emph{Red} agent, 
for deterministic verification of the embedded bug's existence.
However, it does not guarantee that it accepts every semantically valid repair. 
Therefore, two authors conducted a post-hoc manually audit on bug fix failures 
where existing and feature suites pass but the sealed bug suite fails. 
The main tournament contains 158 turns with such outcomes.
We sorted them by turn UUID and drew 50 without
replacement using Python's \texttt{random.Random(20260823).sample}. 
For each sampled turn, we jointly inspected the feature specification, 
admitted PR diff, defect description and location, sealed test, 
Blue explanation, and submitted patch. 
A \emph{genuine miss} means \emph{Blue} agent leaves the embedded bug unrepaired 
while a \emph{false negative} means \emph{Blue} agent implements a legitimate repair 
but fails because the sealed oracle requires an unnecessarily specific 
implementation for the bug fix. 

\paragraph{Result.}
Of 50 sampled outcomes, 45 were genuine misses, three were false negatives, and
two were ambiguous. Treating every sampled scored failure as the denominator,
the estimated false-negative rate is $3/50=6.0\%$ (95\% Wilson interval
$2.1$--$16.2\%$). Excluding the two ambiguous cases gives $3/48=6.25\%$
(95\% Wilson interval $2.1$--$16.8\%$). All three false negatives arose from two
Go JWT tasks: Blue rejected an unverifiable custom-claims token with
\texttt{ErrTokenUnverifiable} rather than the oracle's required
\texttt{ErrTokenRequiredClaimMissing} (two solving runs of one task), or
correctly treated future-\texttt{iat} validation as the existing
\texttt{WithIssuedAt} option's responsibility while the oracle required
\texttt{WithMaxTokenAge} to duplicate it. Thus, over-specification is 
uncommon in this sample.

%% file: Tables/tab_bt_uncertainty.tex
\begin{tabular}{lrr}
\toprule
\textbf{Competitor} & \textbf{BT 95\% CI} & \textbf{95\% rank interval} \\
\midrule
GPT-5.5 [codex] & $[+1.30,\,+9.48]$ & 1--5 \\
Claude Opus 4.8 [mini-swe-agent] & $[+1.10,\,+8.95]$ & 1--6 \\
GPT-5.5 [mini-swe-agent] & $[-0.57,\,+6.95]$ & 1--7 \\
Kimi-K2.7-Code [mini-swe-agent] & $[-0.07,\,+7.44]$ & 1--7 \\
DeepSeek-V4-Pro [mini-swe-agent] & $[-0.61,\,+5.87]$ & 1--7 \\
GLM-5.2 [mini-swe-agent] & $[-1.00,\,+4.44]$ & 1--7 \\
Qwen3.7-Max [mini-swe-agent] & $[-1.45,\,+3.34]$ & 4--8 \\
Claude Opus 4.8 [claude-code] & $[-6.91,\,+1.33]$ & 7--9 \\
Nemotron-3-Ultra [mini-swe-agent] & $[-12.57,\,-1.33]$ & 8--11 \\
MiniMax-M3 [mini-swe-agent] & $[-8.45,\,-2.34]$ & 8--11 \\
Gemini-3.1-Pro [mini-swe-agent] & $[-12.57,\,-2.45]$ & 9--11 \\
\bottomrule
\end{tabular}

%% file: Tables/tab_admission_noise.tex
\begin{tabular}{lrrrrrrr}
\toprule
\textbf{Participant} & \textbf{$n$} & \textbf{3/3} & \textbf{2/3} & \textbf{1/3} & \textbf{0/3} & \textbf{Repairs/runs} & \textbf{Rate (95\% Wilson)} \\
\midrule
DeepSeek-V4-Pro [mini-swe-agent] & 4 & 4 & -- & -- & -- & 12/12 & 100.0\% $[75.7,\,100.0]$ \\
Kimi-K2.7-Code [mini-swe-agent] & 4 & 4 & -- & -- & -- & 12/12 & 100.0\% $[75.7,\,100.0]$ \\
GPT-5.5 [Codex CLI] & 4 & 4 & -- & -- & -- & 12/12 & 100.0\% $[75.7,\,100.0]$ \\
GPT-5.5 [mini-swe-agent] & 4 & 4 & -- & -- & -- & 12/12 & 100.0\% $[75.7,\,100.0]$ \\
Claude Opus 4.8 [Claude Code] & 4 & 3 & 1 & -- & -- & 11/12 & 91.7\% $[64.6,\,98.5]$ \\
Gemini-3.1-Pro [mini-swe-agent] & 4 & 3 & 1 & -- & -- & 11/12 & 91.7\% $[64.6,\,98.5]$ \\
Nemotron-3-Ultra [mini-swe-agent] & 4 & 3 & 1 & -- & -- & 11/12 & 91.7\% $[64.6,\,98.5]$ \\
Qwen3.7-Max [mini-swe-agent] & 4 & 3 & 1 & -- & -- & 11/12 & 91.7\% $[64.6,\,98.5]$ \\
MiniMax-M3 [mini-swe-agent] & 4 & 2 & 2 & -- & -- & 10/12 & 83.3\% $[55.2,\,95.3]$ \\
GLM-5.2 [mini-swe-agent] & 4 & 2 & 2 & -- & -- & 10/12 & 83.3\% $[55.2,\,95.3]$ \\
Claude Opus 4.8 [mini-swe-agent] & 4 & 3 & -- & -- & 1 & 9/12 & 75.0\% $[46.8,\,91.1]$ \\
\midrule
\textbf{All participants} & \textbf{44} & \textbf{35} & \textbf{8} & \textbf{0} & \textbf{1} & \textbf{121/132} & \textbf{91.7\% $[85.7,\,95.3]$} \\
\bottomrule
\end{tabular}

%% file: Tables/tab_decomposition.tex
\begin{tabular}{rlrrrrr}
\toprule
\textbf{Full rank} & \textbf{Competitor} & \textbf{Solving BT} & \textbf{95\% CI} & \textbf{Solving success} & \textbf{Admitted / scheduled slots} & \textbf{Coverage} \\
\midrule
1 & GPT-5.5 [codex] & +2.19 & $[+1.35,\,+4.89]$ & 98.3\% & 13 / 15 & 86.7\% \\
2 & Claude Opus 4.8 [mini-swe-agent] & -0.32 & $[-0.81,\,+0.08]$ & 85.7\% & 15 / 15 & 100.0\% \\
3 & GPT-5.5 [mini-swe-agent] & +1.45 & $[+0.82,\,+2.82]$ & 96.4\% & 13 / 15 & 86.7\% \\
4 & Kimi-K2.7-Code [mini-swe-agent] & +0.09 & $[-0.42,\,+0.59]$ & 88.3\% & 13 / 15 & 86.7\% \\
5 & DeepSeek-V4-Pro [mini-swe-agent] & +0.25 & $[-0.28,\,+0.79]$ & 89.9\% & 15 / 15 & 100.0\% \\
6 & GLM-5.2 [mini-swe-agent] & +0.01 & $[-0.42,\,+0.47]$ & 87.4\% & 13 / 15 & 86.7\% \\
7 & Qwen3.7-Max [mini-swe-agent] & +0.22 & $[-0.29,\,+0.76]$ & 89.0\% & 15 / 15 & 100.0\% \\
8 & Claude Opus 4.8 [claude-code] & -0.23 & $[-0.71,\,+0.16]$ & 83.9\% & 15 / 15 & 100.0\% \\
9 & Nemotron-3-Ultra [mini-swe-agent] & -0.87 & $[-1.46,\,-0.49]$ & 75.6\% & 7 / 15 & 46.7\% \\
10 & MiniMax-M3 [mini-swe-agent] & -1.12 & $[-1.64,\,-0.84]$ & 70.5\% & 5 / 15 & 33.3\% \\
11 & Gemini-3.1-Pro [mini-swe-agent] & -1.69 & $[-2.65,\,-1.08]$ & 61.8\% & 10 / 15 & 66.7\% \\
\bottomrule
\end{tabular}

%% file: Tables/tab_turn_bt.tex
\begin{tabular}{lrrrr}
\toprule
\textbf{Competitor} & \textbf{\shortstack{Full-schedule turn BT\\(95\% CI)}} & \textbf{\shortstack{Full-schedule\\95\% rank interval}} & \textbf{\shortstack{Contested-only turn BT\\(95\% CI)}} & \textbf{\shortstack{Contested-only\\95\% rank interval}} \\
\midrule
Claude Opus 4.8 [mini-swe-agent] & $+0.30$ $[-0.10,\,+0.80]$ & 1--8 & $+0.12$ $[-0.34,\,+0.61]$ & 1--10 \\
DeepSeek-V4-Pro [mini-swe-agent] & $+0.25$ $[-0.15,\,+0.73]$ & 1--8 & $+0.07$ $[-0.31,\,+0.54]$ & 1--10 \\
GPT-5.5 [mini-swe-agent] & $+0.19$ $[-0.20,\,+0.72]$ & 1--9 & $+0.17$ $[-0.28,\,+0.66]$ & 1--9 \\
GPT-5.5 [codex] & $+0.18$ $[-0.20,\,+0.73]$ & 1--9 & $+0.15$ $[-0.29,\,+0.84]$ & 1--9 \\
Kimi-K2.7-Code [mini-swe-agent] & $+0.17$ $[-0.23,\,+0.68]$ & 1--9 & $+0.13$ $[-0.28,\,+0.62]$ & 1--10 \\
Qwen3.7-Max [mini-swe-agent] & $+0.14$ $[-0.28,\,+0.65]$ & 1--9 & $-0.05$ $[-0.46,\,+0.46]$ & 1--10 \\
GLM-5.2 [mini-swe-agent] & $+0.14$ $[-0.29,\,+0.67]$ & 1--9 & $+0.10$ $[-0.33,\,+0.59]$ & 1--10 \\
Claude Opus 4.8 [claude-code] & $-0.10$ $[-0.50,\,+0.34]$ & 2--10 & $-0.33$ $[-0.74,\,+0.14]$ & 4--11 \\
Nemotron-3-Ultra [mini-swe-agent] & $-0.26$ $[-0.70,\,+0.19]$ & 4--11 & $+0.14$ $[-0.35,\,+0.75]$ & 1--10 \\
MiniMax-M3 [mini-swe-agent] & $-0.44$ $[-0.84,\,-0.01]$ & 6--11 & $+0.10$ $[-0.39,\,+0.60]$ & 1--10 \\
Gemini-3.1-Pro [mini-swe-agent] & $-0.57$ $[-0.98,\,-0.12]$ & 7--11 & $-0.60$ $[-1.06,\,-0.10]$ & 7--11 \\
\bottomrule
\end{tabular}

%% file: Tables/tab_harness_ablation.tex
\begin{tabular}{lrrrrrrrr}
\toprule
\textbf{Harness} & \textbf{W} & \textbf{D} & \textbf{L} & \textbf{Score} & \textbf{Rate (95\% CI)} & \textbf{Red win} & \textbf{Blue win} & \textbf{BT (95\% CI)} \\
\midrule
mini-swe-agent & 4 & 1 & 1 & 93/180 & $0.52\,[0.43,\,0.60]$ & 0.13 & 0.90 & $+0.73\,[-0.25,\,+5.56]$ \\
Codex CLI & 2 & 1 & 3 & 90/180 & $0.50\,[0.42,\,0.60]$ & 0.11 & 0.89 & $-0.25\,[-3.50,\,+1.34]$ \\
Claude Code & 1 & 2 & 3 & 87/180 & $0.48\,[0.40,\,0.58]$ & 0.09 & 0.88 & $-0.49\,[-3.83,\,+0.65]$ \\
\bottomrule
\end{tabular}

%% file: Tables/tab_harness_pairings.tex
\begin{tabular}{llrrrr}
\toprule
\textbf{Harness A} & \textbf{Harness B} & \textbf{A score} & \textbf{B score} & \textbf{Margin (95\% CI)} & \textbf{Models} \\
\midrule
Claude Code & Codex CLI & 46 & 44 & $+2\,[-16,\,+22]$ & 3 \\
Codex CLI & mini-swe-agent & 46 & 44 & $+2\,[-16,\,+20]$ & 3 \\
Claude Code & mini-swe-agent & 41 & 49 & $-8\,[-28,\,+12]$ & 3 \\
\bottomrule
\end{tabular}

%% file: Tables/tab_costs.tex
\begin{tabular}{lrrrr}
\toprule
\textbf{Competitor} & \textbf{Gen \$} & \textbf{Val \$} & \textbf{Eval \$} & \textbf{Total \$} \\
\midrule
GPT-5.5 [codex] & 149.56 & 47.54 & 444.79 & 641.89 \\
GPT-5.5 [mini-swe-agent] & 48.38 & 17.15 & 181.65 & 247.18 \\
Claude Opus 4.8 [mini-swe-agent] & 44.93 & 22.12 & 156.54 & 223.59 \\
Gemini-3.1-Pro [mini-swe-agent] & 106.18 & 19.77 & 61.81 & 187.76 \\
Claude Opus 4.8 [claude-code] & 42.31 & 14.04 & 122.70 & 179.05 \\
Nemotron-3-Ultra [mini-swe-agent] & 84.66 & 2.85 & 83.82 & 171.34 \\
Qwen3.7-Max [mini-swe-agent] & 36.44 & 12.40 & 58.38 & 107.22 \\
GLM-5.2 [mini-swe-agent] & 43.51 & 7.14 & 54.32 & 104.97 \\
Kimi-K2.7-Code [mini-swe-agent] & 34.85 & 7.16 & 50.38 & 92.39 \\
MiniMax-M3 [mini-swe-agent] & 46.51 & 2.13 & 38.12 & 86.77 \\
DeepSeek-V4-Pro [mini-swe-agent] & 32.36 & 6.04 & 28.67 & 67.07 \\
\bottomrule
\end{tabular}

%% file: Tables/tab_defense_spend.tex
\begin{tabular}{lrrrrr}
\toprule
\textbf{Competitor} & \textbf{Turns} & \textbf{Success} & \textbf{Output M} & \textbf{Median k} & \textbf{$r$ [95\% CI]} \\
\midrule
Claude Opus 4.8 [mini-swe-agent] & 119 & 0.857 & 0.75 & 5.9 & $-0.04$ $[-0.21,+0.15]$ \\
Claude Opus 4.8 [claude-code] & 118 & 0.839 & 0.80 & 6.9 & $-0.18$ $[-0.35,+0.00]$ \\
GPT-5.5 [mini-swe-agent] & 111 & 0.964 & 0.84 & 7.3 & $-0.11$ $[-0.20,-0.00]$ \\
GPT-5.5 [codex] & 120 & 0.983 & 0.81 & 6.1 & $+0.09$ $[+0.00,+0.18]$ \\
DeepSeek-V4-Pro [mini-swe-agent] & 109 & 0.899 & 1.15 & 10.1 & $-0.01$ $[-0.23,+0.21]$ \\
GLM-5.2 [mini-swe-agent] & 111 & 0.874 & 0.78 & 6.6 & $+0.04$ $[-0.14,+0.22]$ \\
MiniMax-M3 [mini-swe-agent] & 129 & 0.705 & 1.58 & 11.1 & $-0.07$ $[-0.20,+0.08]$ \\
Kimi-K2.7-Code [mini-swe-agent] & 111 & 0.883 & 1.05 & 8.6 & $-0.04$ $[-0.25,+0.18]$ \\
Qwen3.7-Max [mini-swe-agent] & 109 & 0.890 & 0.84 & 7.0 & $-0.12$ $[-0.30,+0.09]$ \\
Nemotron-3-Ultra [mini-swe-agent] & 127 & 0.756 & 1.37 & 10.2 & $+0.38$ $[+0.21,+0.53]$ \\
Gemini-3.1-Pro [mini-swe-agent] & 55 & 0.618 & 0.64 & 8.6 & $+0.01$ $[-0.25,+0.26]$ \\
\bottomrule
\end{tabular}

%% file: Tables/tab_token_budget.tex
\begin{tabular}{rlrrrr}
\toprule
\textbf{Rank} & \textbf{Competitor} & \textbf{Solving BT} & \textbf{95\% interval} & \textbf{Rank interval} & \textbf{Median turns} \\
\midrule
1 & GPT-5.5 [codex] & +2.60 & $[+2.40,+4.94]$ & 1--1 & 94 \\
2 & GPT-5.5 [mini-swe-agent] & +1.38 & $[+0.97,+2.41]$ & 2--2 & 84 \\
3 & DeepSeek-V4-Pro [mini-swe-agent] & +0.17 & $[-0.32,+0.86]$ & 3--7 & 61 \\
4 & Qwen3.7-Max [mini-swe-agent] & +0.15 & $[-0.21,+0.57]$ & 3--6 & 83 \\
5 & Kimi-K2.7-Code [mini-swe-agent] & +0.02 & $[-0.44,+0.56]$ & 3--7 & 68 \\
6 & GLM-5.2 [mini-swe-agent] & -0.08 & $[-0.42,+0.28]$ & 3--7 & 91 \\
7 & Claude Opus 4.8 [mini-swe-agent] & -0.36 & $[-0.71,-0.07]$ & 5--8 & 101 \\
8 & Claude Opus 4.8 [claude-code] & -0.39 & $[-0.73,-0.09]$ & 6--8 & 86 \\
9 & Nemotron-3-Ultra [mini-swe-agent] & -0.98 & $[-1.44,-0.48]$ & 8--10 & 60 \\
10 & MiniMax-M3 [mini-swe-agent] & -1.18 & $[-1.63,-0.68]$ & 9--10 & 53 \\
11 & Gemini-3.1-Pro [mini-swe-agent] & -1.75 & $[-2.07,-1.56]$ & 11--11 & 55 \\
\bottomrule
\end{tabular}

%% file: Tables/tab_newcomer.tex
\begin{tabular}{lrr}
\toprule
\textbf{Cost component} & \textbf{USD} & \textbf{Share} \\
\midrule
\multicolumn{3}{l}{\emph{Main tournament}} \\
Task creation (incl. failed attempts) & 669.69 & 31.8\% \\
Blind self-review validation & 158.34 & 7.5\% \\
Blue evaluation & 1281.18 & 60.7\% \\
Tournament total & 2109.21 & 100.0\% \\
\midrule
\multicolumn{3}{l}{\emph{Entrance: Gemini-3.1-Pro [mini-swe-agent]}} \\
Bank authoring (task creation + validation) & 125.95 & 6.0\% \\
Entrant solving of incumbent tasks & 61.81 & 2.9\% \\
Incumbents solving entrant tasks & 55.64 & 2.6\% \\
Marginal entrance total & 243.40 & 11.5\% \\
\bottomrule
\end{tabular}

%% file: iclr2026_conference.bib
@misc{jimenez_swe-bench_2024,
	title = {{SWE}-bench: {Can} {Language} {Models} {Resolve} {Real}-{World} {GitHub} {Issues}?},
	shorttitle = {{SWE}-bench},
	url = {http://arxiv.org/abs/2310.06770},
	doi = {10.48550/arXiv.2310.06770},
	urldate = {2026-08-05},
	publisher = {arXiv},
	author = {Jimenez, Carlos E. and Yang, John and Wettig, Alexander and Yao, Shunyu and Pei, Kexin and Press, Ofir and Narasimhan, Karthik},
	month = nov,
	year = {2024},
	note = {arXiv:2310.06770 [cs.CL]},
}

@misc{chiang_chatbot_2024,
	title = {Chatbot {Arena}: {An} {Open} {Platform} for {Evaluating} {LLMs} by {Human} {Preference}},
	shorttitle = {Chatbot {Arena}},
	url = {http://arxiv.org/abs/2403.04132},
	doi = {10.48550/arXiv.2403.04132},
	urldate = {2026-08-05},
	publisher = {arXiv},
	author = {Chiang, Wei-Lin and Zheng, Lianmin and Sheng, Ying and Angelopoulos, Anastasios Nikolas and Li, Tianle and Li, Dacheng and Zhang, Hao and Zhu, Banghua and Jordan, Michael and Gonzalez, Joseph E. and Stoica, Ion},
	month = mar,
	year = {2024},
	note = {arXiv:2403.04132 [cs.AI]},
}

@article{bradley_rank_1952,
	title = {Rank {Analysis} of {Incomplete} {Block} {Designs}: {I}. {The} {Method} of {Paired} {Comparisons}},
	volume = {39},
	issn = {0006-3444},
	shorttitle = {Rank {Analysis} of {Incomplete} {Block} {Designs}},
	url = {https://www.jstor.org/stable/2334029},
	doi = {10.2307/2334029},
	number = {3/4},
	urldate = {2026-08-05},
	journal = {Biometrika},
	publisher = {[Oxford University Press, Biometrika Trust]},
	author = {Bradley, Ralph Allan and Terry, Milton E.},
	year = {1952},
	pages = {324--345},
}

@misc{yang_swe-smith_2025,
	title = {{SWE}-smith: {Scaling} {Data} for {Software} {Engineering} {Agents}},
	shorttitle = {{SWE}-smith},
	url = {http://arxiv.org/abs/2504.21798},
	doi = {10.48550/arXiv.2504.21798},
	urldate = {2026-08-05},
	publisher = {arXiv},
	author = {Yang, John and Lieret, Kilian and Jimenez, Carlos E. and Wettig, Alexander and Khandpur, Kabir and Zhang, Yanzhe and Hui, Binyuan and Press, Ofir and Schmidt, Ludwig and Yang, Diyi},
	month = may,
	year = {2025},
	note = {arXiv:2504.21798 [cs.SE]},
}

@misc{xu_swingarena_2026,
	title = {{SwingArena}: {Competitive} {Programming} {Arena} for {Long}-context {GitHub} {Issue} {Solving}},
	shorttitle = {{SwingArena}},
	url = {http://arxiv.org/abs/2505.23932},
	doi = {10.48550/arXiv.2505.23932},
	urldate = {2026-08-05},
	publisher = {arXiv},
	author = {Xu, Wendong and Xiong, Jing and Zhao, Chenyang and Chen, Qiujiang and Wang, Haoran and Shen, Hui and Wan, Zhongwei and Dai, Jianbo and Wu, Taiqiang and Xiao, He and Tao, Chaofan and Mao, Z. Morley and Sheng, Ying and Guo, Zhijiang and Yang, Hongxia and Yu, Bei and Kong, Lingpeng and Gu, Quanquan and Wong, Ngai},
	month = mar,
	year = {2026},
	note = {arXiv:2505.23932 [cs.CL]},
}

@misc{sonwane_bugpilot_2025,
	title = {{BugPilot}: {Complex} {Bug} {Generation} for {Efficient} {Learning} of {SWE} {Skills}},
	shorttitle = {{BugPilot}},
	url = {http://arxiv.org/abs/2510.19898},
	doi = {10.48550/arXiv.2510.19898},
	urldate = {2026-08-05},
	publisher = {arXiv},
	author = {Sonwane, Atharv and White, Isadora and Lee, Hyunji and Pereira, Matheus and Caccia, Lucas and Kim, Minseon and Shi, Zhengyan and Singh, Chinmay and Sordoni, Alessandro and Côté, Marc-Alexandre and Yuan, Xingdi},
	month = oct,
	year = {2025},
	note = {arXiv:2510.19898 [cs.SE]},
}

@inproceedings{zhao_auto-arena_2025,
	address = {Vienna, Austria},
	title = {Auto-{Arena}: {Automating} {LLM} {Evaluations} with {Agent} {Peer} {Battles} and {Committee} {Discussions}},
	isbn = {979-8-89176-251-0},
	shorttitle = {Auto-{Arena}},
	url = {https://aclanthology.org/2025.acl-long.223/},
	doi = {10.18653/v1/2025.acl-long.223},
	urldate = {2026-08-09},
	booktitle = {Proceedings of the 63rd {Annual} {Meeting} of the {Association} for {Computational} {Linguistics} ({Volume} 1: {Long} {Papers})},
	publisher = {Association for Computational Linguistics},
	author = {Zhao, Ruochen and Zhang, Wenxuan and Chia, Yew Ken and Xu, Weiwen and Zhao, Deli and Bing, Lidong},
	editor = {Che, Wanxiang and Nabende, Joyce and Shutova, Ekaterina and Pilehvar, Mohammad Taher},
	month = jul,
	year = {2025},
	pages = {4440--4463},
}

@inproceedings{son_arena-lite_2025,
	address = {Suzhou, China},
	title = {Arena-lite: {Efficient} and {Reliable} {Large} {Language} {Model} {Evaluation} via {Tournament}-{Based} {Direct} {Comparisons}},
	isbn = {979-8-89176-332-6},
	shorttitle = {Arena-lite},
	url = {https://aclanthology.org/2025.emnlp-main.360/},
	doi = {10.18653/v1/2025.emnlp-main.360},
	urldate = {2026-08-09},
	booktitle = {Proceedings of the 2025 {Conference} on {Empirical} {Methods} in {Natural} {Language} {Processing}},
	publisher = {Association for Computational Linguistics},
	author = {Son, Seonil and Oh, Ju-Min and Jin, Heegon and Jang, Cheolhun and Jeong, Jeongbeom and Kim, KunTae},
	editor = {Christodoulopoulos, Christos and Chakraborty, Tanmoy and Rose, Carolyn and Peng, Violet},
	month = nov,
	year = {2025},
	pages = {7057--7075},
}

@misc{deng_swe-bench_2025,
	title = {{SWE}-{Bench} {Pro}: {Can} {AI} {Agents} {Solve} {Long}-{Horizon} {Software} {Engineering} {Tasks}?},
	shorttitle = {{SWE}-{Bench} {Pro}},
	url = {http://arxiv.org/abs/2509.16941},
	doi = {10.48550/arXiv.2509.16941},
	urldate = {2026-08-09},
	publisher = {arXiv},
	author = {Deng, Xiang and Da, Jeff and Pan, Edwin and He, Yannis Yiming and Ide, Charles and Garg, Kanak and Lauffer, Niklas and Park, Andrew and Pasari, Nitin and Rane, Chetan and Sampath, Karmini and Krishnan, Maya and Kundurthy, Srivatsa and Hendryx, Sean and Wang, Zifan and Bharadwaj, Vijay and Holm, Jeff and Aluri, Raja and Zhang, Chen Bo Calvin and Jacobson, Noah and Liu, Bing and Kenstler, Brad},
	month = nov,
	year = {2025},
	note = {arXiv:2509.16941 [cs.SE]},
}

@misc{anthropic_run_2026,
	title = {Run {Claude} {Code} programmatically},
	url = {https://code.claude.com/docs/en/headless},
	language = {en},
	urldate = {2026-08-09},
	journal = {Claude Code Docs},
	author = {Anthropic},
	year = {2026},
}

@misc{openai_non-interactive_2026,
	title = {Non-interactive mode},
	url = {https://learn.chatgpt.com/docs/non-interactive-mode},
	language = {en},
	urldate = {2026-08-09},
	journal = {Non-interactive mode},
	author = {OpenAI},
	year = {2026},
}

@misc{yang_swe-agent_2024,
	title = {{SWE}-agent: {Agent}-{Computer} {Interfaces} {Enable} {Automated} {Software} {Engineering}},
	shorttitle = {{SWE}-agent},
	url = {http://arxiv.org/abs/2405.15793},
	doi = {10.48550/arXiv.2405.15793},
	urldate = {2026-08-09},
	publisher = {arXiv},
	author = {Yang, John and Jimenez, Carlos E. and Wettig, Alexander and Lieret, Kilian and Yao, Shunyu and Narasimhan, Karthik and Press, Ofir},
	month = nov,
	year = {2024},
	note = {arXiv:2405.15793 [cs.SE]},
}

@misc{arenaai_llm_2026,
	title = {{LLM} {Leaderboard} - {Best} {Text} \& {Chat} {AI} {Models} {Compared}},
	url = {https://arena.ai/leaderboard/text},
	language = {en},
	urldate = {2026-08-23},
	journal = {LLM Leaderboard - Best Text \& Chat AI Models Compared},
	author = {Arena.ai},
	year = {2026},
}

@misc{designarenaai_design_2026,
	title = {Design {Arena} {\textbar} {Leaderboards}},
	url = {https://designarena.ai/leaderboard},
	language = {en},
	urldate = {2026-08-23},
	journal = {Design Arena},
	author = {designarena.ai},
	year = {2026},
}

@misc{openai_introducing_2024,
	title = {Introducing {SWE}-bench {Verified}},
	url = {https://openai.com/index/introducing-swe-bench-verified/},
	language = {en-US},
	urldate = {2026-08-23},
	journal = {OpenAI},
	author = {OpenAI},
	month = aug,
	year = {2024},
}

@misc{badertdinov_swe-rebench_2025,
	title = {{SWE}-rebench: {An} {Automated} {Pipeline} for {Task} {Collection} and {Decontaminated} {Evaluation} of {Software} {Engineering} {Agents}},
	shorttitle = {{SWE}-rebench},
	url = {http://arxiv.org/abs/2505.20411},
	doi = {10.48550/arXiv.2505.20411},
	urldate = {2026-08-27},
	publisher = {arXiv},
	author = {Badertdinov, Ibragim and Golubev, Alexander and Nekrashevich, Maksim and Shevtsov, Anton and Karasik, Simon and Andriushchenko, Andrei and Trofimova, Maria and Litvintseva, Daria and Yangel, Boris},
	month = nov,
	year = {2025},
	note = {arXiv:2505.20411 [cs.SE]},
}

@misc{zhang_swe-bench_2025,
	title = {{SWE}-bench {Goes} {Live}!},
	url = {http://arxiv.org/abs/2505.23419},
	doi = {10.48550/arXiv.2505.23419},
	urldate = {2026-08-27},
	publisher = {arXiv},
	author = {Zhang, Linghao and He, Shilin and Zhang, Chaoyun and Kang, Yu and Li, Bowen and Xie, Chengxing and Wang, Junhao and Wang, Maoquan and Huang, Yufan and Fu, Shengyu and Nallipogu, Elsie and Lin, Qingwei and Dang, Yingnong and Rajmohan, Saravan and Zhang, Dongmei},
	month = jun,
	year = {2025},
	note = {arXiv:2505.23419 [cs.SE]},
}

@misc{liu_race-bench_2026,
	title = {{RACE}-{Bench}: {A} {Reasoning}-{Augmented} {Benchmark} for {Repository}-{Level} {Code} {Agents} on {Feature} {Addition}},
	shorttitle = {{RACE}-{Bench}},
	url = {http://arxiv.org/abs/2603.26337},
	doi = {10.48550/arXiv.2603.26337},
	urldate = {2026-09-12},
	publisher = {arXiv},
	author = {Liu, Shuhan and Zhao, Zhiyi and Hu, Xing and Liu, Kui and Yang, Xiaohu and Xia, Xin},
	month = aug,
	year = {2026},
	note = {arXiv:2603.26337 [cs.SE]},
}

@misc{le_swe-evo_2026,
	title = {{SWE}-{EVO}: {Benchmarking} {Coding} {Agents} in {Long}-{Horizon} {Software} {Evolution} {Scenarios}},
	shorttitle = {{SWE}-{EVO}},
	url = {http://arxiv.org/abs/2512.18470},
	doi = {10.48550/arXiv.2512.18470},
	urldate = {2026-09-12},
	publisher = {arXiv},
	author = {Le, Tue and Thai, Minh V. T. and Manh, Dung Nguyen and Nhat, Huy Phan and Bui, Nghi D. Q.},
	month = may,
	year = {2026},
	note = {arXiv:2512.18470 [cs.SE]},
}

@misc{chen_swe-ci_2026,
	title = {{SWE}-{CI}: {Evaluating} {Agent} {Capabilities} in {Maintaining} {Codebases} via {Continuous} {Integration}},
	shorttitle = {{SWE}-{CI}},
	url = {http://arxiv.org/abs/2603.03823},
	doi = {10.48550/arXiv.2603.03823},
	urldate = {2026-09-12},
	publisher = {arXiv},
	author = {Chen, Jialong and Xu, Xander and Wei, Hu and Chen, Chuan and Zhao, Bing},
	month = apr,
	year = {2026},
	note = {arXiv:2603.03823 [cs.SE]},
}

@misc{guan_swe-cycle_2026,
	title = {{SWE}-{Cycle}: {Benchmarking} {Code} {Agents} across the {Complete} {Issue} {Resolution} {Cycle}},
	shorttitle = {{SWE}-{Cycle}},
	url = {http://arxiv.org/abs/2605.13139},
	doi = {10.48550/arXiv.2605.13139},
	urldate = {2026-09-12},
	publisher = {arXiv},
	author = {Guan, Hao and Fu, Lingyue and Zhang, Shao and Zhu, Yaoming and Zhang, Kangning and Qiu, Lin and Cai, Xunliang and Cao, Xuezhi and Liu, Weiwen and Zhang, Weinan and Yu, Yong},
	month = may,
	year = {2026},
	note = {arXiv:2605.13139 [cs.SE]},
}

@misc{vergopoulos_automated_2025,
	title = {Automated {Benchmark} {Generation} for {Repository}-{Level} {Coding} {Tasks}},
	url = {http://arxiv.org/abs/2503.07701},
	doi = {10.48550/arXiv.2503.07701},
	urldate = {2026-09-12},
	publisher = {arXiv},
	author = {Vergopoulos, Konstantinos and Müller, Mark Niklas and Vechev, Martin},
	month = mar,
	year = {2025},
	note = {arXiv:2503.07701 [cs.SE]},
}

@misc{rashid_swe-polybench_2025,
	title = {{SWE}-{PolyBench}: {A} multi-language benchmark for repository level evaluation of coding agents},
	shorttitle = {{SWE}-{PolyBench}},
	url = {http://arxiv.org/abs/2504.08703},
	doi = {10.48550/arXiv.2504.08703},
	urldate = {2026-09-12},
	publisher = {arXiv},
	author = {Rashid, Muhammad Shihab and Bock, Christian and Zhuang, Yuan and Buchholz, Alexander and Esler, Tim and Valentin, Simon and Franceschi, Luca and Wistuba, Martin and Sivaprasad, Prabhu Teja and Kim, Woo Jung and Deoras, Anoop and Zappella, Giovanni and Callot, Laurent},
	month = apr,
	year = {2025},
	note = {arXiv:2504.08703 [cs.SE]},
}

@misc{matton_leakage_2024,
	title = {On {Leakage} of {Code} {Generation} {Evaluation} {Datasets}},
	url = {http://arxiv.org/abs/2407.07565},
	doi = {10.48550/arXiv.2407.07565},
	urldate = {2026-09-12},
	publisher = {arXiv},
	author = {Matton, Alexandre and Sherborne, Tom and Aumiller, Dennis and Tommasone, Elena and Alizadeh, Milad and He, Jingyi and Ma, Raymond and Voisin, Maxime and Gilsenan-McMahon, Ellen and Gallé, Matthias},
	month = oct,
	year = {2024},
	note = {arXiv:2407.07565 [cs.CL]},
}

@misc{guan_is_2026,
	title = {Is {Your} {Benchmark} {Still} {Useful}? {Dynamic} {Benchmarking} for {Code} {Language} {Models}},
	shorttitle = {Is {Your} {Benchmark} {Still} {Useful}?},
	url = {http://arxiv.org/abs/2503.06643},
	doi = {10.48550/arXiv.2503.06643},
	urldate = {2026-09-12},
	publisher = {arXiv},
	author = {Guan, Batu and Wu, Xiao and Yuan, Yuanyuan and Li, Shaohua},
	month = jul,
	year = {2026},
	note = {arXiv:2503.06643 [cs.SE]},
}

@misc{chen_dynamic_2025,
	title = {Dynamic {Benchmarking} of {Reasoning} {Capabilities} in {Code} {Large} {Language} {Models} {Under} {Data} {Contamination}},
	url = {http://arxiv.org/abs/2503.04149},
	doi = {10.48550/arXiv.2503.04149},
	urldate = {2026-09-12},
	publisher = {arXiv},
	author = {Chen, Simin and Pusarla, Pranav and Ray, Baishakhi},
	month = jun,
	year = {2025},
	note = {arXiv:2503.04149 [cs.SE]},
}

@inproceedings{roy_bug_2018,
	address = {Lake Buena Vista FL USA},
	title = {Bug synthesis: challenging bug-finding tools with deep faults},
	isbn = {978-1-4503-5573-5},
	shorttitle = {Bug synthesis},
	url = {https://dl.acm.org/doi/10.1145/3236024.3236084},
	doi = {10.1145/3236024.3236084},
	language = {en},
	urldate = {2026-09-12},
	booktitle = {Proceedings of the 2018 26th {ACM} {Joint} {Meeting} on {European} {Software} {Engineering} {Conference} and {Symposium} on the {Foundations} of {Software} {Engineering}},
	publisher = {ACM},
	author = {Roy, Subhajit and Pandey, Awanish and Dolan-Gavitt, Brendan and Hu, Yu},
	month = oct,
	year = {2018},
	pages = {224--234},
}

@inproceedings{patra_semantic_2021,
	address = {Athens Greece},
	title = {Semantic bug seeding: a learning-based approach for creating realistic bugs},
	isbn = {978-1-4503-8562-6},
	shorttitle = {Semantic bug seeding},
	url = {https://dl.acm.org/doi/10.1145/3468264.3468623},
	doi = {10.1145/3468264.3468623},
	language = {en},
	urldate = {2026-09-12},
	booktitle = {Proceedings of the 29th {ACM} {Joint} {Meeting} on {European} {Software} {Engineering} {Conference} and {Symposium} on the {Foundations} of {Software} {Engineering}},
	publisher = {ACM},
	author = {Patra, Jibesh and Pradel, Michael},
	month = aug,
	year = {2021},
	pages = {906--918},
}

@inproceedings{richter_learning_2022,
	title = {Learning {Realistic} {Mutations}: {Bug} {Creation} for {Neural} {Bug} {Detectors}},
	issn = {2159-4848},
	shorttitle = {Learning {Realistic} {Mutations}},
	url = {https://ieeexplore.ieee.org/document/9787824},
	doi = {10.1109/ICST53961.2022.00027},
	urldate = {2026-09-12},
	booktitle = {2022 {IEEE} {Conference} on {Software} {Testing}, {Verification} and {Validation} ({ICST})},
	author = {Richter, Cedric and Wehrheim, Heike},
	month = apr,
	year = {2022},
	note = {ISSN: 2159-4848},
	pages = {162--173},
}

@inproceedings{zhang_fixreverter_2022,
	title = {{FixReveRteR}: {A} {Realistic} {Bug} {Injection} {Methodology} for {Benchmarking} {Fuzz} {Testing}},
	language = {en},
	author = {Zhang, Zenong and Patterson, Zach},
	year = {2022},
}

@article{romazanov_cdbench_2026,
	title = {{CDBench}: {Benchmarking} the mutation testing capabilities of {LLMs} with code defenders},
	volume = {31},
	issn = {1573-7616},
	shorttitle = {{CDBench}},
	url = {https://doi.org/10.1007/s10664-026-10901-8},
	doi = {10.1007/s10664-026-10901-8},
	language = {en},
	number = {6},
	urldate = {2026-09-12},
	journal = {Empirical Software Engineering},
	author = {Romazanov, Artur and Fraser, Gordon and Herbold, Steffen},
	month = jun,
	year = {2026},
	pages = {166},
}

@misc{google_gemini_2026,
	title = {Gemini 3.1 {Pro}},
	url = {https://deepmind.google/models/gemini/pro/},
	urldate = {2026-09-16},
	author = {Google},
	month = feb,
	year = {2026},
}

@misc{kimi_kimi_2026,
	title = {Kimi {K2}.7 {Code}: {Open}-{Source} {Agentic} {Coding} {Model}},
	shorttitle = {Kimi {K2}.7 {Code}},
	url = {https://www.kimi.ai/resources/kimi-k2-7-code},
	language = {en},
	urldate = {2026-09-16},
	author = {Kimi},
	month = jul,
	year = {2026},
}

@misc{openai_introducing_2026,
	title = {Introducing {GPT}-5.5},
	url = {https://openai.com/index/introducing-gpt-5-5/},
	language = {en-US},
	urldate = {2026-09-16},
	author = {OpenAI},
	month = sep,
	year = {2026},
}

@misc{deepseek-ai_deepseek-v4_2026,
	title = {{DeepSeek}-{V4}: {Towards} {Highly} {Efficient} {Million}-{Token} {Context} {Intelligence}},
	shorttitle = {{DeepSeek}-{V4}},
	url = {http://arxiv.org/abs/2606.19348},
	doi = {10.48550/arXiv.2606.19348},
	urldate = {2026-09-16},
	publisher = {arXiv},
	author = {DeepSeek-AI and Xu, Anyi and Lin, Bangcai and Xue, Bing and Wang, Bingxuan and Xu, Bingzheng and Wu, Bochao and Zhang, Bowei and Lin, Chaofan and Dong, Chen and Ling, Chenchen and Lu, Chengda and Zhao, Chenggang and Deng, Chengqi and Hou, Chengyu and Xu, Chenhao and Shao, Chenze and Ruan, Chong and Sun, Conner and Dai, Damai and Guo, Daya and Yang, Dejian and Chen, Deli and Li, Donghao and Ji, Dongjie and Li, Erhang and Wei, Fang and Lin, Fangyun and Yuan, Fangzhou and Xia, Feiyu and Dai, Fucong and Hao, Guangbo and Chen, Guanting and Cao, Guoai and Meng, Guolai and Li, Guowei and Yu, Han and Zhang, Han and Xu, Hanwei and Li, Hao and Liang, Haofen and Zhang, Haoling and Luo, Haoming and Wei, Haoran and Yuan, Haotian and Zhang, Haowei and Luo, Haowen and Chen, Haoyu and Ji, Haozhe and Zhang, Hengqing and Ding, Honghui and Tang, Hongxuan and Cao, Huanqi and Gao, Huazuo and Qu, Hui and Zeng, Hui and Yang, J. and Zhu, J. Q. and Luo, Jia and Song, Jia and Yu, Jia and Huang, Jialiang and Cai, Jialu and Liang, Jian and Zhou, Jiangting and Ye, Jiasheng and Li, Jiashi and Xu, Jiaxin and Hu, Jiewen and Yang, Jieyu and Chen, Jin and Yan, Jin and Chen, Jingchang and Zhou, Jingli and Xiang, Jingting and Yuan, Jingyang and Cheng, Jingyuan and Zhou, Jingzi and Zhu, Jinhua and Yu, Jiping and Sun, Joseph and Ran, Jun and Jiang, Junguang and Qiu, Junjie and Li, Junlong and Zheng, Junmin and Song, Junxiao and Dong, Kai and Gao, Kaige and Guan, Kang and Zhou, Kexing and Huang, Kezhao and Yu, Kuai and Wang, Lean and Zhang, Lecong and Wang, Lei and Xia, Leyi and Zhang, Li and Zhao, Liang and Guo, Lihua and Luo, Lingxiao and Ma, Linwang and Zhu, Linyan and Wang, Litong and Cai, Liyu and Zhang, Liyue and Chen, Longhao and Di, M. S. and Xu, M. Y. and Mei, Max and Wang, Miaojun and Zhang, Mingchuan and Zhang, Minghua and Tang, Minghui and Li, Mingming and Zhou, Mingxu and Han, Minmin and Wang, Ning and Huang, Panpan and Wang, Panpan and Cong, Peixin and Wang, Peiyi and Zhang, Peng and Wang, Qiancheng and Zhu, Qihao and Li, Qingyang and Chen, Qinyu and Du, Qiushi and Jiang, Qiwei and Tian, Rui and Xu, Ruifan and Lu, Ruijie and Xu, Ruiling and Ge, Ruiqi and Zhang, Ruisong and Pan, Ruizhe and Wang, Runji and Chen, Runqian and Yin, Runqiu and Xu, Runxin and Shen, Ruomeng and Zhang, Ruoyu and Chen, Ruyi and Liu, S. H. and Lu, Shanghao and Sun, Shangmian and Zhou, Shangyan and Chen, Shanhuang and Cai, Shaofei and Nie, Shaoheng and Wu, Shaoqing and Chen, Shaoyuan and Hu, Shengding and Liu, Shengyu and Hu, Shiqiang and Ma, Shirong and Wang, Shiyu and Yu, Shuiping and Zhou, Shunfeng and Pan, Shuting and Yu, Shuying and Zhou, Songyang and Ni, Tao and Yun, Tao and Jin, Tian and Pei, Tian and Ye, Tian and Lin, Tianle and Ji, Tianran and Cui, Tianyi and Yue, Tianyuan and Yu, Tingting and Wang, Tun and Zhang, W. and Xiao, W. L. and Zeng, Wangding and An, Wei and Zhao, Weilin and Liu, Wen and Liang, Wenfeng and Pang, Wenjie and Luo, Wenjing and Yao, Wenjing and Gao, Wenjun and Yang, Wenkai and Huang, Wenlve and Hou, Wenqing and Zhang, Wentao and Ma, Wenting and Gao, Xi and He, Xiang and Wang, Xiangwen and Wang, Xianzu and Bi, Xiao and Liu, Xiaodong and Wang, Xiaohan and Chen, Xiaokang and Zhang, Xiaokang and Nie, Xiaotao and Sun, Xiaowen and Wang, Xiaoxiang and Cheng, Xin and Liu, Xin and Xie, Xin and Liu, Xingchao and Liu, Xingchen and Yu, Xingkai and Li, Xingyou and Yang, Xinyu and Zhang, Xinyu and Chen, Xu and Wang, Xuanyu and Su, Xuecheng and Chen, Xueyin and Lin, Xuheng and Fu, Xuwei and Yan, Y. C. and Wang, Y. Q. and Ma, Y. W. and Luo, Yanfeng and Zhang, Yang and Xu, Yanhong and Ma, Yanru and Huang, Yanwen and Li, Yao and Li, Yao and Xu, Yao and Zhao, Yao and Sun, Yaofeng and Wang, Yaohui and Qian, Yi and Shao, Yi and Yu, Yi and Zhang, Yichao and Ding, Yifan and Shi, Yifan and Wu, Yijia and Xiong, Yiliang and Ma, Yiling and He, Ying and Tang, Ying and Zhou, Ying and Luo, Yingjia and Zhong, Yinmin and Piao, Yishi and Wang, Yisong and Zhang, Yixiang and Chen, Yixiao and Tan, Yixuan and Wei, Yixuan and Ma, Yiyang and Liu, Yiyuan and Yang, Yonglun and Guo, Yongqiang and Wu, Yongtong and Wu, Yu and Li, YuKun and Cheng, Yuan and Ou, Yuan and Xu, Yuanfan and Li, Yuanhao and Wang, Yuduan and Yang, Yuehan and Xu, Yuer and Wu, Yuhan and Meng, Yuhao and Zou, Yuheng and Zha, Yukun and Xiong, Yunfan and Chen, Yupeng and Lin, Yuping and Cao, Yuqian and Wang, Yuqian and Zhang, Yushun and Yan, Yuting and Lin, Yutong and Gu, Yuxian and Luo, Yuxiang and You, Yuxiang and Liu, Yuxuan and Zhou, Yuxuan and Zhou, Yuyang and Huang, Yuzhen and Wu, Z. F. and Wang, Zehao and Zhao, Zehua and Ren, Zehui and Zhang, Zekai and Sha, Zhangli and Fu, Zhe and Ju, Zhe and Xu, Zhean and Xie, Zhenda and Zhang, Zhengyan and Gao, Zheren and Hao, Zhewen and Gou, Zhibin and Ma, Zhicheng and Yan, Zhigang and Shao, Zhihong and Huang, Zhixian and Chen, Zhixuan and Wu, Zhiyu and Ren, Zhizhou and Wu, Zhongyu and Li, Zhuoshu and Zhang, Zhuping and Xu, Zian and Wang, Zihao and Qu, Zihua and Gu, Zihui and Zhu, Zijia and Li, Zilin and Zhang, Zipeng and Xie, Ziwei and Gao, Ziyi and Wan, Ziyi and Pan, Zizheng and Yao, Zongqing},
	month = apr,
	year = {2026},
	note = {arXiv:2606.19348 [cs.CL]},
}

@misc{zai_glm-52_2026,
	title = {{GLM}-5.2: {Built} for {Long}-{Horizon} {Tasks}},
	shorttitle = {{GLM}-5.2},
	url = {https://z.ai/blog/glm-5.2},
	language = {en},
	urldate = {2026-09-16},
	author = {Z.ai},
	month = jun,
	year = {2026},
}

@misc{qwenteam_qwen37_2026,
	title = {Qwen3.7: {The} {Agent} {Frontier}},
	url = {https://qwen.ai/blog?id=qwen3.7},
	urldate = {2026-09-16},
	author = {QwenTeam},
	month = may,
	year = {2026},
}

@misc{nvidia_nemotron_2026,
	title = {Nemotron 3 {Ultra}: {Open}, {Efficient} {Mixture}-of-{Experts} {Hybrid} {Mamba}-{Transformer} {Model} for {Agentic} {Reasoning}},
	shorttitle = {Nemotron 3 {Ultra}},
	url = {http://arxiv.org/abs/2606.15007},
	doi = {10.48550/arXiv.2606.15007},
	urldate = {2026-09-16},
	publisher = {arXiv},
	author = {NVIDIA and Blakeman, Aaron and Thomas, Aaron and Jhunjhunwala, Aastha and Gupta, Abhibha and Khattar, Abhinav and Rajfer, Adam and Renduchintala, Adi and Asif, Adil and Vavre, Aditya and Miranda, Adriana Flores and Bilal, Ahmad and Zaman, Aileen and Hotchandani, Ajay and Shukla, Akanksha and Bercovich, Akhiad and Ficek, Aleksander and Gronskiy, Alex and Kondratenko, Alex and Steiner, Alex and Ye, Alex and Bukharin, Alexander and Milesi, Alexandre and Taghibakhshi, Ali and Gatti, Alice and Liu, Alisa and Kumar, Alok and Phanishayee, Amar and Mahabaleshwarkar, Ameya Sunil and Klein, Amir and Zuker, Amit and Geifman, Amnon and Bhiwandiwalla, Anahita and Subramaniam, Ananth and Santilli, Andrea and Fulks, Andrew and McHarg, Andrew and Tao, Andrew and Skliar, Andrii and Agrusa, Anjulie and Srivastava, Ankur and Verma, Ankur and Shors, Anna and Warno, Anna and Llaquet, Antoni-Joan Solergibert I. and Mehta, Arham and Nowaczynski, Arkadiusz and Jain, Arti and Aithal, Ashwath and Poojary, Ashwin and Ahamed, Asif and Mishra, Asit and Thekkumpate, Asma Kuriparambil and Sohrabizadeh, Atefeh and Kaur, Avinash and Vem, Avinash and Dattagupta, Ayush and Anandan, Barath Subramaniam and Sadeghi, Bardiya and Lanir, Ben and Schifferer, Benedikt and Nushi, Besmira and Kartal, Bilal and Thiede, Bill and Rouhani, Bita Darvish and Deng, Bo and Schatz, Bob and Ginsburg, Boris and Wang, Boxin and Nemire, Brad and Norick, Brandon and Dang, Brian and Westphal, Brian and Yu, Brian and Khailany, Brucek and Catanzaro, Bryan and Mundo, Carlo del and Aarish, Caryln and Lee, Chankyu and Hwang, Chantal and Sakr, Charbel and Wang, Charles and Truong, Charlie and Cui, Chen and Cheng, Cheng and Hsieh, Cheng-Ping and Zhang, Chenghao and Deng, Chenhui and Patel, Chintan and Alexiuk, Chris and Cosgrove, Christian and Munley, Christian and Harvey, Christine and Parisien, Christopher and Shen, Chunyang and Li, Coco and Neale, Collin and Gao, Cynthia and Meurillon, Cyril and Gil, Dan and Su, Dan and Zhao, Dan and Corneil, Dane and Afrimi, Daniel and Egert, Daniel and Korzekwa, Daniel and Lo, Daniel and Machlab, Daniel and Serebrenik, Daniel and Sorokin, Daniil and Gitman, Daria and Levy, Daria and Stosic, Darko and Mosallanezhad, David and Yu, David and Karamyan, Davit and Donia, Deena and Debroy, Deep and Narayanan, Deepak and O'Kelly, Devin and Peri, Dheeraj and Nathawani, Dhruv and Di and Wu and Rekesh, Dima and Kakwani, Divyanshu and Plummer, Donald and Anh, Dong and Yu, Dongfeng and Jiang, Dongfu and Kim, Donnie and Poorkay, Dorrin and Riach, Duncan and Stosic, Dusan and VanStee, Dustin and Meng, Eavan and Minasyan, Edgar and Lin, Edward and Long, Eileen Margaret Peters and Sarafin, Elad and Segal, Elad and Lantz, Elena and Evans, Ellie and Ning, Elliott and Chung, Eric and Harper, Eric and Pham-Hung, Eric and Tramel, Eric and Yang, Eric and Galinkin, Erick and Pounds, Erik and Goncalves, Erika Goncalves and Briones, Evan and Wu, Evan and Bakhturina, Evelina and Tsykunov, Evgeny and Dobrowolska, Ewa and Ladhak, Faisal and Memarian, Farzan and Wang, Fay and Jia, Fei and Soares, Felipe and Frujeri, Felipe Vieira and Chen, Feng and Lin, Fengguang and Galko, Ferenc and Sun, Frank and Siino, Frankie and Hou, Frida and Agam, Gal Hubara and Kaplun, Gal and Bhatt, Gantavya and Prasad, Gargi and Kulshreshtha, Garvit and Armstrong, George and Shen, Gerald and Borghesi, Giulio and Neskovic, Gordana and Batmaz, Gorkem and Lam, Grace and Mason, Greg and Pauloski, Greg and Nalbandyan, Grigor and Chlebus, Grzegorz and Karch, Grzegorz and Liu, Guan-Ting and Zhang, Guoming and Huang, Guyue and Maron, Haggai and Qian, Haifeng and Elisha, Haim and Ren, Haoxing and Kumar, Haran Kumar Shiv and Hud, Haribhau and Nover, Harris and Hall, Harrison Saturley and Iso, Hayate and Ngo, Helen and Hum, Herbert and Sahota, Herman and Wang, Hexin and Soni, Himanshu and Tamoyan, Hovhannes and Li, Hua and Chen, Huanhuan and Li, Hui and Wang, Hui and Nguyen, Huy and Chiles, Ian and Galil, Ido and Shahaf, Ido and Gitman, Igor and Shovkun, Igor and Loshchilov, Ilya and Guehring, Ingo and Schen, Itamar and Levy, Itay and Neeman, Itay and Moshkov, Ivan and Golan, Izik and Putterman, Izzy and Choi, Jaemin and Slowikowski, Jakub and Kautz, Jan and Scowcroft, Jane Polak and Casper, Jared and Mitra, Jatin and Glick, Jeffrey and Chen, Jenny and Oliver, Jesse and Xu, Jiacheng and Zhu, Jiafan and Song, Jialin and Zhang, Jian and Jiao, Jiantao and Zeng, Jiaqi and Lou, Jie and King, Jim and Zhang, Jimmy and Wang, Jingquan and Choi, Jinhang and Chu, Jinju and Conway, Joey and Guman, Joey and Jatko, Johan and Rausch, Johannes and Kamalu, John and Roberts, John and Greco, Johnny and Mensel, Johnny and Alben, Jonah and Yang, Jonas and Cohen, Jonathan and Raiman, Jonathan and Jennings, Joseph and Mabry, Joshua and Pierce, Joshua and Daw, Joyjit and Vialard, Julien Veron and Yi, Junkeun and Parmar, Jupinder and Jain, Kajal and Zhu, Kan and Briski, Kari and Cheung, Katherine and Luna, Katherine and Willowhawk, Keith and Wyss, Keith and Santhanam, Keshav and Shih, Kevin and Kong, Kezhi and Nguyen, Khanh and Bhardwaj, Khushi and Sivamani, Kirthi Shankar and Krommydas, Konstantinos and Puvvada, Krishna C. and Pawelec, Krzysztof and Anik, Kumar and Keprios, Kyle and Day, Kylie and McAfee, Lawrence and Du, Leo and Derczynski, Leon and Ding, Li and Liu, Linda and Wu, Lingjie and Kadoch, Lior and Wei, Lizzie and Vega, Luis and Robison, Luke and Su, Lun and Segbroeck, Maarten Van and Mikulski, Maciej Jakub and Melo, Maer Rodrigues de and Sypula, Magda and Fathi, Mahan and Sreedhar, Makesh Narsimhan and Chandran, Makesh Tarun and Kilaru, Manoj and Ashkenazi, Maor and Cuevas, Marc and Romeijn, Marc and Chochowski, Marcin and Cai, Mark and Mozolewski, Mark and Kliegl, Markus and Stepniewska-Dziubinska, Marta and Patelka, Martyna and Machczynski, Mattei and Novikov, Matvei and Ferrato, Mauricio and Golub, Maximilian and Samadi, Mehrzad and Corpuz, Melissa and Wang, Mengru and Wu, Mengxi and Price, Meredith and Boubdir, Meriem and Schaffer, Micah and Andersch, Michael and Boone, Michael and Gschwind, Michael and Lightstone, Michael and Loh, Michael and Bien, Michal and Zawalski, Michal and Gill, Michelle and Martinez, Miguel and Khona, Mikail and Chrzanowski, Mike and Houston, Mike and Ma, Mingyuan and Lee, Minseok and Fawzy, Mohamed and Dabbah, Mohammad and Shoeybi, Mohammad and Patwary, Mostofa and Mulepati, Nabin and Nabwani, Najeeb and Dhameja, Namit and Hennouni, Narimane and Hereth, Natalie and Pinckney, Nathaniel and Algarici, Nave and Assaf, Nave and Haber, Netanel and Knight, Nicholas and Reamaroon, Nick and Quak, Nickson and Bhatia, Nidhi and Desai, Nikhil and Ludwig, Nikolai and Tajbakhsh, Nima and Xu, Ning and Ailon, Nir and Juluru, Nirmal and Nitin, Nitin and Masad, Ofri and Rybakov, Oleg and Hrinchuk, Oleksii and Kuchaiev, Oleksii and Viessmann, Olivia and Delalleau, Olivier and Olabiyi, Oluwatobi and Argov, Omer Ullman and Puny, Omri and Tropp, Oren and Ribalta, Pablo and Bhattacharya, Pallab and Lampropoulos, Panos and Mannan, Parth and Shamis, Pasha and Legresley, Patrick and Gibbons, Paul and Molchanov, Pavlo and Morkisz, Pawel and Dykas, Peter and Jin, Peter and Aquilanti, Pierre-Yves and Xu, Pinky and Januszewski, Piotr and Laskiewicz, Piotr and Jannaty, Pooya and Gurumurthy, Prakash and Thombre, Pranav Prashant and Varshney, Prasoon and Gundecha, Pritam and Tredak, Przemek and Meng, Puhui and Wan, Qiyu and Mahabadi, Rabeeh Karimi and Oberman, Rachel and Garg, Rachit and Sri-Tharan, Radha and Kandu, Rahul and Sanadhya, Rakshit and El-Yaniv, Ran and Zilberstein, Ran and Shafipour, Rasoul and Macalisang, Ray and Tian, Rayen and Kovacs, Reka and Pi, Renjie and Izzo, Rick and Shahbazyan, Rima and Garg, Rishabh and Puri, Rishi and Neves, Rita Fernandes and Zhao, Ritchie and Borkar, Ritika and Gala, Ritu and Islam, Riyad and Clark, Robert and Hesse, Robert and Kirby, Robert and Waleffe, Roger and Watve, Rohit and Koren, Roi and Banner, Ron and Zhang, Ruoxi and Hewett, Russell J. and Prenger, Ryan and Stewart, Ryan and Egashira, Ryota and Mahdavi, Sadegh and Paliwal, Saee and Singh, Sagar and Modi, Sahil and Dave, Salika and Shinagawa, Samantha and Kriman, Samuel and Bhaskar, Sandip and Lym, Sangkug and Kariyappa, Sanjay and Satheesh, Sanjeev and Murari, Saran Vikas and Pasumarthi, Satish and Mishra, Saurabh and Muralidharan, Saurav and Hara, Scott and Narentharen, Sean and Anandaraj, Selvaraj and Na, Seonjin and Bak, Seonmeyong and Bak, Seonmyeong and Sameni, Sepehr and Mard, Seph and Panev, Serge and Henneman, Seth and Poulos, Seth and Mor, Shahar and Acharya, Shantanu and Ghosh, Shaona and Sreenivas, Sharath Turuvekere and Mendelson, Sharon and Kotek, Shaun and Wang, Shawn and Aharon, Shay and Gharghabi, Shaya and Lin, Sheng-Chieh and Chen, Shi and Fan, Shiqing and Baskaran, Shirish and Gopa, Shreya and Prabhumoye, Shrimai and Pachori, Shubham and Toshniwal, Shubham and Ding, Shuoyang and Krishnamurthy, Shwetha and Singh, Siddharth and Sun, Simeng and Das, Sirshak and Thottakara, Sivakumar Arayandi and Ithape, Smita and Majumdar, Somshubra and Singhal, Soumye and Singudasu, Sri Harsha and Bhuvanapalli, Sridhar and Veccham, Srimukh and Sergienko, Stas and Alborghetti, Stefania and Ge, Stephen and Rong, Su and Devare, Sugam Dipak and Rao, Sukrit and Barua, Sumeet Kumar and Ha, Sungsoo and Gai, Sunny and Gunasekar, Suriya and Panguluri, Suseella and Gupta, Suyog and Hinzburh, Sviataslau and Priyadarshi, Sweta and Akter, Syeda Nahida and Abramovich, Talor and Bui, Tan and Varshney, Tanay and Ter-Hovhannisyan, Tatevik and Ene, Teodor-Dumitru and Kong, Terry and Do, Thanh and Zhang, Tianhe and Moore, Tiffany and Blankevoort, Tijmen and Moon, Tim and Mitra, Tiyasa and Balough, Tom and Grzegorzek, Tomasz and Hliwiak, Tomasz and Asida, Tomer and Natan, Tomer Bar and Keren, Tomer and Ronen, Tomer and Salim, Tony and Wang, Tony and Rebedea, Traian and Konuk, Tugrul and Vashishth, Twinkle and Karpas, Udi and De, Ushnish and Noorozi, Vahid and Srinivasan, Venkat and Elango, Venmugil and Agrawal, Vibhor and Cui, Victor and Korthikanti, Vijay and Mehta, Vikas and Rao, Vinay and Wu, Virginia and Kurin, Vitaly and Lavrukhin, Vitaly and Anisimov, Vladimir and Pham, Vu and Jiang, Wanli and Ahmad, Wasi Uddin and Ishihara, Wataru and Du, Wei and Ping, Wei and Chai, Weiheng and Dai, Wenliang and Helmholz, Wesley and Jennings, Will and Zhu, Will and Prazuch, Wojciech and Ren, Xiaowei and Yu, Xiwen and Breek, Yan and Chen, Yang and Yu, Yang and Chen, Yangyi and Galron, Yaniv and Karnati, Yashaswi and Choi, Yejin and Meyer, Yev and Wu, Yi-Fu and Zhang, Yian and Lin, Ying and Geifman, Yonatan and Fu, Yonggan and Kwon, Youngeun and Yao, Yu and Guvvla, Yugi and Huang, Yuki and Liu, Yunsheng and Moshe, Zach and Newell, Zachary and Wang, Zhilin and Li, Zhiyu and Zhu, Zhongbo and Yang, Zhuolin and Liu, Zihan and Yan, Zijie and Wertheimer, Zsolt-Alon},
	month = jun,
	year = {2026},
	note = {arXiv:2606.15007 [cs.CL]},
}

@misc{minimax_minimax_2026,
	title = {{MiniMax} {M3}: {Frontier} {Coding}, {1M} {Context}, {Native} {Multimodality}},
	url = {https://www.minimax.io/blog/minimax-m3},
	urldate = {2026-09-16},
	author = {MiniMax},
	month = jun,
	year = {2026},
}

@misc{anthropic_introducing_2026,
	title = {Introducing {Claude} {Opus} 4.8},
	url = {https://www.anthropic.com/news/claude-opus-4-8},
	language = {en},
	urldate = {2026-09-16},
	author = {Anthropic},
	month = may,
	year = {2026},
}

@misc{wei_swe-rl_2025,
	title = {{SWE}-{RL}: {Advancing} {LLM} {Reasoning} via {Reinforcement} {Learning} on {Open} {Software} {Evolution}},
	shorttitle = {{SWE}-{RL}},
	url = {http://arxiv.org/abs/2502.18449},
	doi = {10.48550/arXiv.2502.18449},
	urldate = {2026-09-20},
	publisher = {arXiv},
	author = {Wei, Yuxiang and Duchenne, Olivier and Copet, Jade and Carbonneaux, Quentin and Zhang, Lingming and Fried, Daniel and Synnaeve, Gabriel and Singh, Rishabh and Wang, Sida I.},
	month = dec,
	year = {2025},
	note = {arXiv:2502.18449 [cs.SE]},
}

@misc{dataset_swe-duel_2026,
	title = {{SWE}-{Duel} {Experiment Dataset}},
	url = {https://osf.io/n7cdj/overview?view_only=afb5b2f1c939469d8c978996e460786d},
	author = {Wei Minn},
	year = {2026},
}

@misc{toolkit_swe-duel_2026,
	title = {{SWE}-{Duel} {Toolkit}},
	url = {https://github.com/sweduel/SWE-Duel},
	author = {Wei Minn},
	year = {2026},
}

@misc{leaderboard_swe-duel_2026,
	title = {{SWE}-{Duel} {Leaderboard}},
	url = {https://sweduel.github.io/SWE-Duel/},
	author = {Wei Minn},
	year = {2026},
}
